\documentclass[10pt,twocolumn,letterpaper]{article}
\usepackage{iccv}
\usepackage[table]{xcolor} 
\usepackage{times}
\usepackage{graphicx}
\usepackage{amsmath}
\usepackage{amssymb}
\usepackage{caption}
\usepackage{comment}
\usepackage{enumitem}
\usepackage{multirow}
\usepackage{siunitx}
\usepackage[all]{nowidow}
\usepackage{bm}
\usepackage[ruled,linesnumbered]{algorithm2e}
\usepackage{tabularx}
\usepackage{arydshln}
\usepackage{subcaption}
\usepackage[pagebackref=true,breaklinks=true,letterpaper=true,colorlinks,bookmarks=false]{hyperref}

\iccvfinalcopy % *** Uncomment this line for the final submission

\def\iccvPaperID{3285} % *** Enter the ICCV Paper ID here

\newcommand{\model}{ENVIDR}
\newcommand{\citep}{\cite}
\newcommand{\citea}{\cite}

\begin{document}

%%%%%%%%% TITLE
% \title{ENVIDR: Neural Approximation of Physically Based Rendering}
% \title{ENVIDR: Neural Approximation of Physically Based Rendering}
\title{\emph{\model}: Implicit Differentiable Renderer with Neural Environment Lighting}

\newcommand*{\affaddr}[1]{#1} % No op here. Customize it for different styles.
\newcommand*{\affmark}[1][*]{\textsuperscript{#1}}
\newcommand*{\email}[1]{\texttt{#1}}

\author{Ruofan Liang\affmark[1,2]\; Huiting Chen\affmark[1]\;
Chunlin Li\affmark[1]\; Fan Chen\affmark[1]\; Selvakumar Panneer\affmark[3]\; Nandita Vijaykumar\affmark[1,2]\\[8pt]
\affaddr{
\affmark[1]University of Toronto\quad
\affmark[2]Vector Institute\quad
\affmark[3]Intel Labs\quad
}\\[1pt]
% \affmark[2]Vector Institute, Canada}\\
% {\small\affmark[1]\texttt{\{ruofan,nandita\}@cs.toronto.edu}
% \; \affmark[1]\texttt{jiahaoz.zhang@mail.utoronto.ca}\\
% \affmark[2]\texttt{haoda\_li@berkeley.edu}\quad
% \affmark[3]\texttt{ycyangchen@sjtu.edu.cn}
% }
}

% \maketitle
\ificcvfinal\thispagestyle{empty}\fi

\twocolumn[{
\renewcommand\twocolumn[1][]{#1}%
\maketitle
\begin{center}
    \centering
    \small
    % \footnotesize
    \vspace{-8pt}
    \setlength\tabcolsep{0pt}
    \renewcommand{\arraystretch}{1.4}
    \begin{tabularx}{\linewidth}%
    {*{7}{>{\centering\arraybackslash}X}}
    Ground Truth & Ours & Ref-NeRF \citea{verbin2022ref} &
    % Mip-NeRF \citea{barron2021mip} & 
    VolSDF \citea{yariv2021volume} & 
    NVDiffRec \citea{munkberg2022extracting} &
    NVDiffRecMC \citea{hasselgren2022shape} 
    \\
    % \begin{annotate}
    {\includegraphics[width=\linewidth, trim={0 15pt 80pt 15pt},clip]{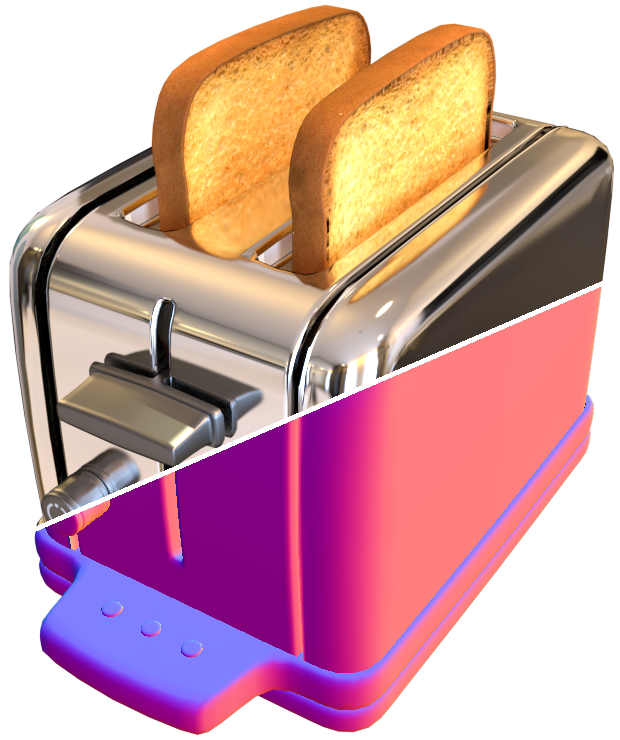}}{1}
    % \note{-0.35\linewidth, 0.6\linewidth}{\textbf{PNSR}} \note{0.39\linewidth, -0.6\linewidth}{\textbf{MAE}\si{\degree}}
    % \end{annotate} 
    &
    % \begin{annotate}
    {\includegraphics[width=\linewidth, trim={0 15pt 80pt 15pt},clip]{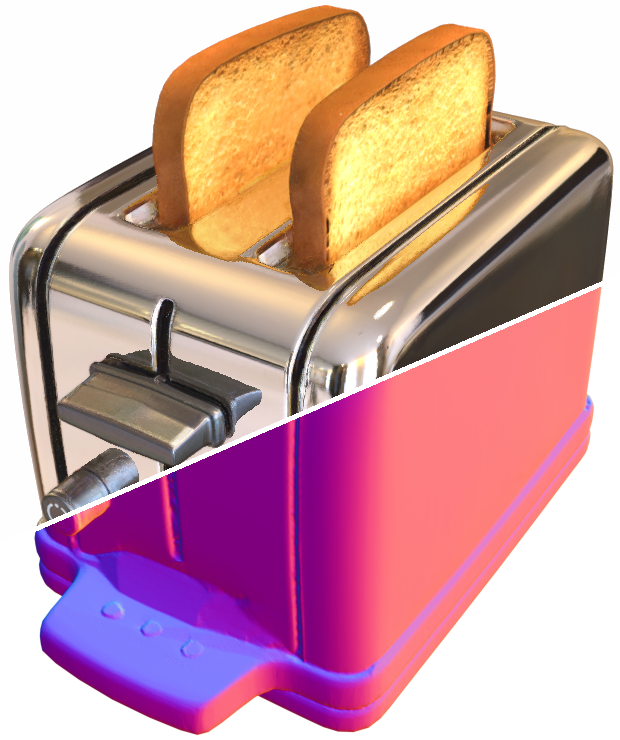}}{1}
    % \note{-0.35\linewidth, 0.6\linewidth}{28.21} \note{0.4\linewidth, -0.6\linewidth}{6.79}
    % \end{annotate} 
    &
    % \begin{annotate}
    {\includegraphics[width=\linewidth, trim={0 15pt 80pt 15pt},clip]{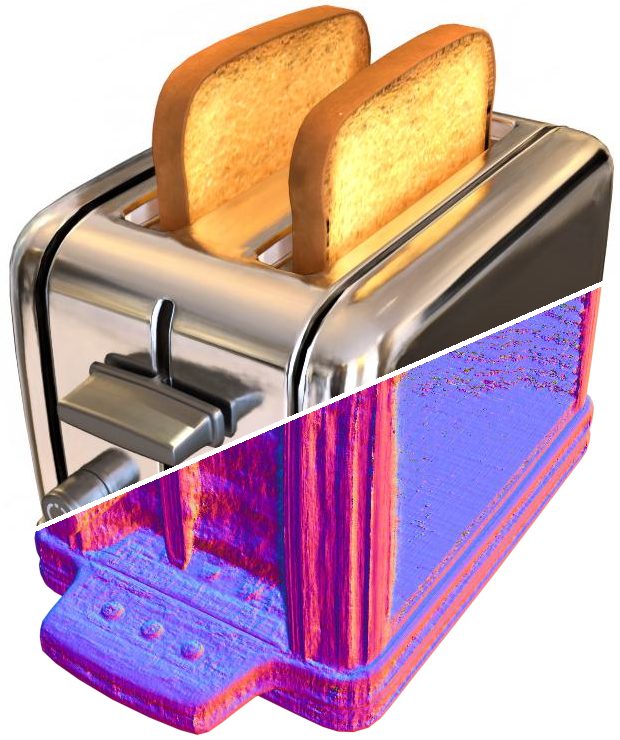}}{1}
    % \note{-0.35\linewidth, 0.6\linewidth}{28.76} \note{0.4\linewidth, -0.6\linewidth}{35.43}
    % \end{annotate}
    &
    % % \begin{annotate}
    % {\includegraphics[width=\linewidth, trim={0 15pt 80pt 20pt},clip]{figures/toaster_blend/mipnerf.png}}{1}
    % \note{-0.35\linewidth, 0.6\linewidth}{PNSR} \note{0.4\linewidth, -0.6\linewidth}{MAE}
    % % \end{annotate} 
    % &
    % \begin{annotate}
    {\includegraphics[width=\linewidth, trim={0 15pt 80pt 15pt},clip]{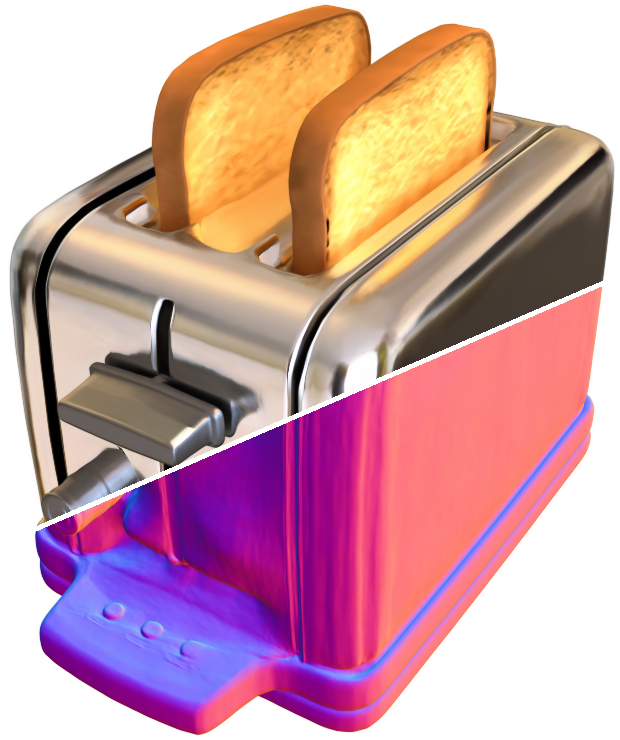}}{1}
    % \note{-0.35\linewidth, 0.6\linewidth}{27.37} \note{0.4\linewidth, -0.6\linewidth}{12.42}
    % \end{annotate} 
    &
    % \begin{annotate}
    {\includegraphics[width=\linewidth,  trim={0 15pt 80pt 15pt},clip]{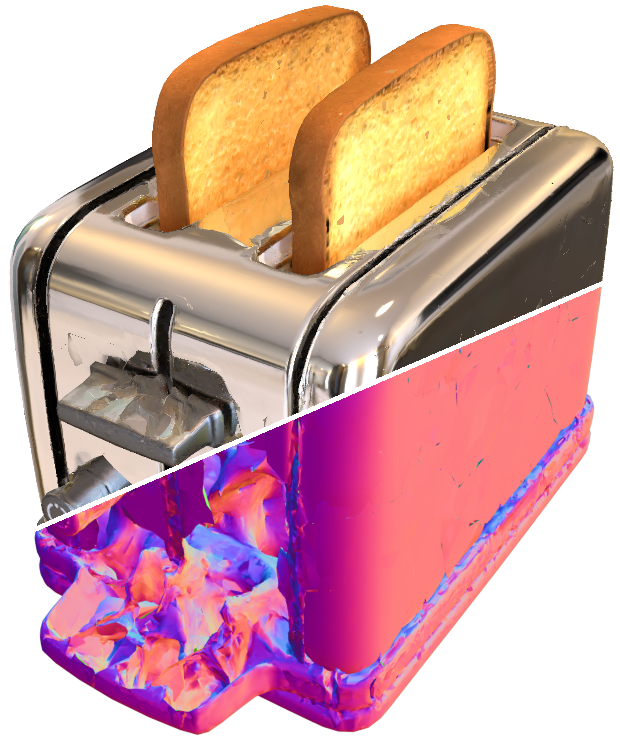}}{1}
    % \note{-0.35\linewidth, 0.6\linewidth}{25.01} \note{0.4\linewidth, -0.6\linewidth}{20.77}
    % \end{annotate} 
    &
    % \begin{annotate}
    {\includegraphics[width=\linewidth, trim={0 15pt 80pt 15pt},clip]{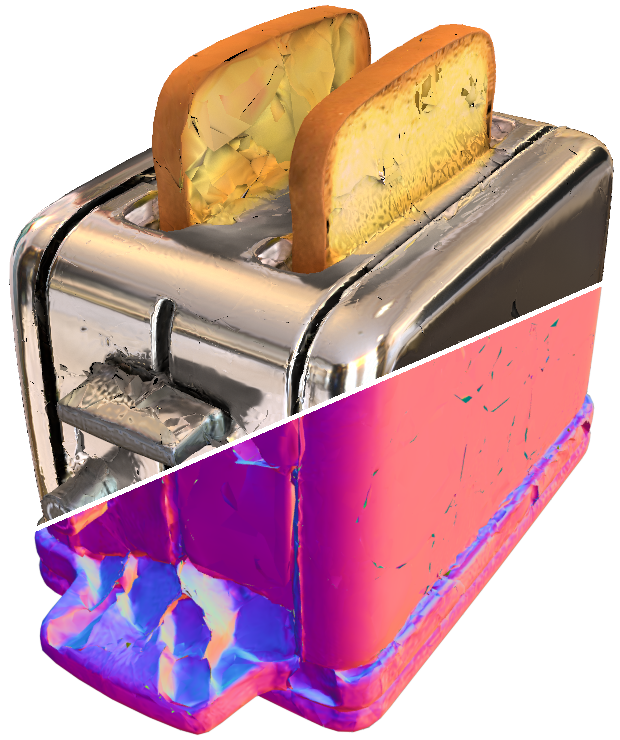}}{1}
    % \note{-0.35\linewidth, 0.6\linewidth}{22.67} \note{0.4\linewidth, -0.6\linewidth}{16.37}
    % \end{annotate}
    \end{tabularx}%
    \vspace{-10pt}
    \captionof{figure}{
    Compared to prior work, \emph{\model} has more accurate surface reconstruction and higher quality rendering for shiny objects with inter-reflections, as depicted for this ``toaster" scene. 
    The project page: \href{https://nexuslrf.github.io/ENVIDR}{\textit{\texttt{nexuslrf.github.io/ENVIDR}}}
    % \ruofan{TODO: may add PSNR/MAE to the figure}
    % NVDIFFREC \citea{munkberg2022extracting} exhibits higher rendering quality but less accurate surface, whereas NVDIFFRECMC \citea{hasselgren2022shape} demonstrates better surface quality but lower rendering quality.
    }
    \label{fig:toaster_compare}
    \label{fig:demo}
    % \vspace{-10pt}
\end{center}%
}]

% \begin{table*}[t]
%     \vspace{-10pt}
%     \centering
%     \footnotesize
%     \setlength\tabcolsep{0pt}
%     % \settowidth\rotheadsize{abcd}
%     \begin{tabularx}{\linewidth}%
%     {*{6}{>{\centering\arraybackslash}X}}
%     \multicolumn{6}{c}{\includegraphics[width=\linewidth]{figs/lego_ablation.pdf}}%
%     \\
%     \multicolumn{6}{c}{\includegraphics[width=\linewidth]{figs/lego_patch.pdf}}%
%     \\
%     Ground truth & w/o $\mathcal{L}_d$ & w/ $\mathcal{L}_d$ & GT normal & Normal, w/o $\mathcal{L}_d$ & Normal, w/ $\mathcal{L}_d$
%     \end{tabularx}%
%     \makeatletter\def\@captype{figure}\makeatother
%     \caption{Ablation on SDF continuity regularization $\mathcal{L}_d$ on Lego scene. We show the results of our non-BRDF model after 80k iterations.
%     The estimated normals shown in the figure are SDF computed normals $\hat{\mathbf{n}}$. \label{tab:lego_ablation}}
%     \vspace{-5pt}
% \end{table*}

%%%%%%%%% ABSTRACT
\begin{abstract}

Recent advances in neural rendering have shown great potential for reconstructing scenes from multiview images. 
However, accurately representing objects with glossy surfaces remains a challenge for existing methods.
In this work, we introduce \model, a rendering and modeling framework for high-quality rendering and reconstruction of surfaces with challenging specular reflections. 
To achieve this, we first propose a novel neural renderer with decomposed rendering components to learn the interaction between surface and environment lighting. This renderer is trained using existing physically based renderers and is decoupled from actual scene representations. 
We then propose an SDF-based neural surface model that leverages this learned neural renderer to represent general scenes.
Our model additionally synthesizes indirect illuminations caused by inter-reflections from shiny surfaces by marching surface-reflected rays. 
We demonstrate that our method outperforms state-of-art methods on challenging shiny scenes, providing high-quality rendering of specular reflections while also enabling material editing and scene relighting. 

% Our method outperforms prior methods on challenging shiny scenes, providing high-quality rendering of view-dependent specular reflections. The decomposed representation also enables our model to achieve scene editing such as scene  relighting and material editing.

% We achieve this by using a renderer with a neural approximation of the physically based rendering equation. We additionally synthesize the one-bounce inter-reflection from glossy surfaces. Our method outperforms prior methods on challenging shiny scenes, providing high-quality rendering of view-dependent specular reflections and flexible physically based scene editing.

\end{abstract}

%%%%%%%%% BODY TEXT
\vspace{-15pt}
\section{Introduction}
\begin{figure}
    \centering
    \includegraphics[width = 0.9\linewidth]{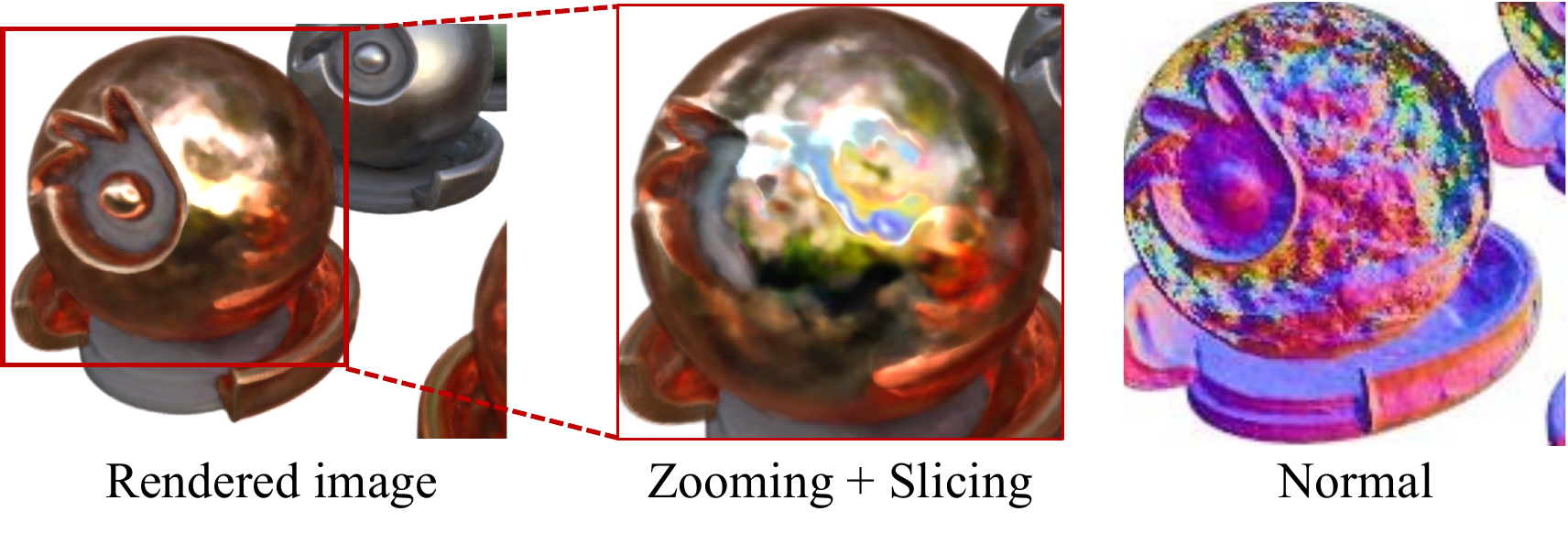}
    \vspace{-10pt}
    \caption{\small Artifacts in rendering surfaces with specular reflections due to the inaccurate interpretation of virtual lights underneath object surfaces (results from mip-NeRF \citea{barron2021mip}).}
    \label{fig:artifacts}
    \vspace{-15pt}
\end{figure}

\vspace{-5pt}
Neural Radiance Fields (NeRF) \citea{mildenhall2021nerf} has emerged as a promising approach to many important 3D computer vision and graphics tasks. By integrating deep learning with traditional volume rendering techniques, NeRF enables high-quality 3D scene modeling and reconstruction with photo-realistic rendering quality with significant recent research that has achieved impressive results \citea{mueller2022instant, munkberg2022extracting, poole2022dreamfusion, lin2022magic3d}.
% Its strong ability in representing 3D scenes is favored by a wide range of related 3D vision or graphics tasks, including novel view synthesis \citea{barron2021mip,barron2022mip}, relighting \citea{nerfactor, boss2021neural, munkberg2022extracting}, 
% geometry editing \citea{Yuan22NeRFEditing, xu2022deforming, liang2022spidr}, and 3D content creation \citea{wang2022clip, poole2022dreamfusion, lin2022magic3d}.
While NeRF can synthesize novel views with photo-realistic quality, they often struggle to accurately represent surfaces with high {specular reflectance}. Instead of learning a solid, smooth surface for these regions, NeRF models tend to interpret the view-dependent specular reflections as virtual lights/images underneath the actual surfaces (Figure \ref{fig:artifacts}). This leads to learning inaccurate surface geometry in the shiny regions.
% because the fake internal light sources have to transmit through the internal volume to synthesize view-dependent color effects. 
These virtual lights can also interfere with normal directions and negatively affect performance in inverse rendering tasks such as relighting and environment estimation.
This challenge has also been observed and analyzed by prior work, Verbin \etal \citea{verbin2022ref}, but is yet to be fully addressed. 

Prior work largely takes one of two major approaches to address the challenge of learning reflection in neural rendering. 
The first approach involves explicitly representing virtual lights or images underneath the surface to account for complex view-dependent appearance \citea{guo2022nerfren, wu2022scalable, kopanas2022neural, tiwary2022orca}. 
The original NeRF \citea{mildenhall2021nerf} and its extensions such as \citea{liu2020neural, barron2021mip, yu2021plenoctrees} also synthesize complex reflections in this way (Figure \ref{fig:artifacts}). 
% More recent work use additional neural models to separately represent virtual lights/images for more higher quality rendering of reflections \citea{guo2022nerfren, wu2022scalable, kopanas2022neural, tiwary2022orca}. 
Although this approach at large can improve rendering quality, it often sacrifices the accuracy of the reconstructed surface and limits the ability to edit scenes, such as relighting.
Alternatively, the second approach incorporates knowledge of inverse rendering to model the  interaction between light and surface \citea{zhang2021physg, nerfactor, boss2021neural, munkberg2022extracting}.
By decomposing rendering parameters, these methods can achieve material editing and scene relighting. 
However, these methods often suffer from relatively low rendering quality compared to top-performing NeRF models without full decomposition. This is because the simplified or approximated rendering equation \citea{kajiya1986rendering} used in these models cannot account for all complex rendering effects. 
Ref-NeRF \citea{verbin2022ref} improves the rendering of glossy objects with some decomposition; however its editability (e.g., relighting) is still limited as it does not fully decompose surface color and environment lighting.
In this work, we aim to further improve the quality of neural rendering for glossy surfaces, while retaining accurate surface geometry and the ability to edit scenes.

In this work, we introduce \emph{\model}, a new rendering and modeling framework for high-quality reconstructing and rendering of 3D objects with challenging specular reflections. It comprises two major parts: 
1) a novel neural renderer and 
2) an SDF-based neural surface model that represents the scene and interacts with the neural renderer. %that approximates physically based rendering with 3 decomposed MLP components accounting for environment lighting, diffuse BRDF rendering, and specular BRDF rendering, respectively (Figure \ref{fig:renderer} \ruofan{better on page 2/3}).

\begin{figure*}[t!]
    \centering
    % \includesvg[width=0.9\textwidth,inkscapelatex=false]{figures/renderer.svg}
    \includegraphics[width=\linewidth, trim={0, 0, 1.5cm, 0}, clip]{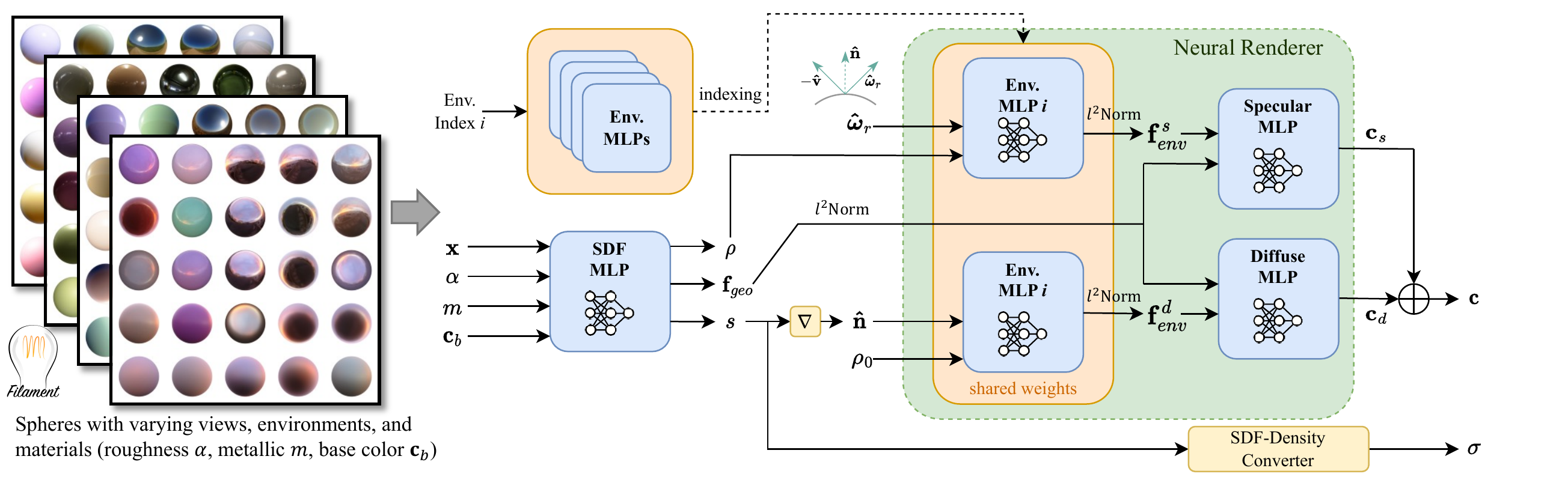}
    \caption{Overview of our proposed neural renderer. Training images are synthesized by Filament PBR engine \citea{google2018filament} during runtime, and probes are from HDRI Haven (CC-0). The MLPs used in our neural renderer are some simple and tiny MLPs.}
    \label{fig:renderer}
    \vspace{-10pt}
\end{figure*}

Our neural renderer is different from prior works \citea{zhang2021physg, nerfactor, boss2021neural, munkberg2022extracting} that incorporate the rendering equations for inverse rendering as we do not use an explicit form of the rendering equation.
Instead, our neural renderer learns an approximation of physically based rendering (PBR) using 3 decomposed MLPs accounting for environment lighting, diffuse rendering, and specular rendering, respectively (Figure \ref{fig:renderer}). 
This neural renderer is trained using images with various materials and environments synthesized by existing PBR renderers. 
In our renderer, the environment MLP is a decoupled component that is trained to represent the pre-integrated lighting of a specific environment with neural features as output (different from prior methods \citea{boss2021neural, gardner2022rotation, liang2022spidr} that outputs RGB). Thus, our neural renderer can be used for scene relighting and material editing by simply swapping out the environment MLP with the one that is trained to represent the desired environment map.
% Unlike prior methods \citea{boss2021neural, gardner2022rotation, liang2022spidr} that output RGB color, our environment MLP outputs neural features fed into diffuse/specular rendering MLPs.

% The environment MLP in our renderer is a replaceable component.
% Each environment MLP is trained to learn a pre-integrated representation of a specific environment map with the output in the form of neural feature vectors (this is different from prior methods \citea{boss2021neural, gardner2022rotation, liang2022spidr} that output RGB pixel).
% Thus, our neural renderer can be used for scene relighting and material editing by replacing the environment MLP with the one representing the desired environment lighting.

To interact with this neural renderer, we present a new neural surface model that employs an SDF-based neural representation (similar to 
 \citea{yariv2021volume}). We, however, use the diffuse/specular MLPs from the neural renderer in place of the commonly used directional color MLP. 
During training, we only train this SDF model and a new environment MLP without changing the pre-trained diffuse/specular MLPs in the neural renderer.

Finally, shiny surfaces may have inter-reflections that cause apparent view-dependent indirect illumination. 
To model this, we approximate the incoming radiance from inter-reflections by marching rays along the surface-reflected view directions. We additionally propose a color blending model that converts the approximated incoming radiance into indirect illumination and blends it into \model's final rendered color.

We demonstrate the effectiveness of our proposed method on several challenging shiny scenes, and our results show that it is quantitatively and qualitatively on par with or superior to previous methods. Our method achieves this while preserving high-quality decomposed rendering components, including diffuse color, specular color, material roughness, and environment light, which enables physically based scene editing.

\section{Related Work}
\textbf{Neural rendering and NeRF.}
Neural rendering is a class of reconstruction and rendering approaches that employ  neural methods to learn complex mappings from captured images to novel images \citea{tewari2020state}.
% The recent advances in neural rendering have shown promising results in various tasks \citea{thies2019deferred, lombardi2021mixture, chan2022efficient}.
% such as texture mapping \citea{thies2019deferred}, surface reconstruction \citea{yariv2020multiview}, image generation \citea{chan2022efficient}, etc. 
Neural radiance field (NeRF) \citea{mildenhall2020nerf} is one representative work that utilizes implicit neural representations and volume rendering for photo-realistic novel view synthesis.
% NeRF can render images from novel unobserved camera views by reconstructing the scene as a neural volume, represented as a coordinate-based multilayer perceptron (MLP).
% Given a set of sparsely-captured scene images, NeRF is able to render images from novel unobserved camera views through its 3D reconstruction of the scene as neural volume represented in the coordinate-based multilayer perceptron (MLP).
NeRF has inspired many follow-up works that achieve state-of-the-art performance in 3D rendering tasks \citea{barron2021mip, barron2022mip, verbin2022ref, tancik2022block}. 
Recent work also utilizes the hybrid neural representation to accelerate the training and rendering speed of NeRF models \citea{mueller2022instant, nerfstudio}, making it practical for real applications such as game and movie productions.
% One major issue is that the original NeRF method has low quality on the represented surface geometry because of the unconstrained volumetric representation. 
One major limitation of the original NeRF method is that its unconstrained volumetric representation leads to low-quality surface geometry.
Follow-up methods combine the implicit surface representation with NeRF \citea{yariv2021volume, wang2021neus} to enable volume rendering on neural surface representations for more accurate and continuous surface reconstruction.
% To improve the reconstructed surface quality, follow-up methods combine the implicit surface representation and NeRF \citea{yariv2021volume, wang2021neus} to enable the NeRF's volume rendering on the SDF-based neural surface representation. 
% These neural surface models can reconstruct accurate and continuous object surfaces through multiview images. 
% Recent work also shows that neural surface approaches can be employed in hybrid neural representations \citea{Yu2022MonoSDF, wang2022go, wu2022voxurf} for efficient and accurate surface reconstruction.
% While our model also utilizes a hybrid neural surface representation for more efficient learning, our primary focus is on accurately rendering and representing glossy surfaces.

\textbf{Rendering reflective and glossy surfaces.}
Rendering views in scenes with complex specular reflections has been challenging. Early methods use light field techniques \citea{gortler1996lumigraph, levoy1996light, wood2000surface, davis2012unstructured}, which require dense image capture. Recent approaches use learning-based methods to reconstruct the light field from a small set of images \citea{zhou2018stereo, flynn2019deepview, wizadwongsa2021nex}, but are limited by the number of available viewpoints.
Recent advances in neural rendering also show promising results in rendering reflective or glossy surfaces. 
NeRFReN \citea{guo2022nerfren} models planar reflections by learning a separate neural field. SNISR \citea{wu2022scalable} treats specular highlights as virtual lights underneath the surface with a reflection MLP.
Neural Catacaustics \citea{kopanas2022neural} uses a neural warp field to approximate the catacaustic through the virtual points for reflections.
% These neural methods synthesize the reflections through the virtual geometry (e.g., virtual lights/points) learned by specialized neural models. 
However, these methods do not model the physically based interaction between lighting and surface, limiting their ability to edit the lighting of represented scenes.
% Various neural methods \citea{guo2022nerfren, wu2022scalable, kopanas2022neural} model the reflections through virtual geometry learned by specialized neural models. However, these models do not account for the physically based interaction between environment light and surfaces, limiting their ability to edit scene lighting.
% 
Ref-NeRF \citea{verbin2022ref} conditions the view-dependent color on the reflected view direction and improves learning of surface normals, but it may still inaccurately learn virtual geometry from complex reflections (see Fig. \ref{fig:toaster_compare}).
% Furthermore, Ref-NeRF does not have a complete decomposition of environment light and surface color, which makes it challenging to do scene relighting.
% following the inspiration of physically based rendering \citea{ramamoorthi2002frequency, ramamoorthi2009precomputation}.
% Although Ref-NeRF improves the surface normal, it can still learn virtual geometry accounting for complex reflections (see Fig. \ref{fig:toaster_compare}).
% Another limitation of Ref-NeRF is that it does not have a complete decomposition of environment light and surface color, which disables its ability to relight scenes.
% Inspired by Ref-NeRF, we incorporate IDE in our decoupled environment light representation to generate pre-integrated environment feature vectors.
% Unlike Ref-NeRF, our model utilizes surface representation to constrain the surface normal, our model also decouples environment light from the directional MLP to achieve scene relighting.
In contrast, our model uses the surface-based representation to constrain the surface normal, and it decouples environment light from the directional MLP to achieve scene relighting.

\textbf{Neural inverse rendering.}
Inverse rendering aims to estimate surface geometry, material properties and lighting conditions from images \citea{marschner1998inverse}. 
% Prior works have shown accurate estimation of reflection properties, such as BRDF, under known lighting conditions \citea{matusik2003data,aittala2016reflectance,deschaintre2018single}, and estimation of environment light from objects with known geometry \citea{richter2016instant, legendre2019deeplight, park2020seeing}.
Recently, NeRF methods have been employed in inverse rendering tasks to learn scene lighting and material properties (e.g., BRDF). 
However, prior work such as \citea{bi2020neural, nerv2021} requires known lighting conditions to learn materials. 
More recent works such as \citea{zhang2021physg, boss2021neural, nerfactor, munkberg2022extracting, hasselgren2022shape} jointly estimate environment light and materials with images under unknown lighting conditions. These methods employ different representations to model the environment light, such as learnable spherical Gaussians \citea{boss2021nerd, zhang2021physg}, pre-integrated environment texture \citea{boss2021neural}, HDR light probes \citea{nerfactor, liang2022spidr, munkberg2022extracting, hasselgren2022shape}.
% Recent work also leverages NeRF methods to learn scene lighting and material reflectance. 
% Methods such as \citea{bi2020neural, nerv2021} attempt to learn material reflectance properties with NeRF, but require known lighting conditions.
% More recent work jointly estimates environment light and reflectance with images under unknown lighting conditions \citea{zhang2021physg, boss2021neural, nerfactor, munkberg2022extracting, hasselgren2022shape}.
% \citea{boss2021nerd, zhang2021physg} use learnable spherical Gaussian (SG) to represent the environment light.
% Neural-PIL \citea{boss2021neural} uses MLP to represent pre-integrated environment maps.
% \citea{nerfactor, liang2022spidr} directly learn the HDR light probes for the environment light.
% Other methods such as \citea{munkberg2022extracting, hasselgren2022shape} learn mesh-based 3D representations and optimize structured material textures and environment textures through a differentiable renderer.
With the estimated decomposed rendering parameters, previous neural inverse rendering methods rely on an approximated or simplified rendering equation \citea{kajiya1986rendering} to synthesize or edit the scene,  limiting their ability to achieve high-quality renderings comparable to top-performing NeRF models.
In contrast, our model uses a neural renderer to learn the physically based interaction between surface and environment through existing PBR renderers, without explicitly formulating the rendering equation. 
Similar to \citea{boss2021neural, gardner2022rotation, liang2022spidr}, our model also uses MLPs to represent environment lights, however, the output of our environment MLP is neural features instead of RGB colors.
% We represent the environment lights as neural feature vectors that are fused with surface features to synthesize the novel views.
Regarding indirect illumination, 
\citea{nerv2021, zhang2022modeling} incorporate indirect illumination in their model, but their approximation may not work well on shiny surfaces. We instead directly march the surface-reflected rays to synthesize indirect lighting on shiny surfaces.

\begin{figure*}
    \centering
    \includegraphics[width=\linewidth]{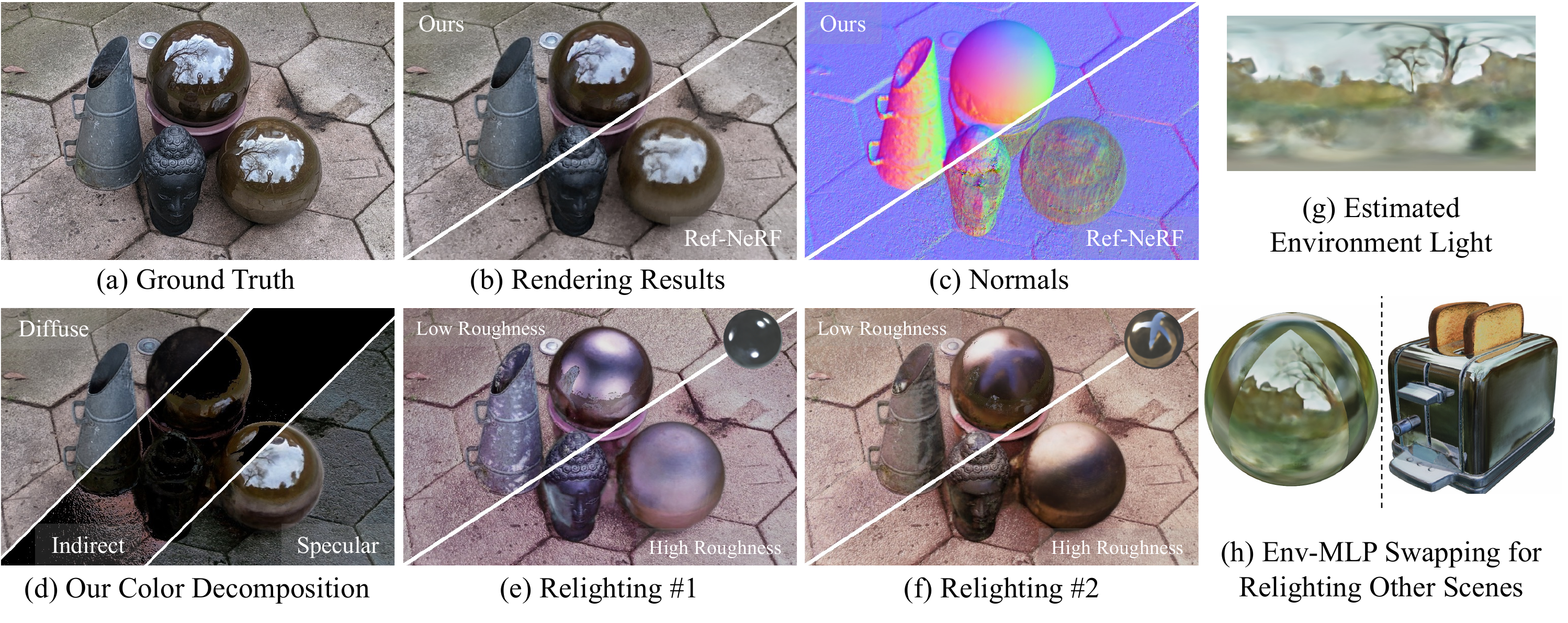}
    \vspace{-20pt}
    \caption{\small
    \emph{\model} achieves high-quality rendering (b) and reconstruction (c) of scenes with glossy/shiny surfaces. Our method represents diffuse, direct specular, and indirect specular colors separately (d), and enables various scene relighting and material editing (e-h) using our proposed neural renderer with decomposed rendering components. Ref-NeRF's results are reproduced based on the official code.
    }
    \label{fig:garden}
\vspace{-10pt}
\end{figure*}

\section{Preliminaries}
\subsection{Volume Rendering with Neural Surface} \label{sec:nerf_surface}
Instead of representing volume density like NeRF \citea{mildenhall2021nerf}, neural surface methods \citea{park2019deepsdf, yariv2020multiview} use implicit neural representation to represent scene geometry as signed distance fields (SDF).  
For a given 3D point $\mathbf x \in \mathbb R^3$, SDF returns the point’s distance to the closest surface $\mathbf x \mapsto s = \bm{F}_\theta(\mathbf{x})$, $\bm{F}_\theta$ denotes the neural spatial representation with learnable parameters $\theta$. $\bm{F}_\theta$ can be either a fully implicit MLP or a hybrid model containing voxel-based features \citea{mueller2022instant, Yu2022MonoSDF, wu2022voxurf}.

% Recent work \citea{oechsle2021unisurf, yariv2021volume,wang2021neus} optimizes such implicit surface representations via NeRF-like differentiable volume rendering. 
To render a pixel, a ray $\mathbf{r}:\mathbf{o} + t \mathbf{\hat{v}}$ is cast from the camera's origin $\mathbf{o}$ along its view direction $\mathbf{\hat{v}}$. 
The SDF value $s_i$ of sampled point $\mathbf{x}_i$ along the ray are then converted to density or opacity value for volume rendering. VolSDF \citea{yariv2021volume} demonstrates a density conversion method with the cumulative distribution function (CDF) of Laplace distribution:
\begin{equation}
    \sigma_\beta(s) = \begin{cases}
\frac{1}{2\beta}\exp(\frac{s}{\beta}) & \text{if } s \leq 0,\\
\frac{1}{\beta}(1 - \frac{1}{2}\exp(-\frac{s}{\beta})) & \text{if } s > 0
\end{cases}\label{eq:conversion}
\end{equation}
Where $\sigma$ is converted volume density, $\beta$ is a learnable parameter.
With the predicted color $\mathbf{c}(\mathbf{x}_i)$ of sampled points along the ray, the color ${\mathbf C}(\mathbf r)$ for the current ray $\mathbf r$ is integrated with volume rendering \citea{max1995optical}:
% \begin{align}
%     {\mathbf C}(\mathbf r) &= \sum_{i} w_i \mathbf c(\mathbf{x}_i) \notag \\
%     \text{with } w_i = \exp(-\sum_{j=1}^{i-1}\sigma&(\mathbf{x}_j) \delta_j)(1 - \exp(-\sigma(\mathbf{x}_j) \delta_j) ) \label{eq_radiance}
% \end{align}
\vspace{-8pt}
\begin{equation}
    {\mathbf C}(\mathbf r) = \sum_{i} \exp(-\sum_{j=1}^{i-1}\sigma_j \delta_j)(1 - \exp(-\sigma_j \delta_j) )  \mathbf c(\mathbf{x}_i)
\label{eq_radiance}
\vspace{-8pt}
\end{equation}
where $\delta_i$ denotes the distance between adjacent sampled positions along the ray.

% \subsection{View-dependent Color Prediction}
% The original NeRF uses directional MLP $\bm{R}(\mathbf f_{geo}, \mathbf{\hat{v}})$, conditioned on additional neural features $\mathbf{f}_{geo}$ from the spatial geometry model $\bm F$ and view direction $\mathbf{\hat{v}}$, to predict view-dependent color. 
% IDR \citea{yariv2020multiview} includes normal direction $\mathbf{\hat{n}}$, as the extra input to the color prediction MLP for the better modeling of the underlying bidirectional reflectance distribution function (BDRF) of the surface. The normal direction $\mathbf{\hat{n}}$ can be computed by the analytical gradient of neural field (SDF or density) $\bm{s}$ w.r.t. position $\mathbf x$: $\mathbf{\hat{n}} = \pm \frac{\nabla \bm{s}(\mathbf x)}{\|\nabla \bm{s}(\mathbf x)\|_2}$
% % \begin{equation}
% %     \mathbf{\hat{n}} = \pm \frac{\nabla \bm{s}(\mathbf x)}{\|\nabla \bm{s}(\mathbf x)\|_2}
% % \end{equation}
% % $c(\mathbf z, \mathbf{\hat{v}}, \mathbf{\hat{n}})$

% Some works \citea{zhang2021physg, boss2021neural, verbin2022ref} use the reflected view direction $\bm{\hat{\omega}}_r$ to help neural rendering model better learn continuous spatially-varying illumination, where %$\bm{\hat{\omega}}_r$ can be computed by 
% $\bm{\hat{\omega}}_r = \mathbf{\hat{v}} - 2(\mathbf{\hat{v}} \cdot \mathbf{\hat{n}})\mathbf{\hat{v}}$.
% \begin{equation}
%  \bm{\hat{\omega}}_r = \mathbf{\hat{v}} - 2(\mathbf{\hat{v}} \cdot \mathbf{\hat{n}})\mathbf{\hat{v}}   
% \end{equation}

\subsection{The Rendering Equation} \label{sec:re} 
Mathematically, the outgoing radiance of a surface point $\mathbf{x}$ with normal $\hat{\mathbf{n}}$ from outgoing direction $\hat{\bm{\omega}}_o$ can be described by the physically based rendering equation \citea{kajiya1986rendering}:
\vspace{-5pt}
\begin{equation}
% \vspace{-5pt}
    L_o(\mathbf{x}, \hat{\bm{\omega}}_o) = \int_{\Omega}  L_i(\mathbf{x},  \hat{\bm{\omega}}_i) f_r(\mathbf{x}, \hat{\bm{\omega}}_i, \hat{\bm{\omega}}_o) (\hat{\mathbf{n}}\cdot \hat{\bm{\omega}}_i)  d \hat{\bm{\omega}}_i \label{eq:re}
\end{equation}
where $\hat{\bm{\omega}}_i$ denotes incoming light direction, $\Omega$ denotes the hemisphere centered at $\hat{\mathbf{n}}$, $L_i(\mathbf{x},  \hat{\bm{\omega}}_i)$ is the incoming radiance of $\mathbf{x}$ from $\hat{\bm{\omega}}_i$, $f_r$ is the BRDF that describes the surface response of incoming lights. BDRF can be expressed as a function with diffuse $f_d$ and specular $f_s$ components \citea{walter2007microfacet}:
\begin{equation}
    f_r = f_d + f_s \label{eq:brdf_decompose}
\end{equation}
% Diffuse BRDF $f_d$ is view-independent, while $f_r$ depends on the view and surface normal directions. However, both components interact with the light denoted by $L_i$.

% \subsection{Multi-resolution Hash encoding.}
% Recently, Instant-NGP~\citea{mueller2022instant} proposes to represent the whole 3D space with multi-resolution grids stored in a hash table.
% Instead of using a fully implicit MLP, Instant-NGP uses a combination of implicit and explicit neural networks to represent the 3D scene. For each query point position $\mathbf{x}$, instant-NGP's hash encoding outputs its neural feature $\mathbf{h}_{\mathbf{x}}^{(l)}$ by interpolating the feature grids at each level $l$. Then the features from all resolution levels are concatenated together as the point's encoded neural features $\mathbf{h}_{\mathbf{x}}$. Then these features are fed into a shallow MLP for predicting density and radiance.
% Thanks to this efficient structure, Instant-NGP accelerates the training and rendering stage of NeRF by a large margin without obvious performance degradation.
% \section{Methods}

% In this section, we first describe how we build and learn a neural renderer with the approximation of a PB renderer. By utilizing this neural renderer, \model\ can learn accurate decomposition of the rendering parameter and synthesize high-quality novel views. The decomposition of the environment light also enables \model\ to relight scenes.
% Finally, we introduce our neural blending for indirect illumination.

\section{Neural Renderer: Approximating the PBR} \label{sec:renderer}
% \ruofan{Repeat: Instead of explicitly ...}
From the rendering equation described in \ref{sec:re}, we know PBR  is the result of the interaction between the surface material and lighting. %, and BRDF can be decomposed into diffuse and specular components.
The neural renderer proposed in this work attempts to learn a neural approximation of the PBR, instead of \emph{explicitly} formulating a rendering equation.
In addition to the geometry MLP $\bm{F}$ for surface geometry representation, we use three MLPs: environment light MLP $\bm{E}(\cdot)$, diffuse MLP $\bm{R}_d(\cdot)$, and specular MLP $\bm{R}_s(\cdot)$ in place of NeRF's directional MLP. 
Similar to the decomposition of BDRF in PBR (Eqn. \ref{eq:brdf_decompose}), our diffuse MLP and specular MLP implicitly learn the rendering rules for the corresponding color components.
The environment light MLP encodes the distant light probes of a specific environment into neural features that interact with surface geometry features (i.e., feature fusion) through diffuse/specular MLPs.

\subsection{Decomposed Rendering MLPs}\label{sec:color_mlp}
\noindent
\textbf{Environment Light MLP.} Similar to \citea{boss2021neural, liang2022spidr, gardner2022rotation}, we represent the environment light probes as a coordinate-based MLP, conditioned on light directions. However, the output of our environment MLP is a high-dimensional neural feature vector instead of a valid HDR pixel. These neural environment features help our neural renderer capture 
 complex surface-lighting interactions compared to RGB pixels. 
For efficient rendering,  we use a similar approach as Neural-PIL \citea{boss2021neural} to represent pre-integrated environment light. Therefore, our environment MLP also requires roughness as the additional input. We employ the integrated directional encoding (IDE) \citea{verbin2022ref} over input direction and roughness to better learn continuous, high-frequency environment feature vectors. 
More specifically, given a light direction $\bm{\hat{\omega}}$, and a roughness value\footnote{It should note that the roughness $\rho$ used in IDE does not have the same meaning as the perceptual roughness $\alpha$ used in analytic BRDF models.}
$\rho$, our environment MLP $\bm{E}$ returns a neural feature vector $\mathbf{f}_{env}$:
\begin{equation}
    \mathbf{f}_{env} = \bm{E}(\bm{\hat{\omega}}, \rho)
\end{equation}
$\mathbf{f}_{env}$ is then used for synthesizing diffuse/specular colors. 
To change the environment lighting of the rendered scene, we can swap the environment MLP to achieve this. 

\noindent
\textbf{Diffuse MLP.} The diffuse MLP learns color synthesis from the diffuse (Lambertian) BRDF. Since the irradiance of diffuse color is a cosine-weighted integration of environment light over a hemisphere centered at surface normal direction $\mathbf{\hat{n}}$, the diffuse color is independent of view direction and surface roughness. Based on this, we use the normal direction $\mathbf{\hat{n}}$ and a constant high roughness value $\rho_0$ (we empirically set $\rho_0=0.64$) as the input to our environment MLP to query the environment neural feature vector $\mathbf{f}_{env}^{d} = \bm{E}(\mathbf{\hat{n}}, \rho_0)$. Environment features $\mathbf{f}_{env}^{d}$ are then concatenated with geometry features $\mathbf{f}_{geo}$ as the input to the diffuse MLP $\bm{R}_d$:
\begin{equation}
\vspace{-3pt}
    \mathbf{c}_d = \bm{R}_d(\mathbf{f}_{geo},\ \mathbf{f}_{env}^d)
\vspace{-1pt}
\end{equation}

\noindent
\textbf{Specular MLP.} As the counterpart of the diffuse MLP, specular MLP learns color synthesis from the specular BRDF. The commonly used analytic BRDF model \citea{cook1982reflectance, walter2007microfacet} also depends on view direction and roughness. The specular BRDF lobe\footnote{We leave the anisotropic or refraction effects to the future exploration.} is centered around the direction of specular reflection and its shape is controlled by material roughness $\rho$ and the angle between the outgoing direction $\bm{\hat{\omega}}_o$ ($\bm{\hat{\omega}}_o=-\mathbf{\hat{v}}$) and surface normal $\mathbf{\hat{n}}$.
Similarly, the outgoing radiance at $\bm{\hat{\omega}}_o$ is also a integral over the weight distribution of incoming lights that is centered around reflected view direction $\bm{\hat{\omega}}_r$.
Therefore, we use the reflected view direction $\bm{\hat{\omega}}_r$ and the predicted roughness $\rho$ (via geometry MLP $\bm F$) to query environment MLP $\bm E$ for the environment feature vector $\mathbf{f}_{env}^{s} = \bm{E}(\bm{\hat{\omega}}_r, \rho)$.
Environment features, geometry features, and the dot product between  $\bm{\hat{\omega}}_o$ and $\mathbf{\hat{n}}_r$. are then combined as the input to the specular MLP $\bm{R}_s$:
\begin{equation}
    \mathbf{c}_s = \bm{R}_s(\mathbf{f}_{geo},\ \mathbf{f}_{env}^s, \bm{\hat{\omega}}_o\cdot \mathbf{\hat{n}})
\end{equation}
Finally, the synthesized diffuse and specular colors after volume rendering (Eq. \ref{eq_radiance}) are additively combined in the linear space and then converted to sRGB space with gamma tone mapping \citea{anderson1996proposal}:
\begin{equation}
    \mathbf{C} = \gamma(\mathbf{C}_d + \mathbf{C}_s)
\end{equation}

% \begin{figure}[t]
%     \centering
%     \includesvg[width=0.9\linewidth,inkscapelatex=false]{figures/sphere_demo.svg}
%     \caption{\textbf{TODO}}
%     \label{fig:sphere_demo}
% \end{figure}

\subsection{Training the Neural Renderer} \label{sec:train_render}
% Combining the decomposed color MLPs described in \ref{sec:color_mlp}, we then make our color MLPs learn to approximate the actual PBR.

We train our neural renderer using synthesized images of a sphere with various materials and environment lighting rendered by an existing PBR renderer as depicted in Figure \ref{fig:renderer}. 
Specifically, we use Filament \citea{google2018filament}, a real-time PBR engine to synthesize these images by varying perceptual roughness $\alpha$, metallic value $m$, and base color $\mathbf{c}_b$ for the surface material, as well as different distant light probes for environment lighting.
To render the same sphere with our neural renderer, we employ a simple MLP $\bm{F}_{sphere}$ (similar to the one introduced in \ref{sec:nerf_surface}) to represent the sphere surface with SDF and output geometry features $\mathbf{f}_{geo}$. % for diffuse and specular MLP. 
$\bm{F}_{sphere}$ is also conditioned on the three material attributes ($\alpha, m, \mathbf{c}_b$) to account for the changes in geometry features caused by varying material properties. 
% The structure of our rendering model is illustrated in Figure \ref{fig:renderer}. 

To train our model, we construct an L1 photometric loss between images synthesized by the PBR renderer $\mathbf{C}^*$ and ones synthesized by our renderer $\mathbf{C}$. 
Other than the photometric loss, we also use the ground truth SDF $s^*$ to supervise the SDF prediction of the sphere (MSE). The loss function is formulated as:
\begin{equation}
\vspace{-3pt}
    \mathcal{L}_r = \mathcal{L}_{rgb}(\mathbf{C}, \mathbf{C}^*) + \lambda_1 \mathcal{L}_{SDF}(s, s^*) + \lambda_2 \mathcal{L}_{eik}(\nabla{s})
\end{equation}
Where $L_{eik}$ is Eikonal loss \citea{gropp2020implicit}, $\lambda_1$ \& $\lambda_2$ are loss weights which we set to 0.1 and 0.01 respectively. Figure \ref{fig:sphere_demo} showcases the controllable rendering results of our renderer.
Once the neural renderer is trained, we will freeze the weights of diffuse/specular MLPs for the rest experiments.
% \ruofan{frozen}
% we use the existing PBR renderers to synthesize images with varied 
% \ruofan{How to train, data}
% \ruofan{}

% For simplicity, we build a neural surface model to render a reflective sphere. In addition to the color models, we include a special SDF MLP $\bm{F}_{sphere}$ to represent a sphere with standard material \footnote{We do not consider anisotropic or transparent effects in our renderer.}. 
% In addition to the spatial coordinate, $\bm{F}_{sphere}$ is also conditioned on material attributes including perceptual roughness $\alpha$, metallic $m$, and the base color $\mathbf{c}_b$ to generate varying neural geometry features:
% \begin{equation}
%     s, \rho, \mathbf{f}_{geo} = \bm{F}_{sphere}(\mathbf{x}, \alpha, m, \mathbf{c}_b)
% \end{equation}
% For the environment light, we use multiple environment MLPs to represent different environment maps appearing in the training images, while the rest MLP modules are always shared.
% To render a pixel, SDF $s$ is converted to density value following Eq. \ref{eq:conversion} for volume rendering. 
% Figure \ref{fig:renderer} gives an overview of our neural renderer and its learning process. 

\begin{table}[]
    \centering
    % \small
    \setlength\tabcolsep{0pt}
        \begin{tabularx}{\linewidth}%
        {*{6}{>{\centering\arraybackslash}X}} % p{0.19\linewidth}
 \multicolumn{6}{c}{\includegraphics[width=\linewidth, trim={0 0 0 0},clip]{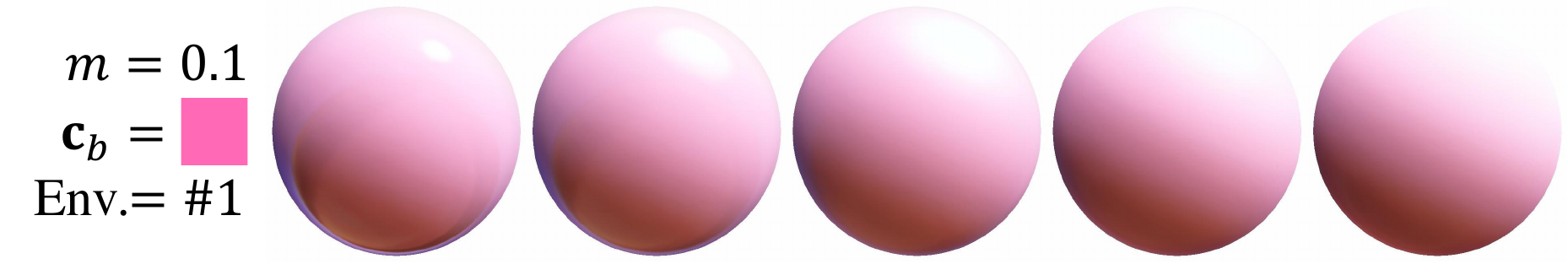}}\\
 \multicolumn{6}{c}{\includegraphics[width=\linewidth, trim={0 0 0 0},clip]{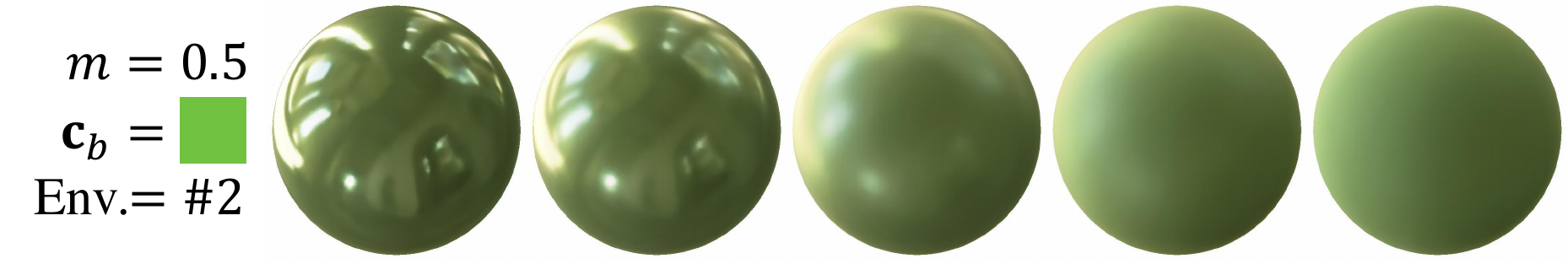}}\\
 \multicolumn{6}{c}{\includegraphics[width=\linewidth, trim={0 0 0 0},clip]{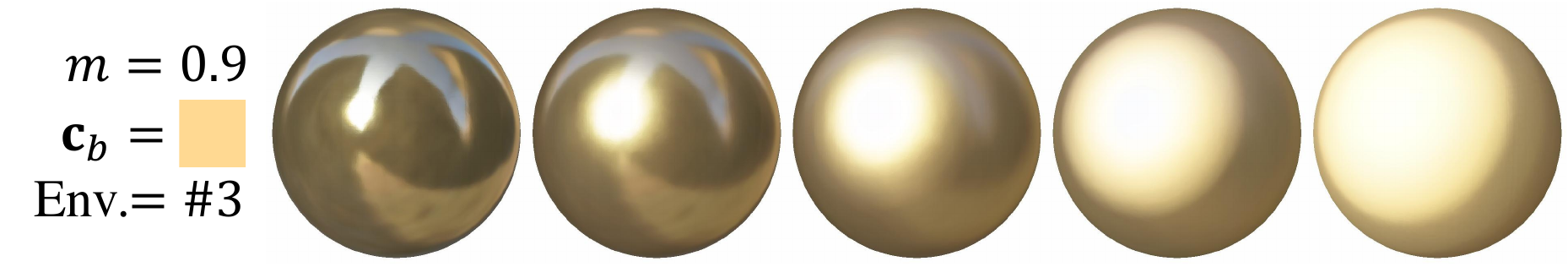}}\\
           $\alpha$  & 0.0 & 0.2 & 0.4 & 0.6 & 0.8 
        \end{tabularx}%
    \vspace{-8pt}
    \makeatletter\def\@captype{figure}\makeatother
    \caption{Spheres sythesized by our neural renderer with varying metallic $m$, base color $\mathbf{c}_b$, roughness $\alpha$, and light probes. 
    We will show an interactive web demo in the future.
    \label{fig:sphere_demo}}
    \vspace{-20pt}
\end{table}

\subsection{Feature Normalization} \label{sec:normalization}
% Our learned neural renderer is not only for rendering synthetic spheres, we can combine these pre-trained color MLPs with other spatial neural field backbone models for rendering arbitrary scenes. 
% In our neural renderer, we model the interaction between environment light and object surface as the feature fusion (concatenation) of environment features $\mathbf{f}_{env}$ and geometry features $\mathbf{f}_{geo}$. The fused features are then processed by diffuse/specular MLPs. 
Unlike the prior work that estimates all rendering parameters with physical meanings, our high-dimensional neural features are unconstrained, since they are simply the outputs of a linear net layer. There could be multiple possible feature mappings to the RGB colors in the high-dimensional space, which could cause mismatched colors in relighting results when swapping the environment MLP learned from two different scenes.
To constrain the neural features for more plausible relighting results with environment MLP swapping, we propose to normalize the neural features ($\mathbf{f}_{env}$ and $\mathbf{f}_{geo}$) with $l^2$-norm of each feature vector:
\vspace{-5pt}
\begin{equation}
    \mathbf{f'} = \mathbf{f} / \|\mathbf{f}\|_2
\vspace{-5pt}
\end{equation}
This normalization maps neural features to the unit vector on a hypersphere manifold, which improves the feature interchangeability among different represented neural scenes. 
We will give an empirical analysis of this in Section \ref{sec:ablation_renderer}.
% https://stats.stackexchange.com/questions/248511/purpose-of-l2-normalization-for-triplet-network

\section{Neural Rendering for General Scenes} \label{sec:envidr}
% In this section, we describe how we use the trained color MLPs from \ref{sec:train_render} together with neural geometry representation models to reconstruct and render general 3D scenes. We then introduce our approach to rendering indirect illumination caused by inter-reflection.
\subsection{Neural Surface Representations}
Following \citea{mueller2022instant, wang2022go, Yu2022MonoSDF}, our approach utilizes a hybrid neural SDF representation $\bm{F}_{g}$ with multi-resolution feature grids and hash encoding for the efficient learning and rendering of scene surfaces.
Given an input query position $\mathbf{x}$, $\bm{F}_g$  converts coordinate input $\mathbf{x}$ into a concatenated feature vector from the multi-resolution hash encoding sampled with trilinear interpolation (the encoding used in Instant-NGP \citea{mueller2022instant}). The encoded features are then fed into a shallow MLP to predict all of SDF $s$, roughness $\rho$, and geometry feature $\mathbf{f}_{geo}$.
% Note that compared to $\bm{F}_{sphere}$  that is used for learning the neural renderer, $\bm{F}_{g}$ for general scenes is not conditioned on any explicit material properties. 
Note that unlike $\bm{F}_{sphere}$ used for learning the neural renderer, $\bm{F}_{g}$ for general scenes is not conditioned on any explicit material properties. Instead, $\bm{F}_{g}$ implicitly learns the material properties through the multiview training images and encodes the knowledge into its geometry feature $\mathbf{f}_{geo}$.
\begin{equation}
    s, \rho, \mathbf{f}_{geo} = \bm{F}_{g}(\mathbf{x})
\vspace{-3pt}
\end{equation}

For the color rendering, we utilize pre-trained diffuse and specular MLPs ($\bm{R}_d$ \& $\bm{R}_s$) from \ref{sec:train_render} to synthesize output colors. The weights of both $\bm{R}_d$ and $\bm{R}_s$ remain \emph{frozen} throughout training. To estimate unknown environment light from training images, we randomly initialize an environment MLP $\bm{E}_g$ and optimize it alongside the geometry model $\bm{F}_{g}$. 
By combining all these components, we introduce our neural scene representation and rendering model, which we call \emph{\model}.
Similar to the training of other neural surface models \citea{yariv2021volume, wang2021neus}, 
Our training is supervised by L1 photometric loss and the Eikonal constraint \citea{gropp2020implicit}:
\begin{equation}
    \mathcal{L} = \mathcal{L}_{rgb}(\mathbf{C}, \mathbf{C}^*) + \lambda_{eik} \mathcal{L}_{eik}(\nabla{s})
\end{equation}
where $\lambda_{eik}$ is a weight hyperparameter which we set to 0.01.
To ensure a smooth geometry initialization at the beginning, we add additional SDF regularizations at the early training iterations, please refer to the supplement for details.

\begin{figure}[t]
    \centering
    \begin{subfigure}{0.49\linewidth}
    \centering\captionsetup{width=0.96\linewidth}%
    \includegraphics[height = 85pt, trim={10pt 0 0 0},clip]{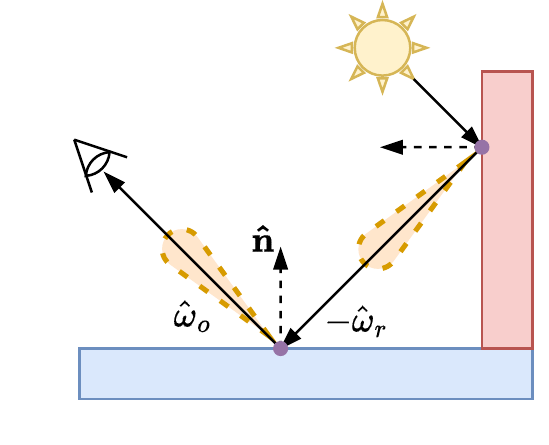}
    \vspace{-10pt}
    \caption{Tracing ray's reflection path to the camera.}
    \label{fig:mic_normal}
    \end{subfigure}
    \begin{subfigure}{0.49\linewidth}
    \centering\captionsetup{width=0.96\linewidth}%
    \includegraphics[height = 85pt,  trim={10pt 0 0 0},clip]{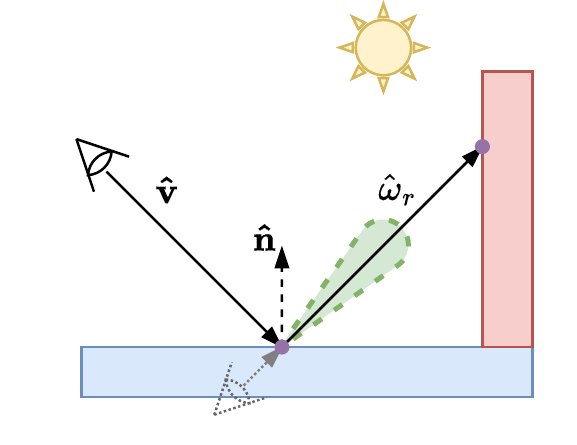}
    \vspace{-10pt}
    \caption{The raymarching process to capture pixel color.}
    \label{fig:mic_normal}
    \end{subfigure}
    \caption{The illustrations of inter-reflections from the (a) physical lighting process and (b) raymarching rendering process. We assume the surface has a low roughness value. 
    }
    \label{fig:indir_ref}
    \vspace{-10pt}
\end{figure}

\subsection{Ray Marching for Inter-reflections} \label{sec:indir}
Our learned neural renderer from \ref{sec:renderer} only models the image-based lighting from distant light probes. As a result, \model\ in Section \ref{sec:envidr} cannot effectively handle the indirect illumination caused by surface inter-reflection. Inter-reflection can negatively impact the inverse rendering process for mirror-like specular surfaces as shown in Figure \ref{fig:toaster_compare}. 
We observe that most indirect illumination effects arise from reflective surfaces with low roughness values. Although rougher surfaces can also be affected by indirect illumination, the resulting visual effects are less apparent. Thus, we focus on synthesizing inter-reflection on reflective surfaces with predicted roughness lower than a threshold $\rho_s$ (set to 0.1 in our experiments).

The weight distribution of incoming lights for outgoing radiance at $\bm{\hat{\omega}}_o$
on surfaces with low roughness is similar to a specular BRDF with rays concentrated at the reflected view direction $\bm{\hat{\omega}}_r$. Thus, to efficiently approximate the incoming radiance of the indirect illumination caused by inter-reflection on these surfaces, we can perform additional raymarching (one-bounce) along the reflected view directions (see Fig. \ref{fig:indir_ref}). This raymarching is similar to the raymarching along camera rays, but the origin and direction are set to the surface point $\mathbf{p}_{s}$ and $\bm{\hat{\omega}}_r$.
% This process is similar to screen space reflection (SSR) \citea{sousa2011secrets}, but we perform the raymarching in the 3D object space.

To render indirect illumination from approximated incoming radiance $\mathbf{e}_{r}$ from rendered reflected rays, 
we introduce another color encoding MLP $\bm{E}_{ref}$ to convert rendered reflected ray color into neural features $\mathbf{f}_{env}^{ref}$ compatible with our specular MLP $\bm{R}_s$. The rendered indirect illumination $\mathbf{c}_{ref}$ is then output by:
\begin{equation}
    \mathbf{c}_{ref} = \bm{R}_s(\mathbf{f}_{geo}, \mathbf{f}_{env}^{ref}, \bm{\hat{\omega}}_o\cdot \mathbf{\hat{n}}),\; \mathbf{f}_{env}^{ref} = \bm{E}_{ref}(\mathbf{e}_{r}, \rho)
\end{equation}
To blend the rendered indirect illumination into our final rendering results, we let the geometry MLP $\bm{F}_{g}$ to additionally predict a blending factor $\eta \in [0,1]$ (with Sigmoid activation) to combine the original direct specular color $\mathbf{c}_s$ and indirect specular color $\mathbf{c}_{ref}$ into new specular color $\mathbf{c}_s^{\prime}$:
\begin{equation}
    \mathbf{c}_s^\prime = \mathbf{c}_s + \eta \mathbf{c}_{ref}
\end{equation}

\section{Experiments}
We evaluate our method on various challenging shiny scenes and demonstrate the qualitative and quantitative results. We compare against prior methods based on view synthesis, scene relighting, and environment light estimation.

\noindent
\textbf{Datasets.} We use all 6 scenes in the Shiny Blender dataset proposed in \citea{verbin2022ref}, 2 shiny scenes (``ficus" and ``materials") from NeRF's Blender dataset \citea{mildenhall2021nerf}, and one real captured shiny scene (``garden spheres") from SNeRG \citea{hedman2021baking}.

\noindent
\textbf{Baselines.} We choose Ref-NeRF \citea{verbin2022ref} as the top-performing view synthesis model, NVDIFFREC \citea{munkberg2022extracting} and NVDIFFRECMC \citea{hasselgren2022shape} as two top-performing neural inverse rendering models. We also include VolSDF \citea{yariv2021volume} as a baseline neural surface model.

\subsection{Novel View Synthesis}
Following prior works, we use PSNR, SSIM, and LPIPS \citea{zhang2018unreasonable} to measure the view synthesis quality. Similar to \citea{verbin2022ref}, we use mean angular error (MAE) to evaluate the estimated surface normals.
We show the novel view synthesis results for all evaluated scenes in Table \ref{tab:nvs} and visual results in Figure \ref{fig:toaster_compare} and \ref{fig:garden}. 
Our model consistently shows better qualities in perceptually based metrics (SSIM and LPIPS).
\model\ significantly outperforms previous neural inverse rendering and neural surface methods. 
\model\ also performs on par with Ref-NeRF, and with higher PSNR scores in some scenes.
However, we should note that Ref-NeRF has a much lower surface quality (depicted by MAE) and does not support scene relighting.

\begin{table}[thb]
\centering
\footnotesize

\small
\footnotesize
\setlength\tabcolsep{0.5pt}
\begin{tabularx}{\linewidth}
    {{>{\raggedright\arraybackslash}l|}*{2}{>{\raggedleft\arraybackslash}X}|*{6}{>{\raggedleft\arraybackslash}X}|*{1}{>{\raggedleft\arraybackslash}X}}
\hline
\multicolumn{1}{c|}{} &\multicolumn{1}{c}{ficus} &\multicolumn{1}{c|}{mat.} &  \multicolumn{1}{c}{car} & 
\multicolumn{5}{c|}{ball\:\:\:helmet\:teapot\:toaster\:coffee} & \multicolumn{1}{c}{garden.} 

\\\hline 
\multicolumn{1}{c}{}&\multicolumn{8}{c}{\textbf{PSNR $\uparrow$}}                                                                  
\\ \hline
{VolSDF} & 22.91 & 29.13 & 27.41 & 33.66 & 28.97 & 44.73 & 24.10 & 31.22 & -\\ 
{NVDiffRec} & 29.88 & 26.89 & 27.98 & 21.77 & 26.97 & 40.44 & 24.31 & 30.74 & -\\ 
{NVDiffMC} & 27.05 & 25.68 & 25.93 & 30.85 & 26.27 & 38.44 & 22.18 & 29.60 & -\\ 
{ReF-NeRF} &  \cellcolor{orange!25}33.91 & \cellcolor{orange!25}35.41 & \cellcolor{orange!25}30.82 & \cellcolor{orange!25}47.46 & \cellcolor{yellow!25}29.68 & \cellcolor{orange!25}47.90 & \cellcolor{yellow!25}25.70 & \cellcolor{yellow!25}34.21 & \cellcolor{orange!25}23.46\\ 
{Ours} &  \cellcolor{yellow!25}30.53 & \cellcolor{yellow!25}29.51 & \cellcolor{yellow!25}29.88 & \cellcolor{yellow!25}41.03 & \cellcolor{orange!25}36.98 & \cellcolor{yellow!25}46.14 & \cellcolor{orange!25}26.63 & \cellcolor{orange!25}34.45 & 22.67\\ 
\hline
\multicolumn{1}{c}{}&\multicolumn{8}{c}{\textbf{SSIM $\uparrow$}}                                                                  
\\ \hline
{VolSDF} & 0.929 & 0.954 & 0.955 & 0.985 & \cellcolor{yellow!25}0.968 & \cellcolor{yellow!25}0.998 & \cellcolor{yellow!25}0.928 & \cellcolor{yellow!25}0.977 & -\\ 
{NVDiffRec} & \cellcolor{yellow!25}0.985 & 0.955 & \cellcolor{yellow!25}0.963 & 0.858 & 0.951 & 0.996 & \cellcolor{yellow!25}0.928 & 0.973 & -\\ 
{NVDiffMC} & 0.969 & 0.943 & 0.940 & 0.940 & 0.940 & 0.995 & 0.886 & 0.965 & -\\ 
{ReF-NeRF} & 0.983 & \cellcolor{orange!25}0.983 & 0.955 & \cellcolor{yellow!25}0.995 & 0.958 & \cellcolor{yellow!25}0.998 & 0.922 & 0.974 & 0.601\\ 
{Ours} & \cellcolor{orange!25}0.987 & \cellcolor{yellow!25}0.971 & \cellcolor{orange!25}0.972 & \cellcolor{orange!25}0.997 & \cellcolor{orange!25}0.993 & \cellcolor{orange!25}0.999 & \cellcolor{orange!25}0.955 & \cellcolor{orange!25}0.984 & \cellcolor{orange!25}0.695\\ 
\hline
\multicolumn{1}{c}{}&\multicolumn{8}{c}{\textbf{LPIPS $\downarrow$}}                                                                  
\\ \hline
{VolSDF} & 0.068 & 0.048 & 0.047 & \cellcolor{yellow!25}0.056 & \cellcolor{yellow!25}0.053 & \cellcolor{yellow!25}0.004 & 0.105 & \cellcolor{yellow!25}0.061 & -\\ 
{NVDiffRec} & \cellcolor{yellow!25}0.012 & 0.047 & 0.045 & 0.297 & 0.118 & 0.011 & 0.169 & 0.076 & -\\ 
{NVDiffMC} & 0.026 & 0.080 & 0.077 & 0.312 & 0.157 & 0.014 & 0.225 & 0.097 & -\\ 
{ReF-NeRF} & 0.019 & \cellcolor{orange!25}0.022 & \cellcolor{yellow!25}0.041 & 0.059 & 0.075 & \cellcolor{yellow!25}0.004 & \cellcolor{orange!25}0.095 & 0.078 & \cellcolor{orange!25}0.138\\ 
{Ours} & \cellcolor{orange!25}0.010 & \cellcolor{yellow!25}0.026 & \cellcolor{orange!25}0.031 & \cellcolor{orange!25}0.020 & \cellcolor{orange!25}0.022 & \cellcolor{orange!25}0.003 & \cellcolor{yellow!25}0.097 & \cellcolor{orange!25}0.044 & 0.372\\ 
\hline
\multicolumn{1}{c}{}&\multicolumn{8}{c}{\textbf{MAE\si{\degree} $\downarrow$}}                                                               \\ \hline
{VolSDF} & 39.80 & \cellcolor{orange!25}8.28 & \cellcolor{yellow!25}7.84 & \cellcolor{yellow!25}1.10 & \cellcolor{yellow!25}5.97 & \cellcolor{yellow!25}4.61 & \cellcolor{yellow!25}11.48 & \cellcolor{orange!25}7.68 & -\\ 
{NVDiffRec} & \cellcolor{yellow!25}32.39 & 15.42 & 11.78 & 32.67 & 21.19 & 5.55 & 16.04 & 15.05 & -\\ 
{NVDiffMC} & \cellcolor{orange!25}29.69 & 10.78 & 11.05 & 1.55 & 9.33 & 7.63 & 13.33 & 22.02 & -\\ 
{ReF-NeRF} & 41.05 & 9.53 & 14.93 & 1.55 & 29.48 & 9.23 & 42.87 & 12.24 & -\\ 
{Ours} & 34.44 & \cellcolor{yellow!25}8.47 & \cellcolor{orange!25}7.10 & \cellcolor{orange!25}0.74 & \cellcolor{orange!25}1.66 & \cellcolor{orange!25}2.47 & \cellcolor{orange!25}6.45 & \cellcolor{yellow!25}9.23 & -\\  \hline
\end{tabularx}
\vspace{-3pt}
\caption{\small Quantitative comparison among evaluated models. ``NVDiffMC" is short for NVDIFFRECMC. Ref-NeRF's results are imported from their original paper \citea{verbin2022ref}.}
\label{tab:nvs}
\vspace{-15pt}
\end{table}

In terms of learned surface quality,
\model\ achieves the lowest MAEs on almost all evaluated synthetic scenes, indicating superior surface quality. 
We attribute this improvement primarily to the VolSDF-like neural surface representation employed in our model, as VolSDF also demonstrates competitive MAE values.
Combining our neural renderer and neural surface model can further enhance the quality of the learned surface geometry.

\subsection{Environment Estimation}
Although the environment MLP in \model\ does not directly represent RGB values of environment light, it encodes the environment light as neural features. Our learned neural renderer can convert these neural features into RGB colors on a metallic sphere. By unwrapping such a sphere, we can obtain a panorama view of the environment light, which is similar to a light-probe image. 
This approach allows us to extract the environment light and compare it against the results obtained from other methods. In this section, we choose NVDIFFREC as a strong baseline to evaluate the accuracy of our environment light estimation.

 %Therefore, 
% We show the visual comparisons\footnote{Indeterminable color scaling and incomplete estimation from limited views make it difficult to quantify the quality of estimated light probes.} of environment light estimation in Figure \ref{fig:env_compare}.
Figure \ref{fig:env_compare} visually compares our estimated environment light maps with those of NVDIFFREC\footnote{Indeterminable color scaling and incomplete estimation from limited views make it difficult to quantify the quality of estimated light probes.}.
Figure \ref{fig:env_compare} demonstrates that our model effectively captures high-frequency environment lighting through the training with multiview images. 
Both our approach and NVDIFFREC accurately capture the high-quality environment light from highly reflective objects like the ``toaster" and ``helmet".
However, NVDIFFREC struggles to capture the detailed patterns of environment light for less reflective objects such as the ``coffee" and ``teapot", whereas our model still captures these patterns with precision. 
% As demonstrated in Figure \ref{fig:env_compare}, our model is able to capture high-frequency environment lighting through the training of the differentiable rendering. 
% Both our model and NVDIFFREC are able to capture high-quality environment light from more reflective objects such as toaster and helmet. 
% For less reflective objects such as coffee and teapot, our model is still able to capture detailed patterns of environment light, while the NVDIFFREC fails to capture these patterns in its estimations.
% It is worth noting that our primary focus in this work is not on environment light estimation.

% \textbf{Baseline:} NVDIFFREC

% \textbf{Scenes:} Helmet, Teapot, Coffee

% \begin{figure}
%     \centering
%     \includegraphics[width=\linewidth]{figures/env_compare.pdf}
%     \caption{Caption}
%     \label{fig:my_label}
% \end{figure}

\begin{table}[t]
    % \vspace{-10pt}
    \centering
    \small
    \setlength\tabcolsep{0pt}
    % \settowidth\rotheadsize{abcd}
    \begin{tabularx}{\linewidth}%
    {>{\centering\arraybackslash}p{0.15\linewidth}*{3}{>{\centering\arraybackslash}X}}
    \multicolumn{4}{c}{\includegraphics[width=\linewidth]{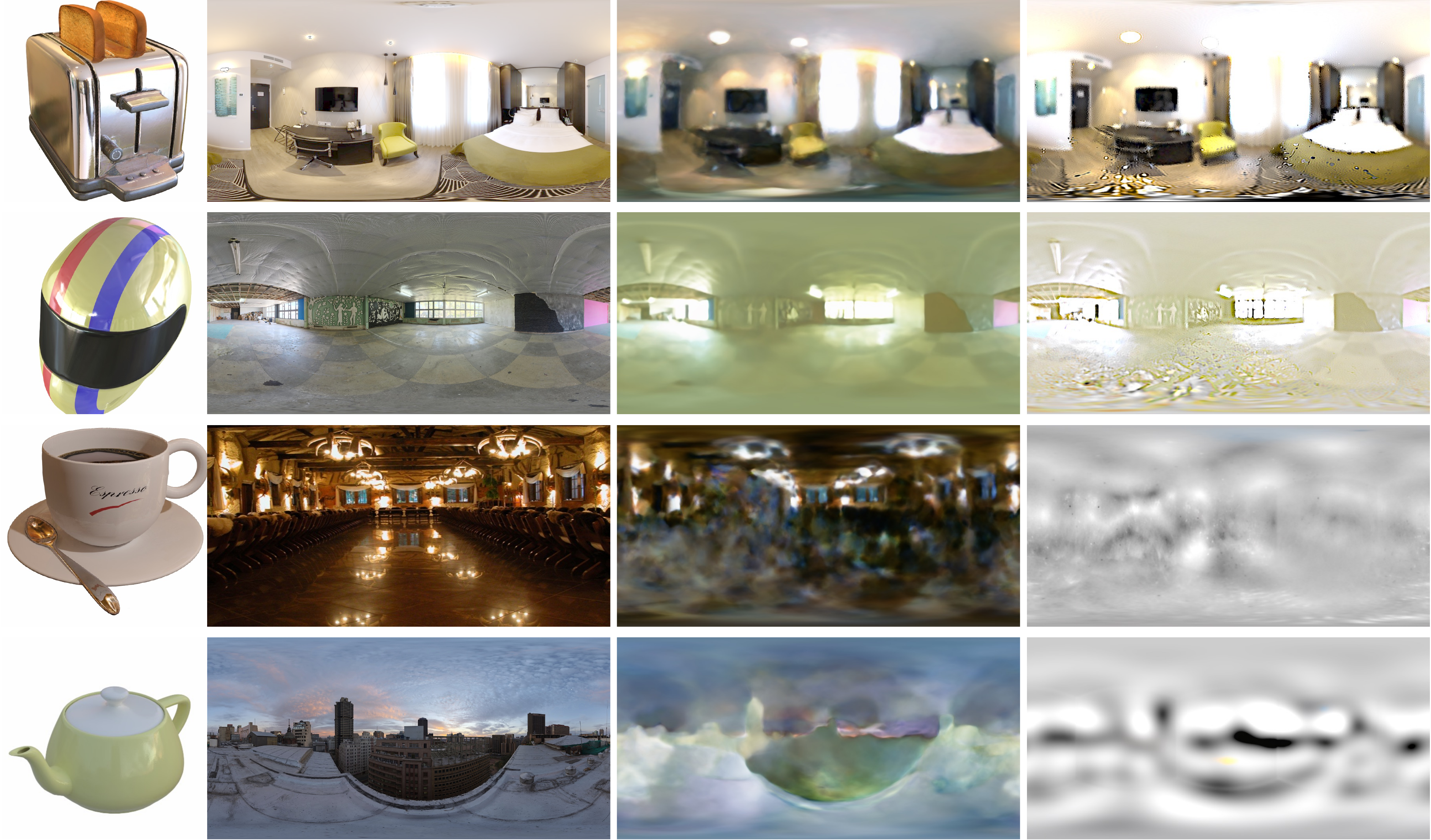}}%
    \\
    Scenes & Reference & Ours & NVDiffRec \citea{munkberg2022extracting}
    \end{tabularx}%
    \vspace{-2pt}
    \makeatletter\def\@captype{figure}\makeatother
    \caption{\small The comparison of estimated environment light probes. HDR light probes are converted to sRGB by gamma correction. 
    % We manually adjust the roughness of our Env-MLP for the best visual quality. The scene images shown on the left are synthesized by our model.
    Note that NVDIFFRECMC's extracted probes have similar or worse qualities compared to NVDIFFREC.
    }
    \label{fig:env_compare}
    % \vspace{-2pt}
\end{table}

\begin{table}[t]
    \centering
    \footnotesize
    \setlength\tabcolsep{0pt}
        \begin{tabularx}{\linewidth}%
        {p{1em}*{4}{>{\centering\arraybackslash}X}}
\rotatebox{90}{Ref.} & \multicolumn{4}{c}{\includegraphics[width=0.98\linewidth, trim={0 0 8cm 0},clip]{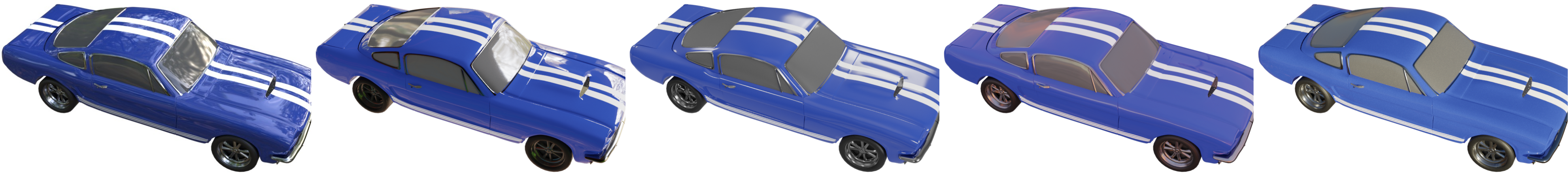}}\\
\rotatebox{90}{NV.Rec} & \multicolumn{4}{c}{\includegraphics[width=0.98\linewidth, trim={0 0 8cm 0},clip]{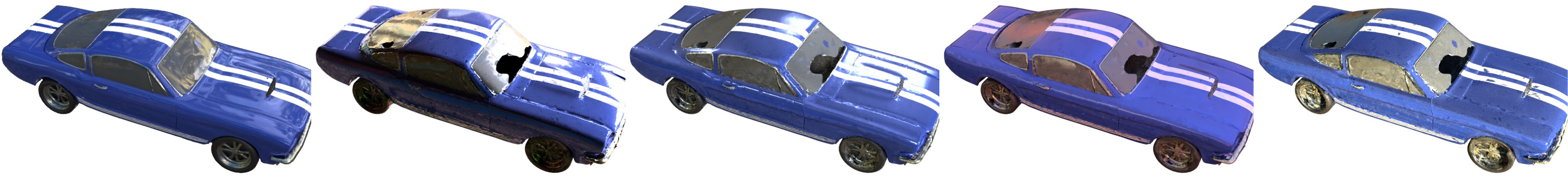}}\\
\rotatebox{90}{NV.MC} & \multicolumn{4}{c}{\includegraphics[width=0.98\linewidth, trim={0 0 8cm 0},clip]{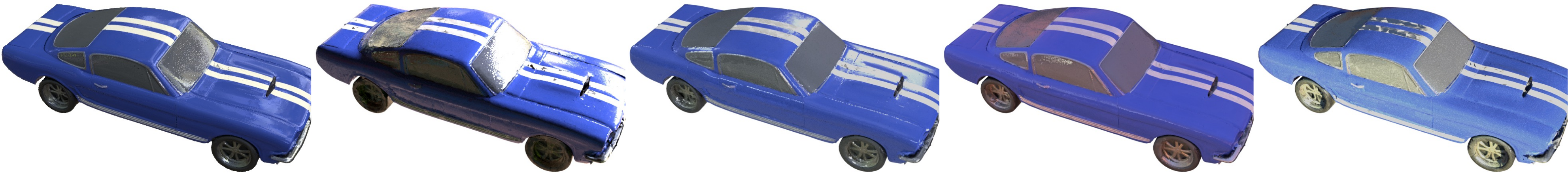}}\\
\rotatebox{90}{Ours} & \multicolumn{4}{c}{\includegraphics[width=0.98\linewidth, trim={0 0 8cm 0},clip]{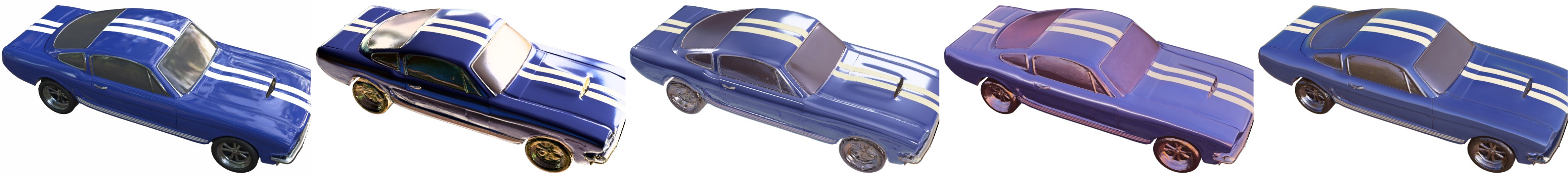}}\\
\rotatebox{90}{Env.} & \multicolumn{4}{c}{\includegraphics[width=0.98\linewidth, trim={-4cm 0 35cm 0},clip]{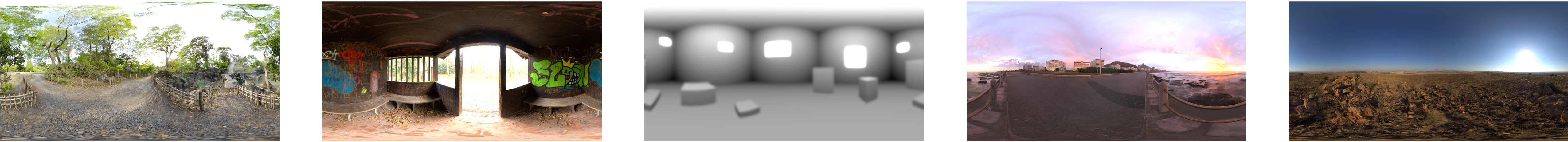}}\\
             & Original & Relighting \#1 & Relighting \#2 & Relighting \#3 
        \end{tabularx}%
        
    \makeatletter\def\@captype{figure}\makeatother
    \caption{\small
    Relighting results for ``car". 
    Except for ours, other results are rendered by Blender.
    % NVDiffRec and NVDiffRecMC use Blender to render relighting, while our results are rendered through our neural model.
    Since our relighting is synthesized by a neural model, the intensity of environment light represented by our env MLP may not match the Blender rendering.
    \label{fig:relighting}}
    \vspace{-10pt}
\end{table}

% \vspace{}
\subsection{Relighting}\label{sec:relight}
Similar to the relighting process in the traditional PBR, we can relight the scene represented by our model by replacing the environment MLP with environment MLPs representing new environment lights (these pre-trained env. MLPs can be obtained from our neural renderer).

In Figure \ref{fig:relighting}, we present a comparison between the scene relighting results obtained from our neural renderer and those obtained from NVDIFFREC and NVDIFFRECMC (rendered by Blender).
% we compare the scene relighting results obtained from our neural renderer with results from NVDIFFREC and NVDIFFRECMC rendered by Blender. 
Table \ref{tab:relighting} also provides a quantitative comparison. 
Our model outperforms NVDIFFREC since NVDIFFREC fails to accurately reconstruct surfaces on reflective regions (e.g., artifacts shown in Figure \ref{fig:relighting}).
Even though the baseline models directly use the same Blender rendering as the ground-truth reference, our model using a fully neural approach still provides comparable results with specular reflections on relit surfaces.
Artifacts from our approach are primarily due to: 1) the mismatch of rendering parameters (e.g., light intensity) used by our neural renderer and Blender; 
2) no synthesized shadowing effects on surfaces with occluded visibility due to the use of pre-integrated environment representation. We intend to address these limitations in our future work.
% Since our model uses the pre-integrated environment representation, which assumes full light visibility on the object surface, our neural renderer cannot synthesize the shadowing effects caused by surface occlusion. 

\begin{table}[t]
    \centering
    
% \small
% \footnotesize
% \setlength\tabcolsep{10pt}
% \begin{tabular}{c|c|c|c}
% \hline
% & car & ficus & materials \\
% \hline
% \multicolumn{1}{c}{}&\multicolumn{2}{c}{\textbf{PSNR $\uparrow$}}    
% \\\hline                        
% {NVDiffRec} & 22.44 & 20.21 & 22.07 \\ 
% {NVDiffMC}  & \textbf{24.71} & 23.16 & \textbf{23.69} \\ 
% {Ours}      & 22.45 & \textbf{24.64} & 22.48 \\
% \hline
% \multicolumn{1}{c}{}&\multicolumn{2}{c}{\textbf{SSIM $\uparrow$}}    
% \\\hline                        
% {NVDiffRec} & 0.920 & 0.888 & 0.895 \\ 
% {NVDiffMC}  & \textbf{0.936} & 0.921 & \textbf{0.914} \\ 
% {Ours}      & 0.921 & \textbf{0.936} & 0.900 \\
% \hline
% \end{tabular}

\small
\setlength\tabcolsep{6.5pt}
\begin{tabular}{l|c|c|c}
\hline
& car & ficus & materials \\
\hline
NVDiffRec & 22.44 / 0.920 & 20.21 / 0.888 & 22.07 / 0.895 \\
NVDiffMC  & \textbf{24.71} / \textbf{0.936} & 23.16 / 0.921 & \textbf{23.69} / \textbf{0.914} \\
Ours & 22.45 / 0.921 & \textbf{24.64} / \textbf{0.936} & 22.48 / 0.900\\
\hline
\end{tabular}
    \vspace{-6pt}
    \caption{Quantitative results (PSNR/SSIM) of relighting on three synthetic scenes with 4 light probe images, each with 50 uniformly sampled views}
    \label{tab:relighting}
\vspace{-13pt}
\end{table}

\vspace{-5pt}
\section{Ablation}\label{sec:ablation}

\subsection{The Design Choice of Neural Render}\label{sec:ablation_renderer}
Our proposed neural renderer is able to generalize to scenes with various shapes and materials for achieving reasonable relighting effects.
To justify the design choices, 
we conduct ablation studies from two aspects and present the visual comparisons in Figure \ref{fig:ficus_ablation}. 

\begin{table}[h]
    \centering
    \footnotesize
    \setlength\tabcolsep{1pt}
        \begin{tabularx}{\linewidth}%
        {*{4}{>{\centering\arraybackslash}X:}*{1}{>{\centering\arraybackslash}X}}
\includegraphics[width=\linewidth, trim={3cm 0 2.7cm 8cm},clip]{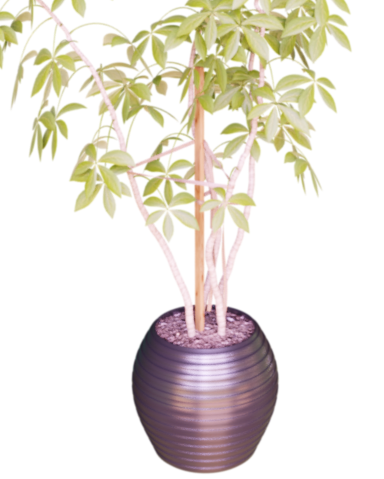} &
\includegraphics[width=\linewidth, trim={3cm 0 2.7cm 8cm},clip]{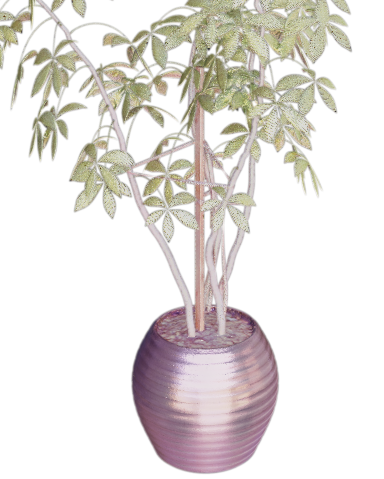} &
\includegraphics[width=\linewidth, trim={3cm 0 2.7cm 8cm},clip]{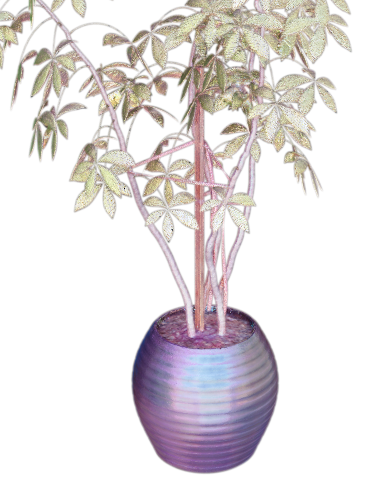} &
\includegraphics[width=\linewidth, trim={3cm 0 2.7cm 8cm},clip]{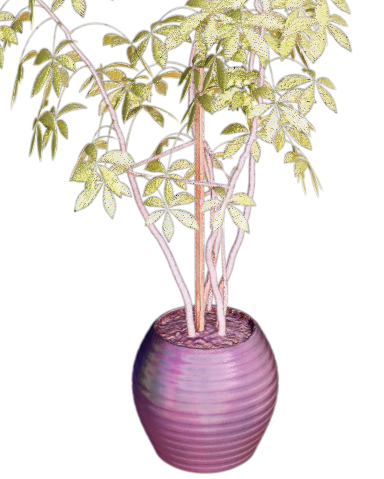} &
\includegraphics[width=\linewidth, trim={3cm 0 2.7cm 8cm},clip]{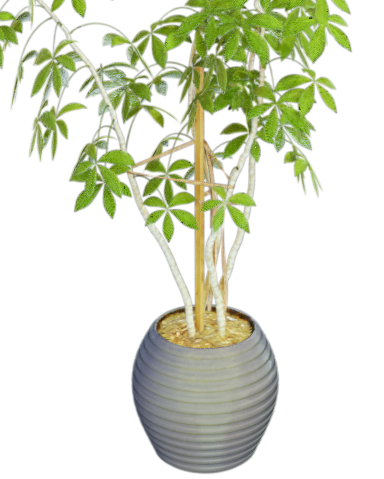}
\\
Reference & $l^2$-Norm & Inst-Norm &  w/o Norm & w/o Pre-train
\end{tabularx}%
        
    \makeatletter\def\@captype{figure}\makeatother
    \caption{The effects of design choices on scene relighting. 
    Note that models trained with different design options have the same level of rendering quality before relighting.
    \label{fig:ficus_ablation}}
\vspace{-6pt}
\end{table}

\noindent
\textbf{The use of pre-trained rendering MLPs.}
The use of pre-trained diffuse/specular MLPs allows the model to enforce a consistent feature-color mapping across different scenes. Without pre-training, the model cannot synthesize accurate reflections during relighting (``w/o Pre-train" in Fig. \ref{fig:ficus_ablation}).

\noindent
\textbf{Feature normalizations.} The feature normalization (Sec. \ref{sec:normalization}) is another non-trivial part of our design. 
To demonstrate the effectiveness of our $l^2$-Norm, we train two additional models: one without any normalization (``w/o Norm") and one with instance normalization (``Inst-Norm"). 
Compared to our $l^2$-Norm, both w/o Norm and Inst-Norm fail to synthesize the accurate specular highlight in the relit scene (e.g., pot in Fig. \ref{fig:ficus_ablation}).

\subsection{The Effect of Indirect Illuminations.}
\vspace{-3pt}
To demonstrate the effectiveness of our modeling of indirect illuminations, we train our model without specific modeling of indirect illumination (``w/o Indir.") on 4 scenes that contain obvious inter-reflections. Table \ref{tab:indir_vs_dir} and Figure \ref{fig:indir_vs_dir} provide the quantitative and qualitative comparisons, respectively.
The results demonstrate that our additional modeling of indirect illumination can help improve model's rendering quality, as well as the accuracy of surface geometry.

\vspace{-4pt}
\begin{table}[h]
    \centering
    \small
\setlength\tabcolsep{2pt}
\begin{tabular}{r|c|c|c|c}
\hline
& materials & toaster & coffee & garden. \\
\hline
w/o Indir. & 29.40 / 8.85 & 25.46 / 7.64 & 33.86 / 10.44 & 22.57 /\quad- \\
w/ Indir.  & \textbf{29.51} / \textbf{8.47} & \textbf{26.63} / \textbf{6.45} & \textbf{34.45} / \textbf{\;9.23} & \textbf{22.67} /\quad- \\
\hline
\end{tabular}

%%% can you replace with ssim with mae?
% sure

% \small
% \setlength\tabcolsep{1.3pt}
% \begin{tabular}{r|c|c|c|c}
% \hline
% & materials & toaster & coffee & garden. \\
% \hline
% w/o Indir. & 29.40 / 0.968 & 25.46 / 0.946 & 33.86 / 0.981 & 22.57 / 0.684   \\
% w/ Indir.  & \textbf{29.51} / \textbf{0.971} & \textbf{26.63} / \textbf{0.954} & \textbf{34.45} / \textbf{0.983} & \textbf{22.67} / \textbf{0.695}\\
% \hline
% \end{tabular}
    % \vspace{-4pt}
    \caption{
    Comparison of PSNR/MAE scores for models trained with and without indirect illumination modeling.}
    \label{tab:indir_vs_dir}
    \vspace{-15pt}
\end{table}

\begin{table}[h]
    \centering
    \footnotesize
    \setlength\tabcolsep{0pt}
        \begin{tabularx}{\linewidth}%
        {*{2}{>{\centering\arraybackslash}X:}*{2}{>{\centering\arraybackslash}X}}
Ground Truth & w/o Indir. & \multicolumn{2}{c}{w/ Indirect Illumination} \\
\includegraphics[width=\linewidth, trim={0 0cm 0 0cm},clip]{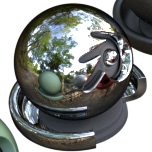} & 
\includegraphics[width=\linewidth, trim={0 0cm 0 0cm},clip]{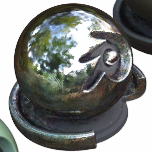} & 
\includegraphics[width=\linewidth, trim={0 0cm 0 0cm},clip]{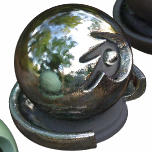} & 
\includegraphics[width=\linewidth, trim={0 0cm 0 0cm},clip]{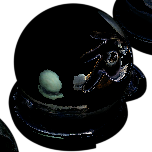} %\\
            % Ground Truth & W/o Indir. & W/ Indir. & Indir. Specular
        \end{tabularx}%
    \vspace{-5pt}
    \makeatletter\def\@captype{figure}\makeatother
    \caption{
    The visual comparison between models trained with and without indirect illumination modeling, the rightmost figure shows our synthesized indirect specular color.
    \label{fig:indir_vs_dir}}
\vspace{-20pt}
\end{table}

% \noindent
% \textbf{The color blending options.}
% \ruofan{may put in supp.}
\section{Limitations and Conclusions}
The main limitation of our neural renderer is the absence of explicit modeling of light visibility, which is crucial for synthesizing shadowing effects.
Our use of a pre-integrated representation of environment light, assuming full visibility of the lights to the surface, may result in lower rendering quality of shaded surfaces in complex objects (e.g., the base of material balls in Fig. \ref{fig:indir_vs_dir}).
This issue is also present in other neural inverse rendering methods \citea{zhang2021physg, boss2021neural, munkberg2022extracting}.
Given the high-quality surface geometry reconstructed by our model, our future work could incorporate geometry-based visibility approximations in proposed recent works \citea{hasselgren2022shape, liang2022spidr} to deal with shadowing effects.
Other limitations of our approach include the lack of fine-grained rendering parameter decomposition, the inability to handle semi-transparent or unbounded scenes, 
and limited support for indirect illuminations only on low-roughness surfaces. 

To summarize, we show that our approximation of physically based rendering with decomposed neural net components can help implicit neural surface models improve rendering and reconstruction of glossy surfaces, with results on par with or better than the state-of-art for view synthesis and inverse rendering, while enabling more accurate surface reconstruction and scene editing. 
The design of our neural renderer is inspired by PBR and models the implicit interaction between material and environment lighting. We use various plausible scene relighting and material editing examples in the paper to show the applicability of our approach.
 We believe our approach can benefit other implicit neural representation methods, leading to higher rendering and reconstruction quality with enhanced scene editability.

% In conclusion, our work demonstrates the effectiveness of using decomposed neural network components to approximate physically based rendering for improved rendering and reconstruction of glossy surfaces in implicit neural surface models. Our proposed neural renderer leverages the interaction between surface material and environment lighting, and achieves results on par with or better than previous state-of-the-art methods for view synthesis and inverse rendering. 
% We also showcase the versatility of our approach with various scene relighting and material editing examples. We believe that our method can be beneficial to other implicit neural scene representation methods, leading to improved rendering and reconstruction quality with enhanced scene editability.

% \begin{figure}
%     \centering
%     \begin{annotate}{\includegraphics[width=1\linewidth]{figures/toaster_blend/mipnerf.png}}{1}
%     \note{-0.5\linewidth,0}{Center}
%   \end{annotate}
%     \caption{Caption}
%     \label{fig:my_label}
% \end{figure}

% \clearpage
% {\small
% \bibliographystyle{ieee_fullname}
% \bibliography{egbib}
% }

% \newpage
% \ 
% \newpage

% xxx

% \newpage
% \ 
% \newpages

\clearpage
% \twocolumn[{
% \centering
% \renewcommand\twocolumn[1][]{#1}%
% \title{\emph{\model}: Implicit Differentiable Renderer with Neural Environment Lighting\\Supplement}
% \maketitle
% }]

\section*{Supplementary}
\appendix

\section{Additional SDF Regularizations}
% Learning implicit neural surface representations is more challenging compared to learning neural volume representations. 
In addition to the Eikonal term \citep{gropp2020implicit} that is commonly used for constraining the learned SDF field,
neural surface methods also require special weight initialization or warm-up steps to stabilize the learning of implicit surfaces \citep{yariv2021volume, wang2021neus, Yu2022MonoSDF}.
However, unlike prior works that use a large fully implicit MLP \citep{yariv2021volume, wang2021neus} or attach spatial coordinates to the grid-interpolated geometry features \citep{Yu2022MonoSDF}, \model's geometry MLP uses an NGP-like \citep{mueller2022instant} model that uses multi-level hash encoding as the only input to the tiny geometry MLP. Therefore, the geometric initialization \citep{atzmon2020sal} that assumes spatial coordinates as MLP inputs cannot be applied to our model. Thus, our model requires new ways to initialize its learned geometry to better represent continuous and smooth surfaces. 

\begin{figure}[ht]
    \centering
    \includegraphics[width=\linewidth, trim={0.5cm, 0.1cm, 0.5cm, 0cm}, clip]{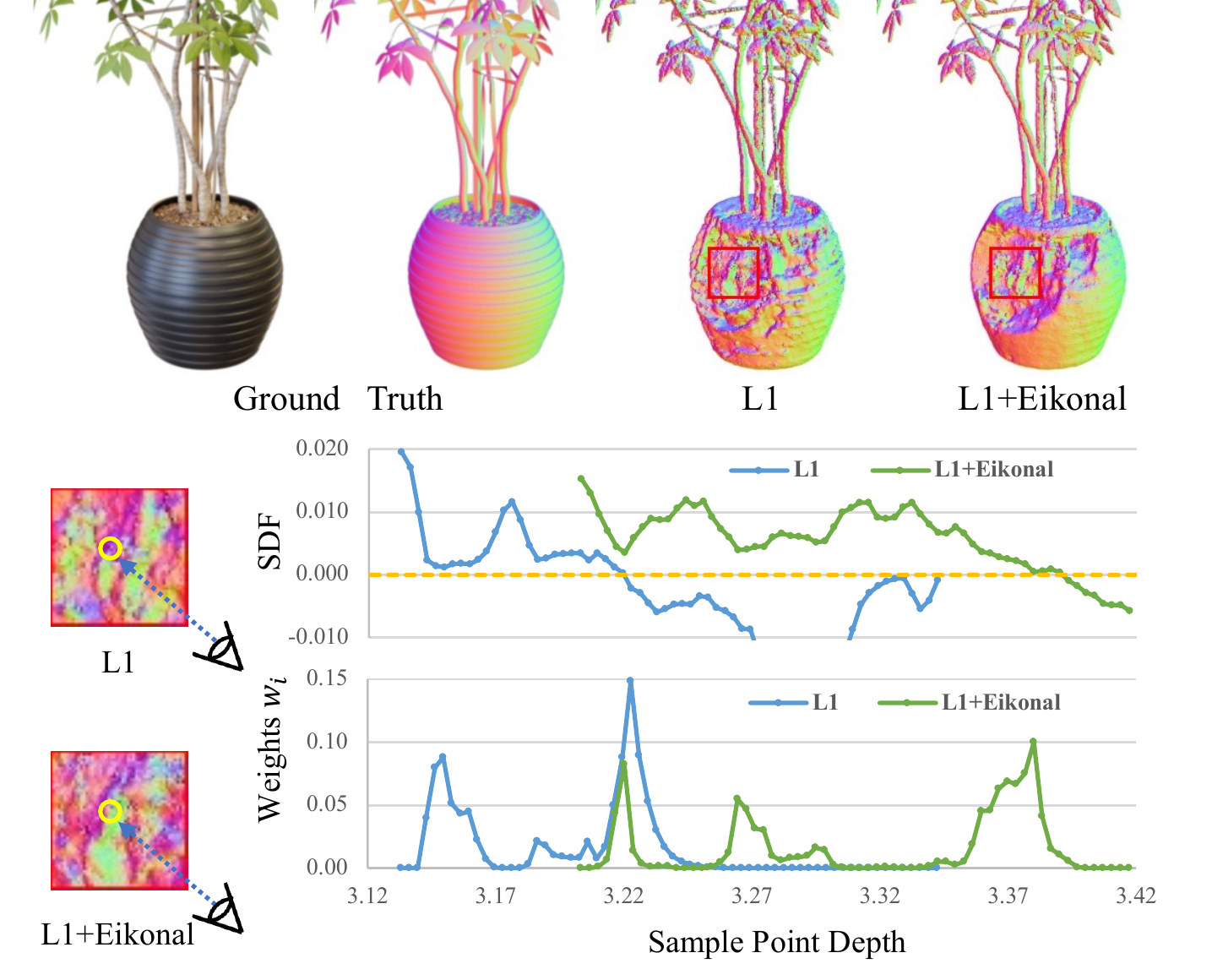}
    \vspace{-10pt}
    \caption{The normals of the surfaces learned by models without additional constraints. The results are from models after only 20k training iterations.
    We also plot the curves of SDF and compositing weight $w_i$ of the sampled points along the ray for rendering the pixel shown on the left.}
    \label{fig:ficus_curve}
    \vspace{-12pt}
\end{figure}

To demonstrate the importance of incorporating special constraints when learning implicit neural surfaces, we first show the results of models that do not include additional SDF constraints. Figure \ref{fig:ficus_curve} shows the surface normals of two such models, one trained solely with L1 photometric loss (``L1") and the other with L1 loss and Eikonal term (``L1+Eikonal"). 
Both of these models fail to capture accurate surface geometry on glossy regions, and their corresponding SDF curves oscillate around the zero-level SDF, which results in a compositing weight $w_i$ (obtained from Eqn. \ref{eq_radiance}) distribution with multiple peaks along the ray. 
These scattered compositing weights further cause the reconstructed surface to ``collapse" into the actual object.
To avoid this surface collapse, additional regularizations are needed on the initial SDF predictions.
% we can add extra regularizations to reduce the SDF oscillation and thus avoid surface collapse.

\begin{figure}
\centering
\small
\begin{subfigure}[c]{0.243\linewidth}
\includegraphics[width=\linewidth, trim={3cm 0 2.7cm 6cm},clip]{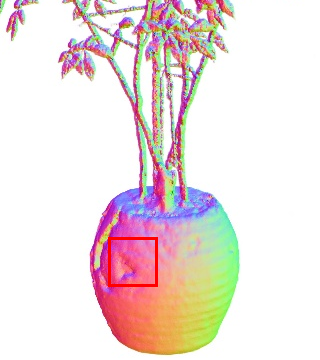}
\caption{$\mathcal{L}_n$ Only}\label{fig:ln_only}
\end{subfigure}
\begin{subfigure}[c]{0.243\linewidth}
\includegraphics[width=\linewidth, trim={3cm 0 2.7cm 6cm},clip]{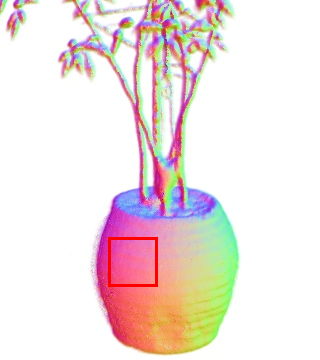}
\caption{$\mathcal{L}_b$ Only}\label{fig:lb_only}
\end{subfigure}
\begin{subfigure}[c]{0.243\linewidth}
\includegraphics[width=\linewidth, trim={3cm 0 2.7cm 6cm},clip]{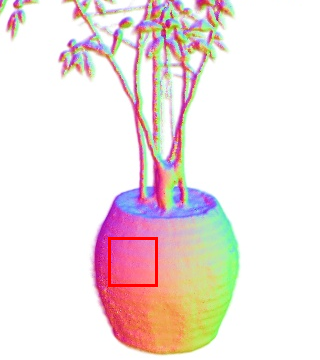}
\caption{$\mathcal{L}_n + \mathcal{L}_b$}\label{fig:ln_lb}
\end{subfigure}
\begin{subfigure}[c]{0.243\linewidth}
\includegraphics[width=\linewidth, trim={3cm 0 2.7cm 6cm},clip]{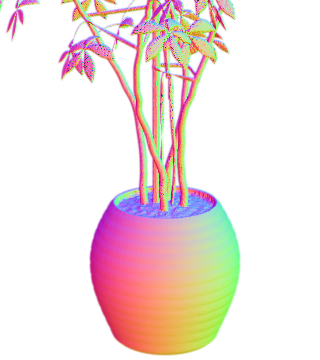}\label{fig:our_200k}
\caption{200k iter.}
\end{subfigure}
% $\mathcal{L}_n$ Only & $l^2$-Norm & Inst-Norm &  full, 200k iter.
        
    \caption{The estimated surface normal with our regularization terms. (a-c) show the results after 20k iterations, and (d) shows the results of the model with our full regularizations after 200k training iterations ($\mathcal{L}_n$ \& $\mathcal{L}_b$ stop at 40k).
    \label{fig:ficus_w_reg}}
\vspace{-6pt}
\end{figure}

In this work, we employ two SDF regularization terms in the early training steps to stabilize the learning of the initial SDF. 
The first regularization term is to force the SDF predictions of positions inside object surfaces to be away from zero-level SDF, which can suppress the multiple peaks on the compositing weight curves. 
To avoid over-suppression of the existing SDF predictions that are far away from the zero-level SDF, we employ a modified Cauchy loss \citep{hedman2021baking} on the SDF-converted (via Eqn. \ref{eq:conversion}) density values $\sigma_i$:
\begin{equation}
    \mathcal{L}_n = \frac{1}{N}\sum_{i} \log\left(1+ \frac{(1-\beta \sigma_i^2}{c^2}\right)
\label{eq:l_s}
\end{equation}
where $\beta$ is the parameter used in Eqn. \ref{eq:conversion} and $c$ is a hyperparameter that controls the loss scale which we set to 4. This regularization term is uniformly applied to all the sampled $N$ points per iteration. The effectiveness of this regularization is shown in Figure \ref{fig:ln_only}. Compared to the results in Figure \ref{fig:ficus_curve}, $\mathcal{L}_n$ is able to help our model to get a better initial surface structure.

The second regularization is to eliminate the fluctuations of SDF curve segments that are close to zero-level SDF. The zero level set of SDF denotes the actual position of the surface, and frequent fluctuations near the surface can significantly impact the quality of surface geometry as well as surface normals. 
To ensure that our estimated SDF has a stable decreasing curve when it hits the surface at ray-sampled points, we propose a back-face suppression regularization. This regularization penalizes SDF curve segments with positive slopes and high corresponding compositing weights, as shown in the following equation:
\begin{equation}
    \mathcal{L}_b = \sum_i w_i\max(\Delta s_i, 0)\frac{\Delta s_i}{\delta_i^2 + \Delta s_i^2}
\end{equation}
In this equation, $\Delta s_i = s({\mathbf{x}_{i+1}}) - s({\mathbf{x}_{i}})$ represents the difference in estimated SDF between two adjacent sampled points ${\mathbf{x}_{i}}$ and ${\mathbf{x}_{i+1}}$, $\delta_i$ is the actual distance between ${\mathbf{x}_{i}}$ and ${\mathbf{x}_{i+1}}$. 
The compositing weight $w_i$ for volume rendering can be obtained from Equation \ref{eq_radiance}.
Figure \ref{fig:lb_only} demonstrates the effectiveness of this regularization term. We can observe that the smoother estimated surface is achieved with the same number of training iterations compared to the results obtained using only $\mathcal{L}_n$. However, it should be noted that the object's surface shrinks slightly compared to the other results.

By combining the two regularization terms introduced above, we are able to achieve a more stable and accurate SDF estimation. Specifically, we formulate the combination of the two regularization terms as:
\begin{equation}
\mathcal{L}_{reg} =\lambda_n \mathcal{L}_n + \lambda_b \mathcal{L}_b
\end{equation}
where $\lambda_n$ and $\lambda_b$ are weight hyperparameters.
Setting overly large values of $\lambda_n$ and $\lambda_b$ may prohibit the model from learning fine-grained geometry details. However, they can be beneficial in providing smooth surfaces for the later training steps.
The effectiveness of the combined regularization can be seen in Figure \ref{fig:ln_lb}. By using these additional SDF regularizations during the early training steps, we can produce much better initial surfaces for accurate reconstruction and illumination.

% \subsection{Relighting}

\begin{table}[t]
    \centering
    \footnotesize
    \setlength\tabcolsep{0pt}
        \begin{tabularx}{\linewidth}%
        {p{1em}*{4}{>{\centering\arraybackslash}X}}
\hdashline
\rotatebox[origin=l]{90}{NVDiffRec} & \multicolumn{4}{c}{\includegraphics[width=0.98\linewidth, trim={0 0 0 0},clip]{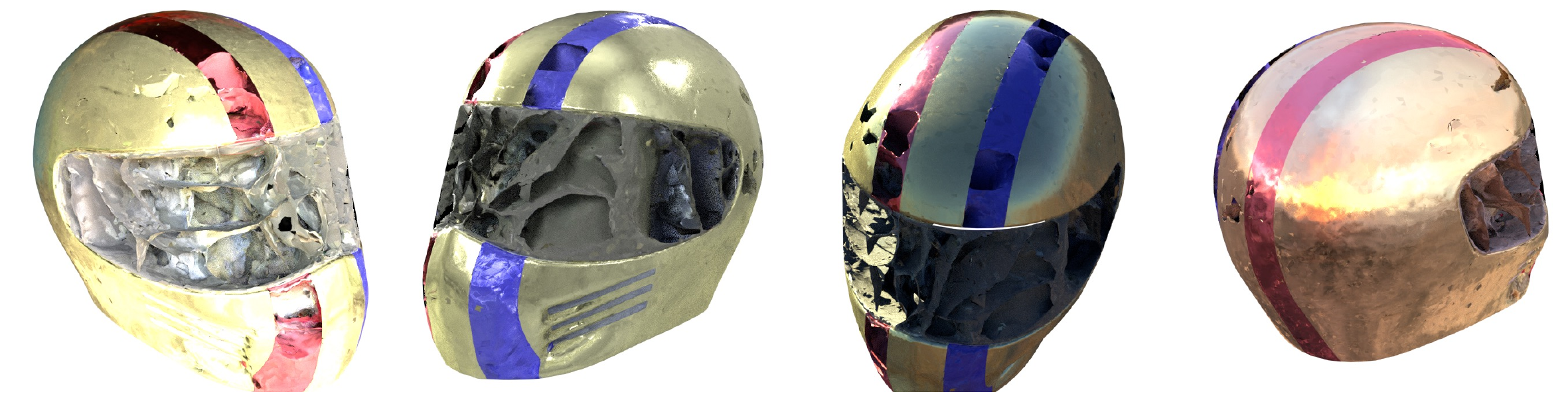}}\\
\rotatebox[origin=l]{90}{NVDiffMC} & \multicolumn{4}{c}{\includegraphics[width=0.98\linewidth, trim={0 0 0 0},clip]{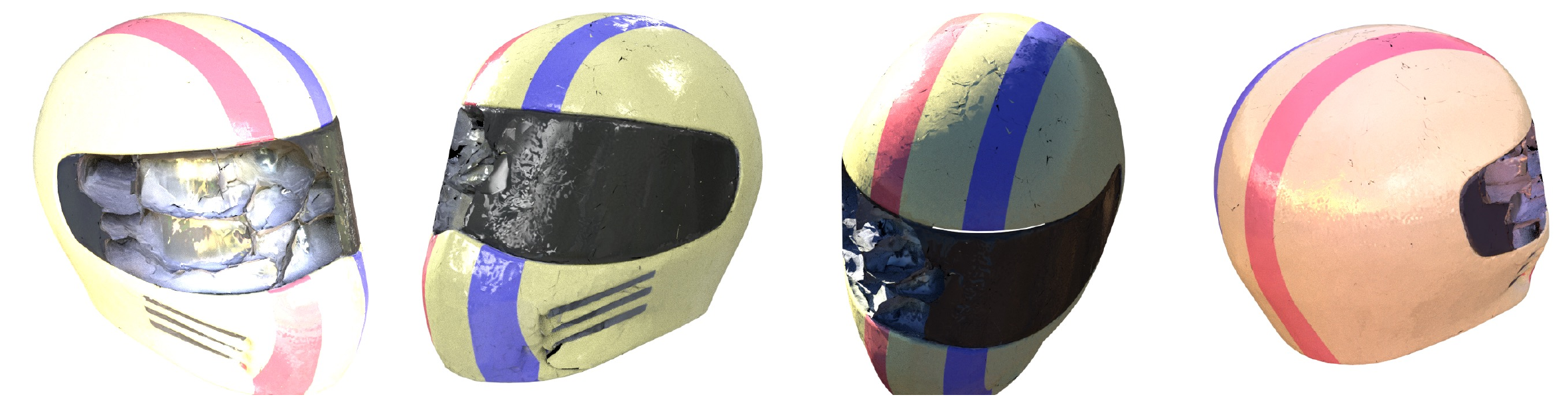}}\\
\rotatebox[origin=l]{90}{\quad\quad Ours} & \multicolumn{4}{c}{\includegraphics[width=0.98\linewidth, trim={0 0 0 0},clip]{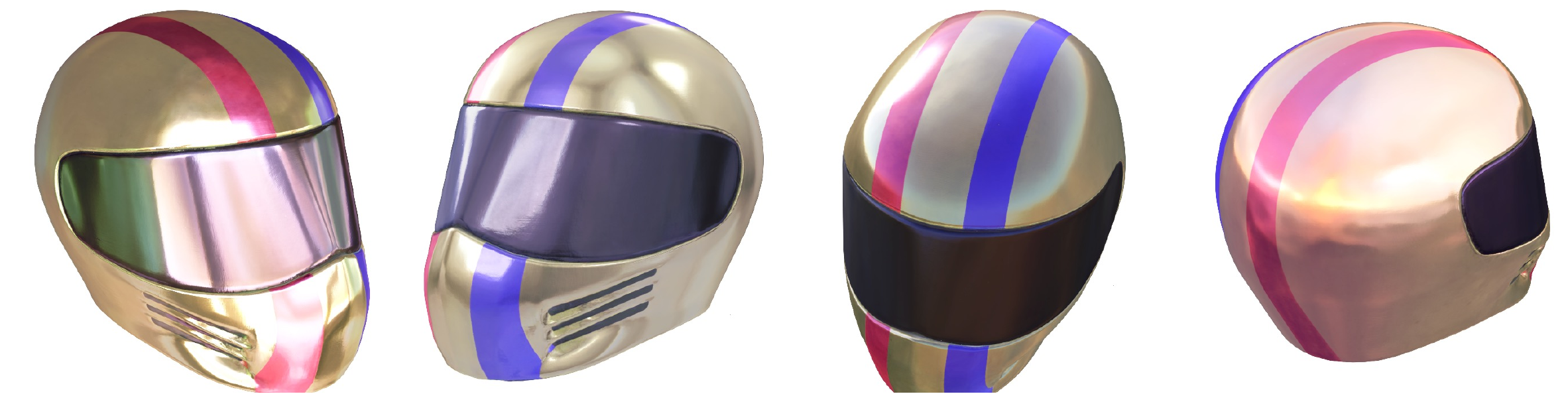}}\\
\hdashline
\rotatebox[origin=l]{90}{NVDiffRec} & \multicolumn{4}{c}{\includegraphics[width=0.98\linewidth, trim={0 0 0 0},clip]{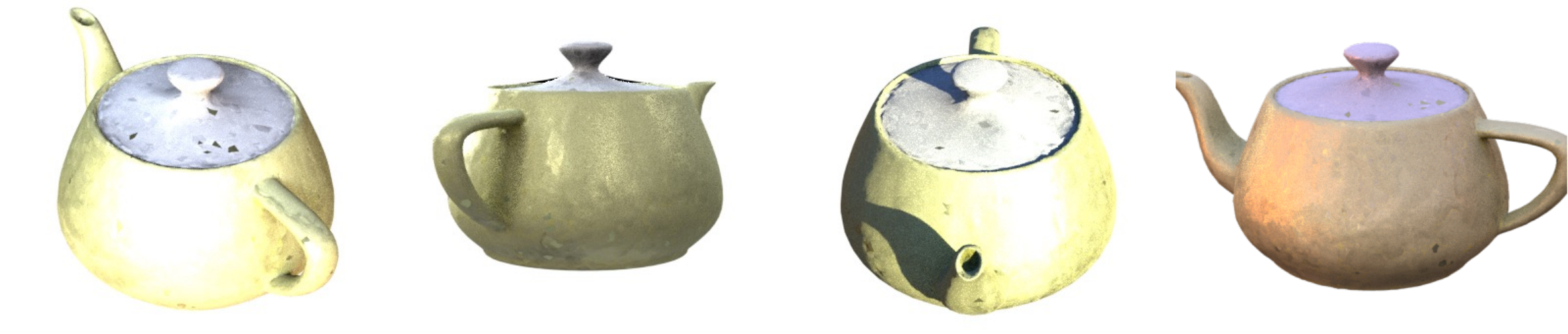}}\\
\rotatebox[origin=l]{90}{NVDiffMC} & \multicolumn{4}{c}{\includegraphics[width=0.98\linewidth, trim={0 0 0 0},clip]{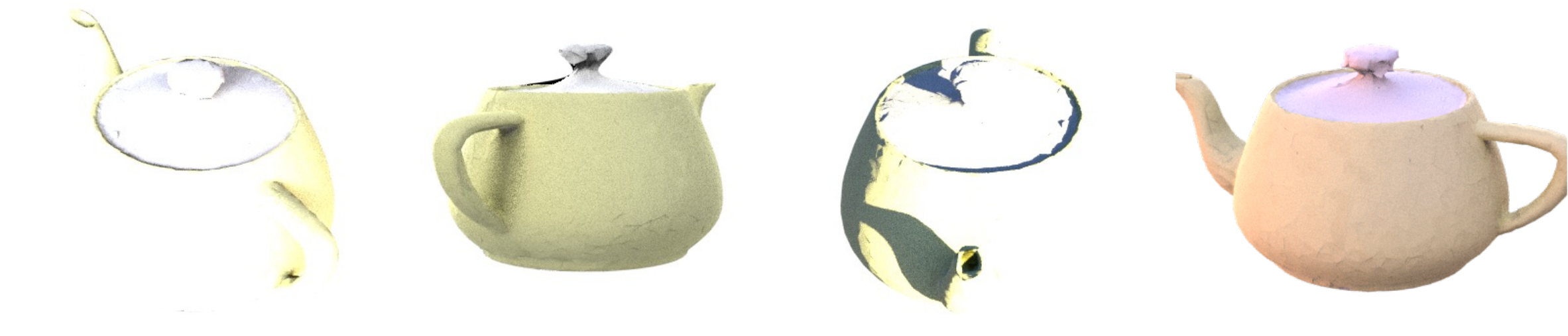}}\\
\rotatebox[origin=l]{90}{\quad\quad Ours} & \multicolumn{4}{c}{\includegraphics[width=0.98\linewidth, trim={0 0 0 0},clip]{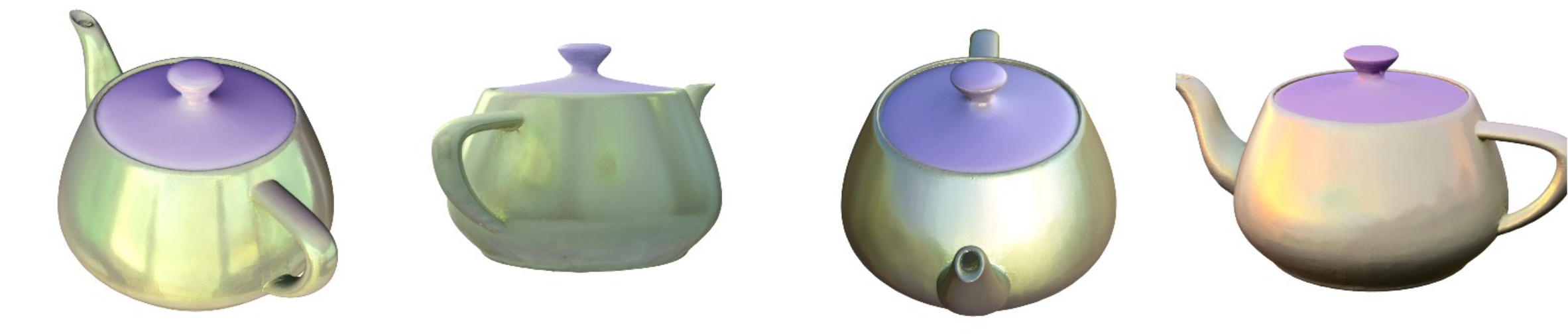}}\\
\hdashline
\rotatebox[origin=l]{90}{NVDiffRec} & \multicolumn{4}{c}{\includegraphics[width=0.98\linewidth, trim={0 0 0 0},clip]{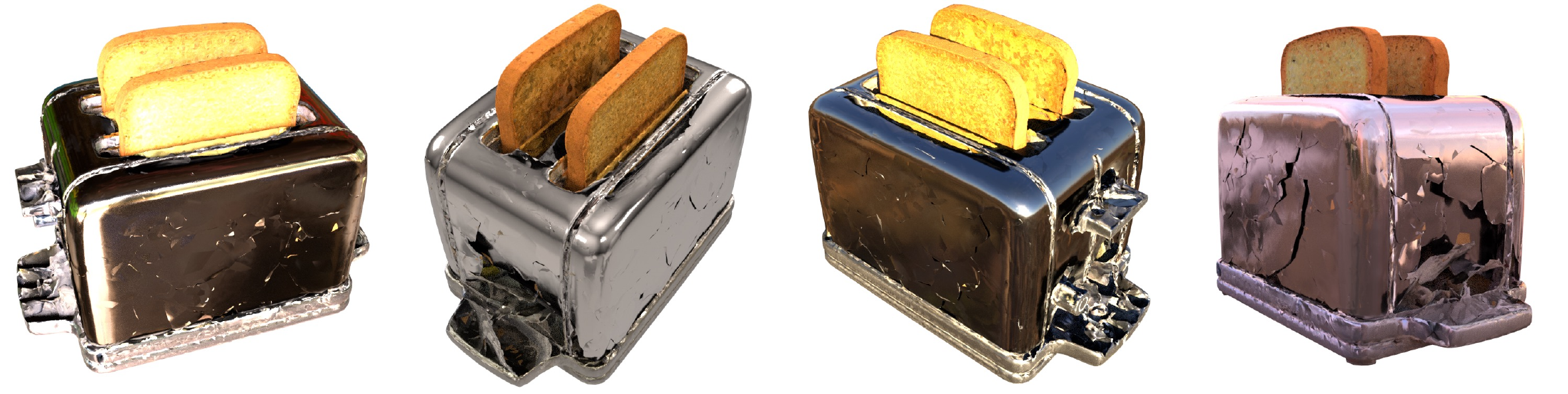}}\\
\rotatebox[origin=l]{90}{NVDiffMC} & \multicolumn{4}{c}{\includegraphics[width=0.98\linewidth, trim={0 0 0 0},clip]{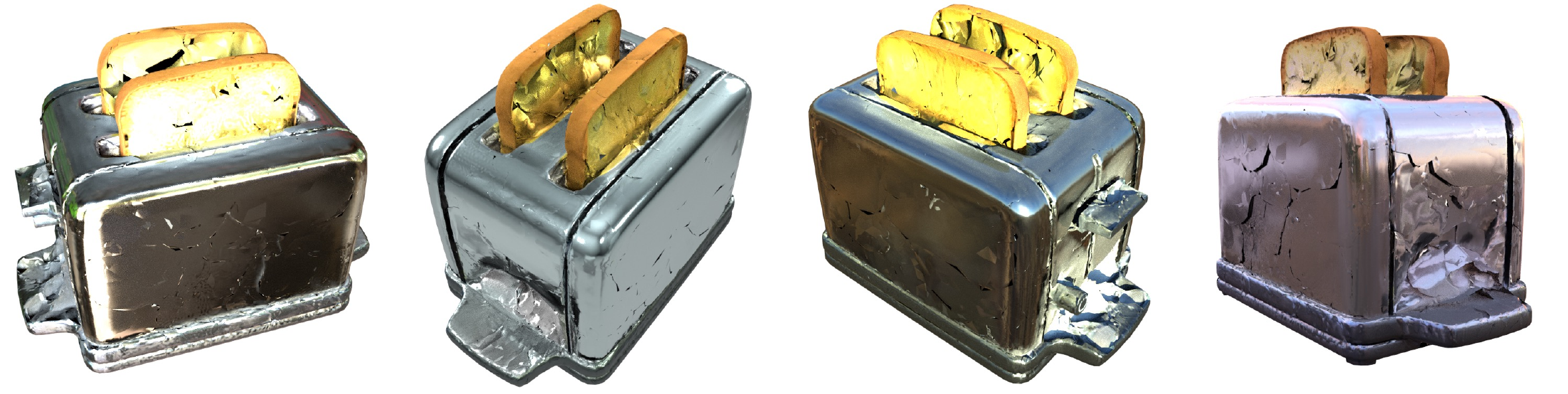}}\\
\rotatebox[origin=l]{90}{\quad\quad Ours} & \multicolumn{4}{c}{\includegraphics[width=0.98\linewidth, trim={0 0 0 0},clip]{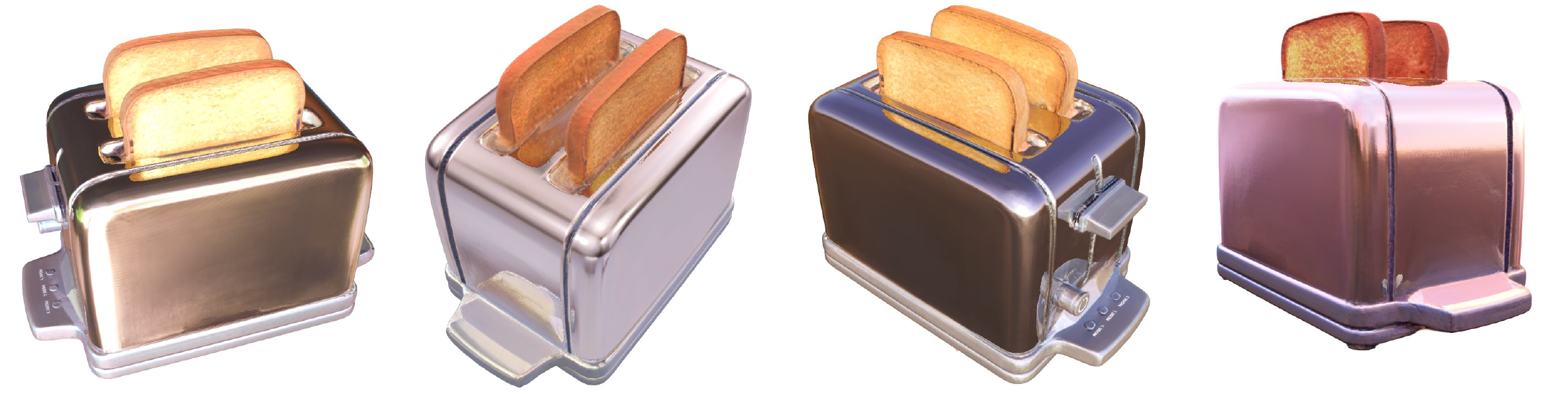}}\\
\hdashline
\rotatebox{90}{Probes} & \multicolumn{4}{c}{\includegraphics[width=0.98\linewidth, trim={0 0 0 0},clip]{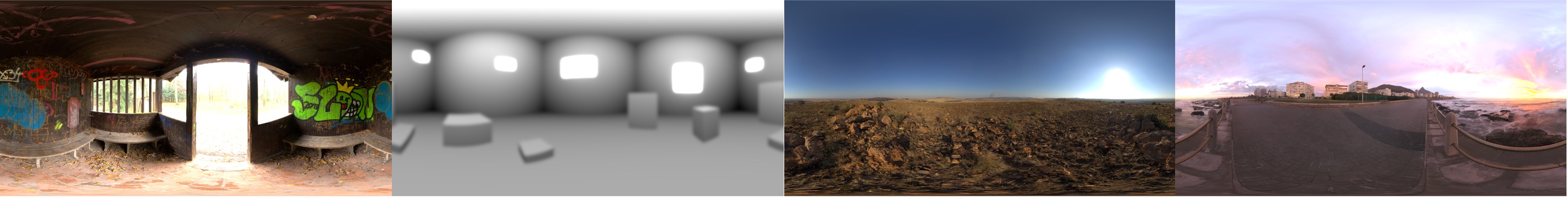}}\\
             & Relighting \#1 & Relighting \#2 & Relighting \#3 & Relighting \#4 
        \end{tabularx}%
        
    \makeatletter\def\@captype{figure}\makeatother
    \caption{\small
Additional relighting results for various scenes. Due to the unavailability of Blender files for these objects, reference relighting images are not provided for comparison. 
NVDIFFREC and NVDIFFRECMC are rendered by Blender Cycles. 
    \label{fig:add_relighting}}
    \vspace{-10pt}
\end{table}

\section{Implementation Details}
\textbf{Architecture and hyperparameters.} 
The geometry MLP $\bm{F}_g$ used in \model\ is similar to Instant-NGP \cite{mueller2022instant}. We employ a hash encoding with 16 grid levels where each level encodes a 2-d feature vector. The MLP $\bm{F}_g$ itself is a tiny 3-layer MLP with 64 neurons per hidden layer.
The output geometry feature $\mathbf{f}_{geo}$ is a 12-d feature vector. 
The environment MLP $\bm{E}$ is a relatively large  MLP with 4 layers and 256 neurons per layer (still smaller than a 1k HDR light probe image). The integrated directional encoding (IDE) \cite{verbin2022ref} utilized by $\bm{E}$ encodes unit directions using the first 5 bands of spherical harmonics. The output environment feature $\mathbf{f}_{env}$ is also a 12-d feature vector. Specular MLP $\bm{R}_s$ is a 3-layer MLP with 64 neurons per hidden layer, while Diffuse MLP $\bm{R}_d$ is a 2-layer MLP with 32 neurons per hidden layer. We implemented our model using a PyTorch version of Instant-NGP \footnote{\url{https://github.com/ashawkey/torch-ngp}}. 

\textbf{Training details.}
For the neural renderer's training, we use Filament PBR engine \cite{google2018filament} to randomly synthesize new frames on the fly, with varying materials and environment lights (we use 11 light probes images, provided by Filament's repository\footnote{\url{https://github.com/google/filament/tree/main/third\_party/environments}}). 
We train the neural renderer for 100k iterations, each with 32000 sampled rays, using the Adam optimizer with an initial learning rate of 0.001. This training takes about 3 hours on a single RTX3090 GPU.
For the training of representing general scenes, we train our model for 200k iterations with 4096 sampled rays per iteration, using the Adam optimizer with an initial learning rate of 0.0005. 
We first train the model only with a photometric loss for 4k iterations to obtain a coarse geometry for ray-sampling acceleration. We then apply our additional SDF regularizations for about 40k iterations with exponentially decaying loss weights. The Eikonal loss term is added to the training after the first 10k training iterations. If the indirect illumination module is used, we initiate the extra raymarching pass after the first 40k training iterations.
The training speed depends on the complexity of the target scene, but most scenes can be trained within 3 hours (5 hours if indirect illumination is used) on a single RTX3090 GPU.

\textbf{Runtime.}
The time required to render a single 800$\times$800 image is approximately between 0.5 to 1.2 seconds (without indirect illumination) on a single RTX3090 GPU. If the indirect illumination pass is enabled, rendering may take 1.6 times longer. Although our code base is not yet optimized for runtime performance, further optimizations are possible to achieve faster rendering.

\section{Additional Results}
In this section, we present additional visual results to demonstrate \model's ability to reconstruct and render glossy surfaces. We also provide a demo video on our \href{https://nexuslrf.github.io/ENVIDR/}{web page} to showcase results in motion.

\textbf{Relighting}. Additional relighting results on challenging synthetic shiny scenes are demonstrated in Figure \ref{fig:add_relighting}. \model\ is capable of synthesizing lighting effects that are comparable to Blender's path-tracing rendering.  while maintaining significantly better surface geometry compared to the two baseline models. However, these relighting results also reveal some limitations of our method. For instance, in the ``teapot" example, our current method cannot synthesize shadowing effects caused by surface occlusions. We plan to address this in our future work.

\textbf{Scene Decompositions}
The rendering decomposition of all evaluated synthetic scenes is shown in Figure \ref{fig:perscene}. \model~can effectively decompose both the view-independent diffuse color and view-dependent specular color from multiview training images. Moreover, it can successfully distinguish between different types of materials present in the scene, such as metallic materials in the "toaster" and "coffee" scenes. In addition, our model also captures high-fidelity environment light probes from these shiny objects. 

% \subsection{Scene Decompositions}

\begin{table*}[ht]
\vspace{-5pt}
    \centering
    \footnotesize
    \setlength\tabcolsep{0.5pt}
\begin{tabularx}{\linewidth}%
{p{1em}*{6}{>{\centering\arraybackslash}X}}
\hdashline
%%%%%%%%%%%%%%%%%%%%%%
{\rotatebox{90}{ficus}} & 
\multicolumn{6}{c}
{
\begin{subfigure}[c]{0.166\linewidth}%
\includegraphics[width=0.94\linewidth]{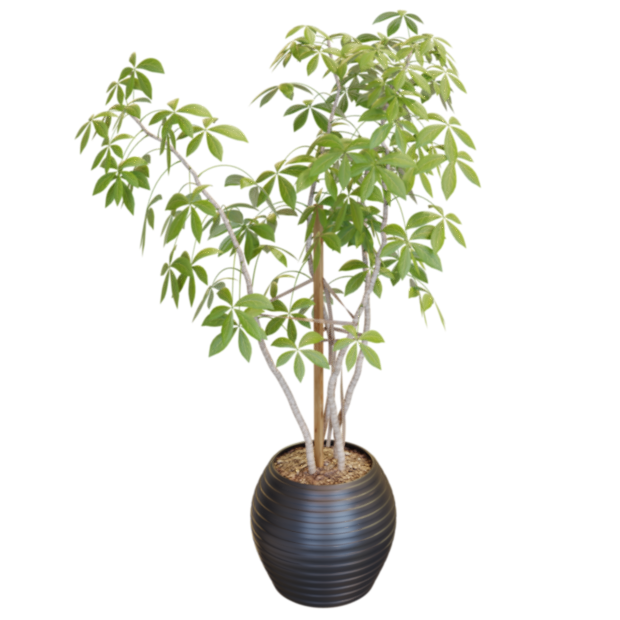}%
\end{subfigure}%
\begin{subfigure}[c]{0.166\linewidth}%
\includegraphics[width=0.94\linewidth]{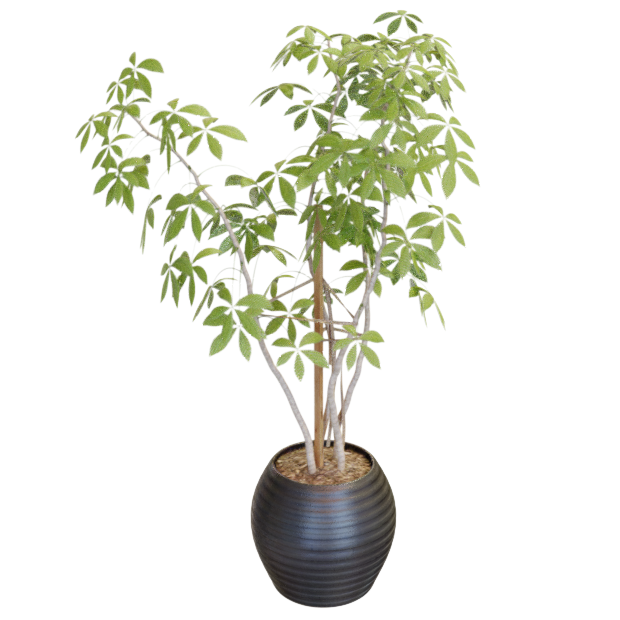}%
\end{subfigure}%
\begin{subfigure}[c]{0.166\linewidth}%
\includegraphics[width=0.94\linewidth]{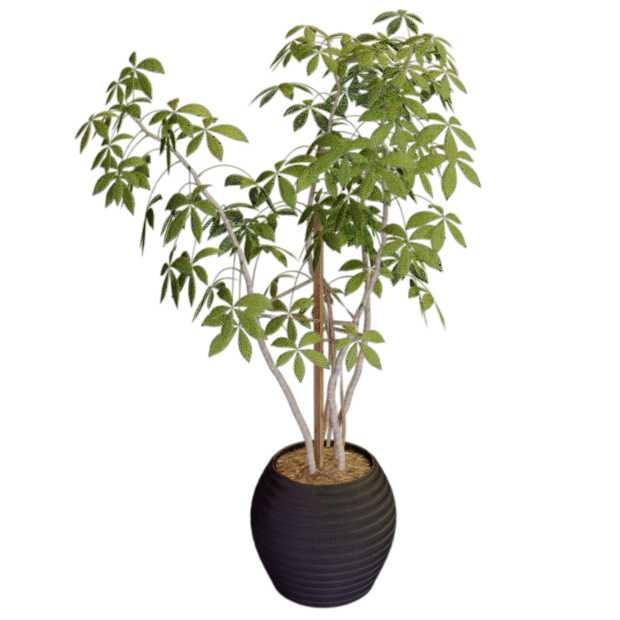}%
\end{subfigure}%
\begin{subfigure}[c]{0.166\linewidth}%
\includegraphics[width=0.94\linewidth]{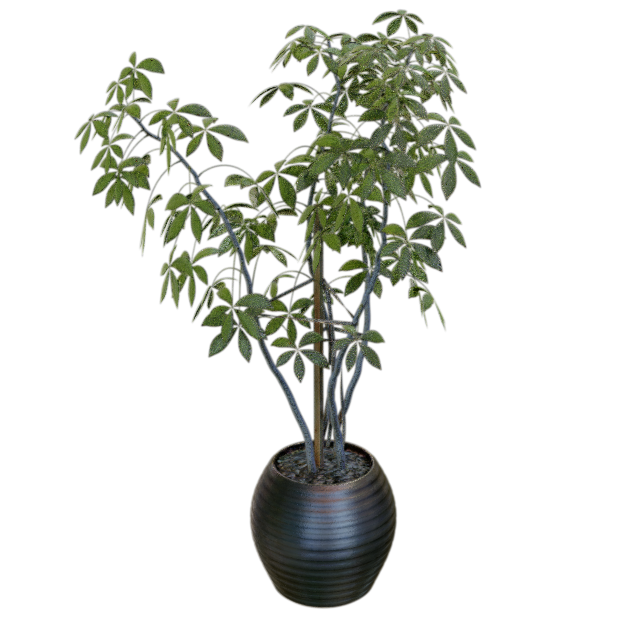}%
\end{subfigure}%
\begin{subfigure}[c]{0.166\linewidth}%
\includegraphics[width=0.94\linewidth]{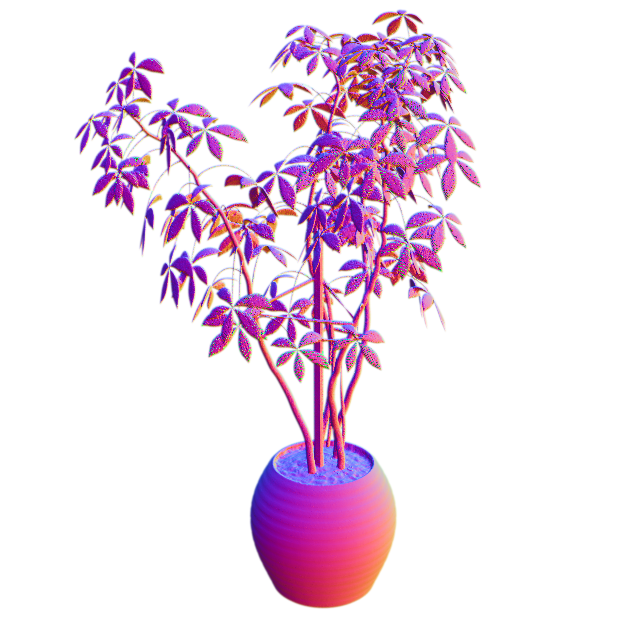}%
\end{subfigure}%
\begin{subfigure}[c]{0.166\linewidth}%
\begin{subfigure}[t!]{0.9\linewidth}%
\includegraphics[width=\linewidth]{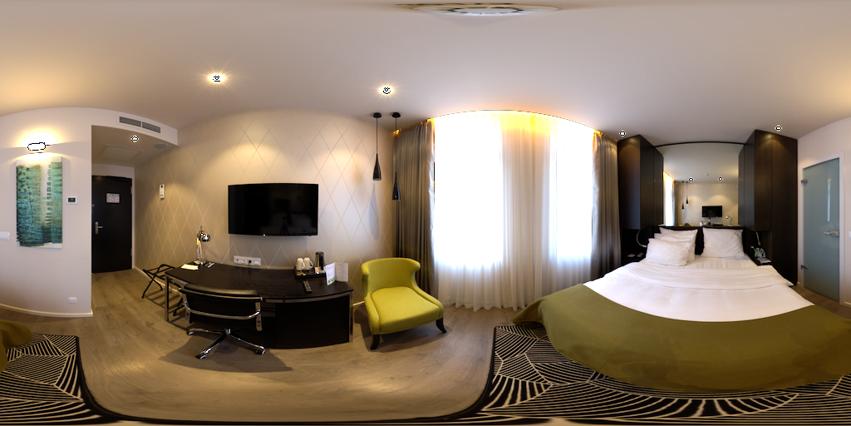}%
\end{subfigure}%
\\
\begin{subfigure}[b!]{0.9\linewidth}%
\includegraphics[width=\linewidth]{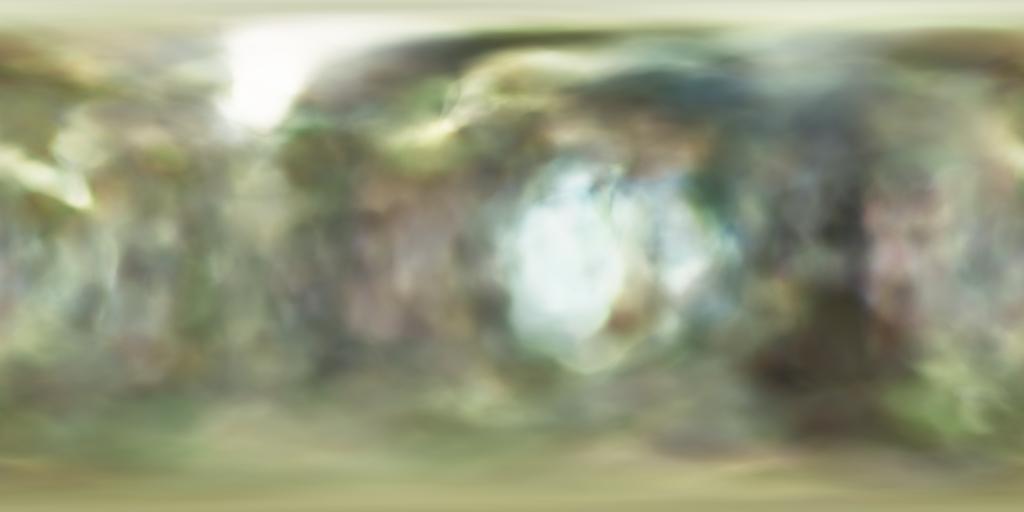}%
\end{subfigure}%
\end{subfigure}%
}
\\ \hdashline
%%%%%%%%%%%%%%%%%%%%%
{\rotatebox{90}{materials}} & 
\multicolumn{6}{c}
{
\begin{subfigure}[c]{0.166\linewidth}%
\includegraphics[width=0.94\linewidth]{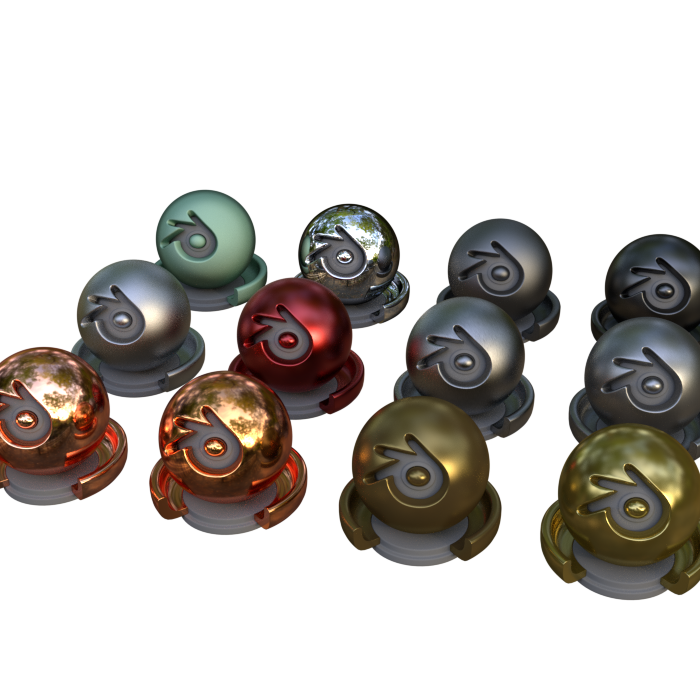}%
\end{subfigure}%
\begin{subfigure}[c]{0.166\linewidth}%
\includegraphics[width=0.94\linewidth]{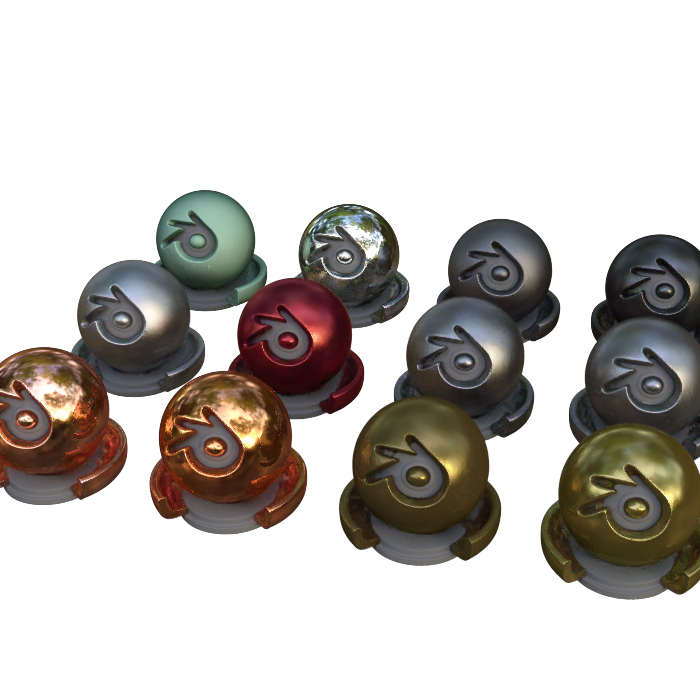}%
\end{subfigure}%
\begin{subfigure}[c]{0.166\linewidth}%
\includegraphics[width=0.94\linewidth]{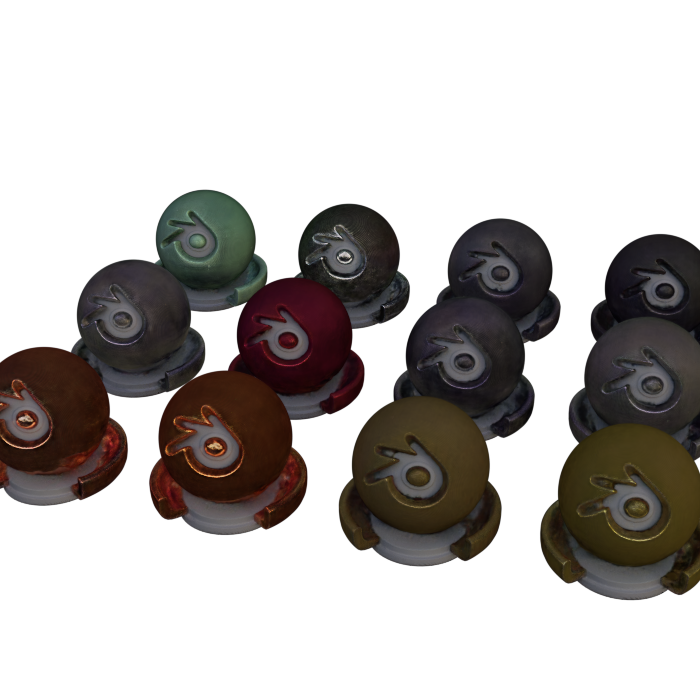}%
\end{subfigure}%
\begin{subfigure}[c]{0.166\linewidth}%
\includegraphics[width=0.94\linewidth]{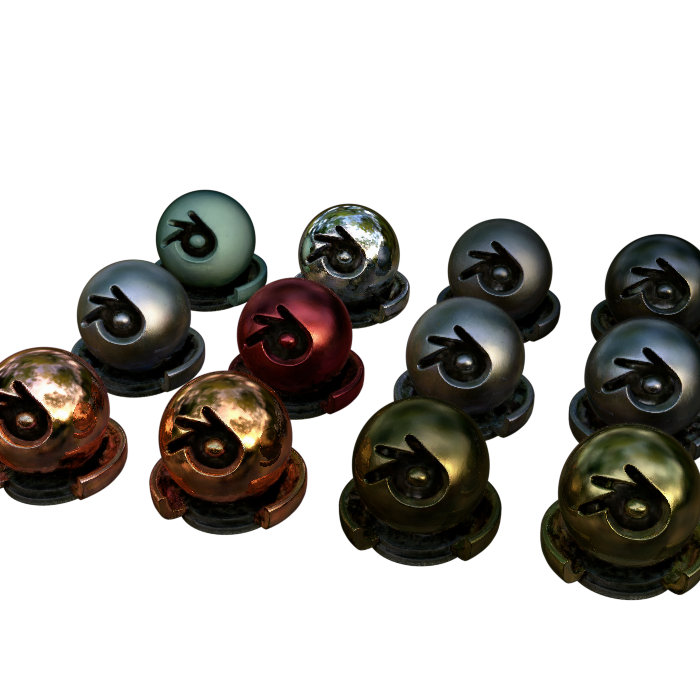}%
\end{subfigure}%
\begin{subfigure}[c]{0.166\linewidth}%
\includegraphics[width=0.94\linewidth]{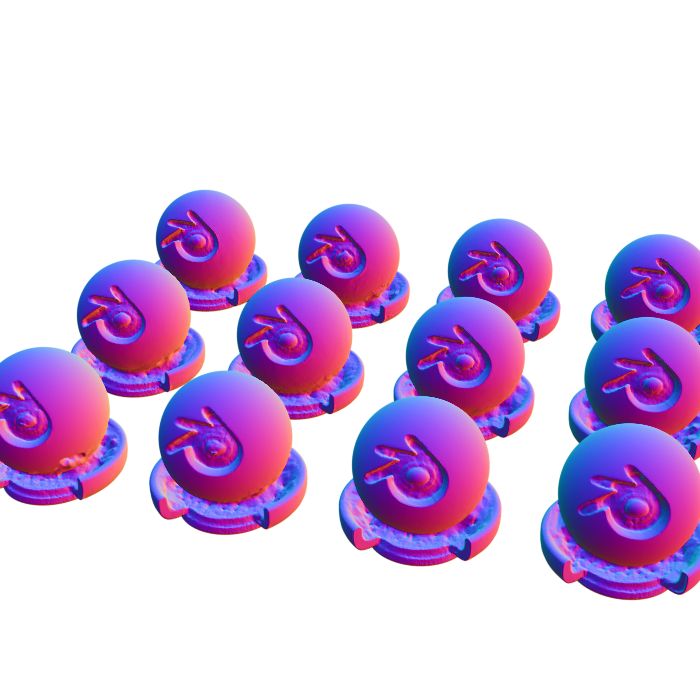}%
\end{subfigure}%
\begin{subfigure}[c]{0.166\linewidth}%
\begin{subfigure}[t!]{0.9\linewidth}%
\includegraphics[width=\linewidth]{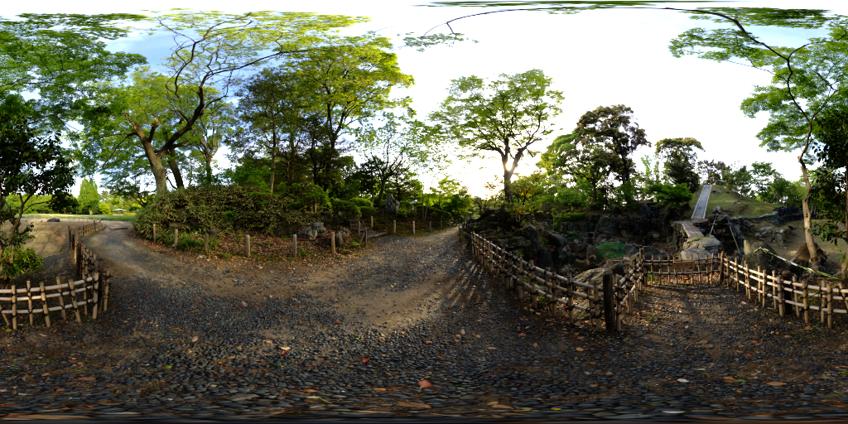}%
\end{subfigure}%
\\
\begin{subfigure}[b!]{0.9\linewidth}%
\includegraphics[width=\linewidth]{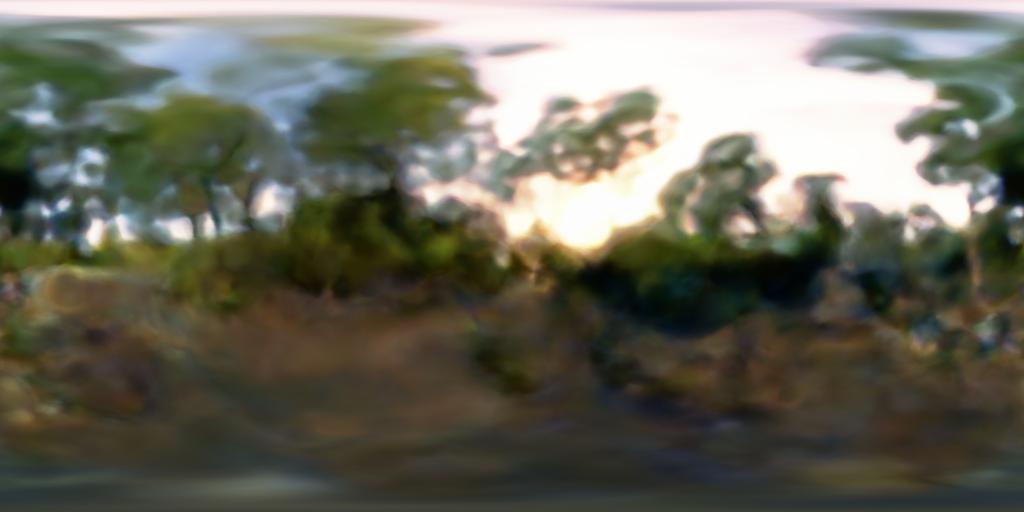}%
\end{subfigure}%
\end{subfigure}%
}
\\ \hdashline
%%%%%%%%%%%%%%%%%%%%%%
{\rotatebox{90}{ball}} & 
\multicolumn{6}{c}
{
\begin{subfigure}[c]{0.166\linewidth}%
\includegraphics[width=0.94\linewidth]{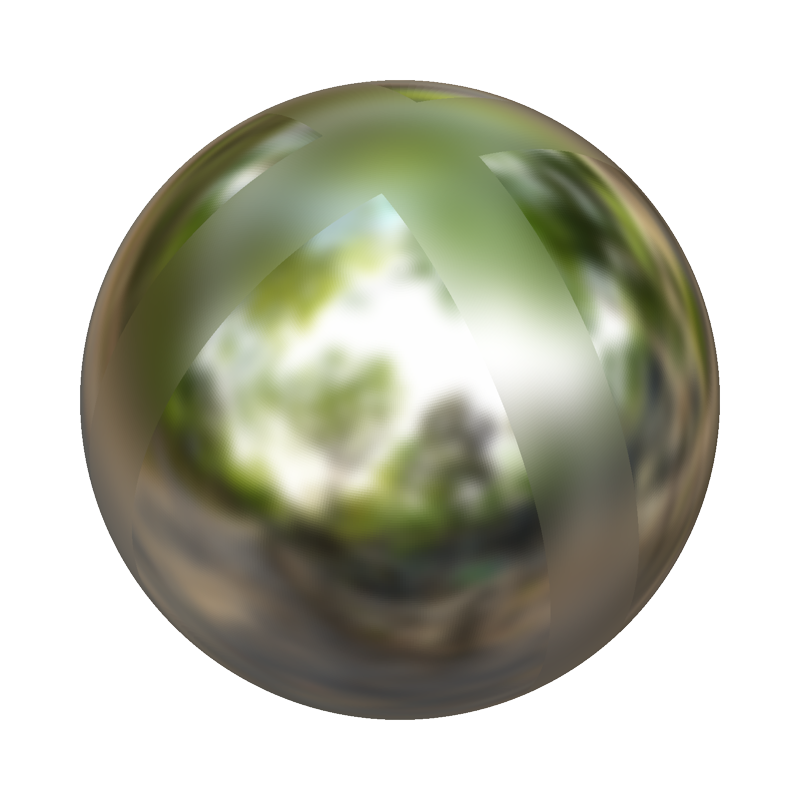}%
\end{subfigure}%
\begin{subfigure}[c]{0.166\linewidth}%
\includegraphics[width=0.94\linewidth]{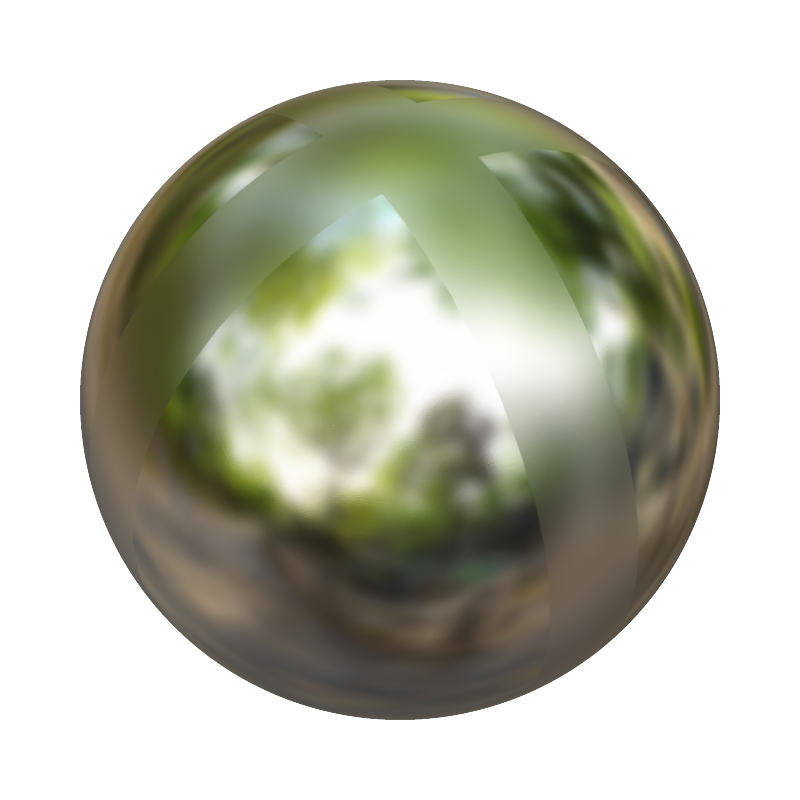}%
\end{subfigure}%
\begin{subfigure}[c]{0.166\linewidth}%
\includegraphics[width=0.94\linewidth]{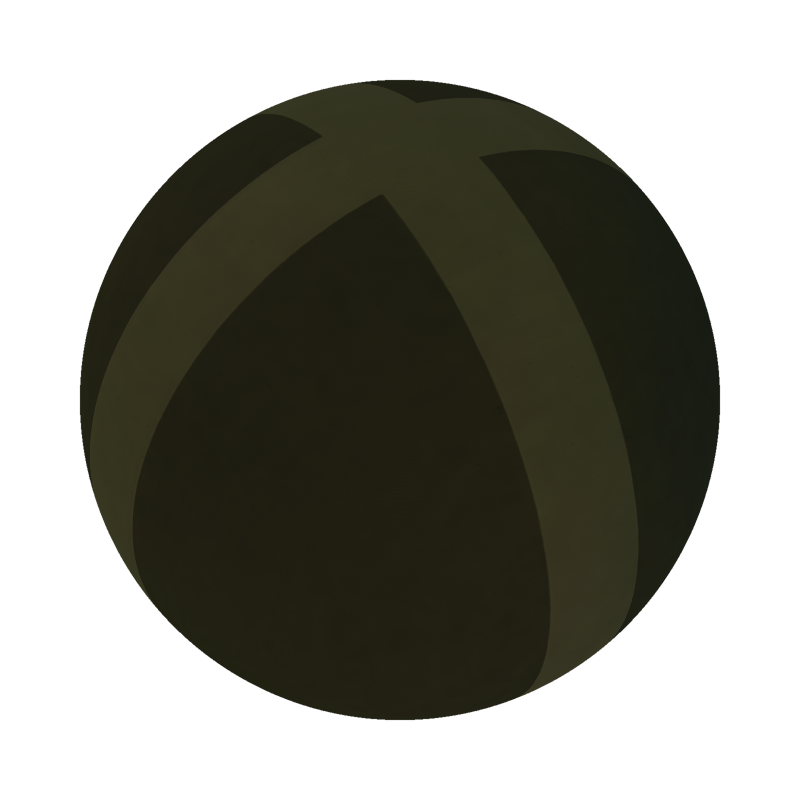}%
\end{subfigure}%
\begin{subfigure}[c]{0.166\linewidth}%
\includegraphics[width=0.94\linewidth]{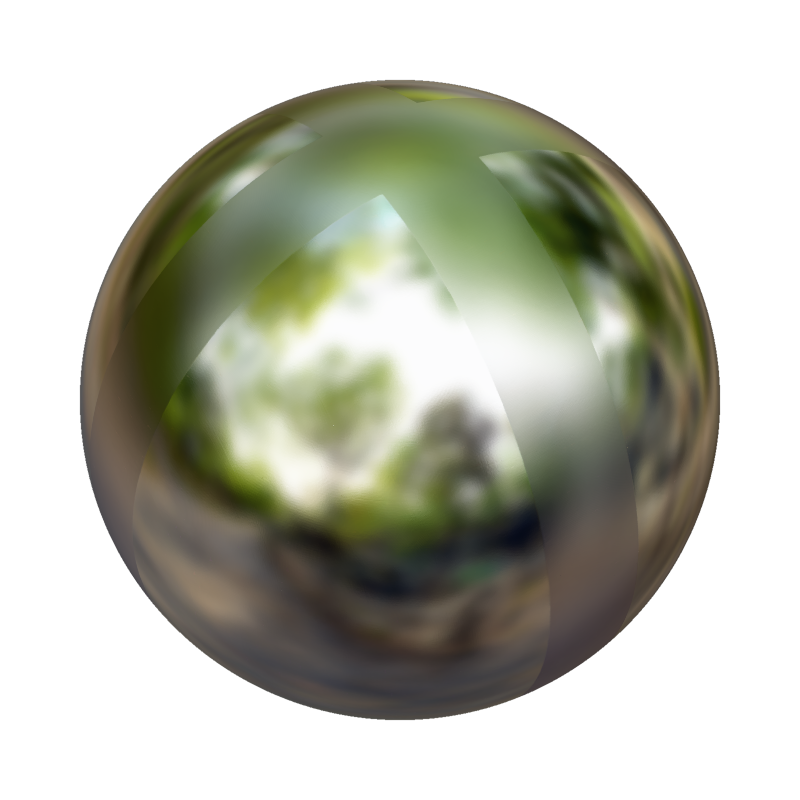}%
\end{subfigure}%
\begin{subfigure}[c]{0.166\linewidth}%
\includegraphics[width=0.94\linewidth]{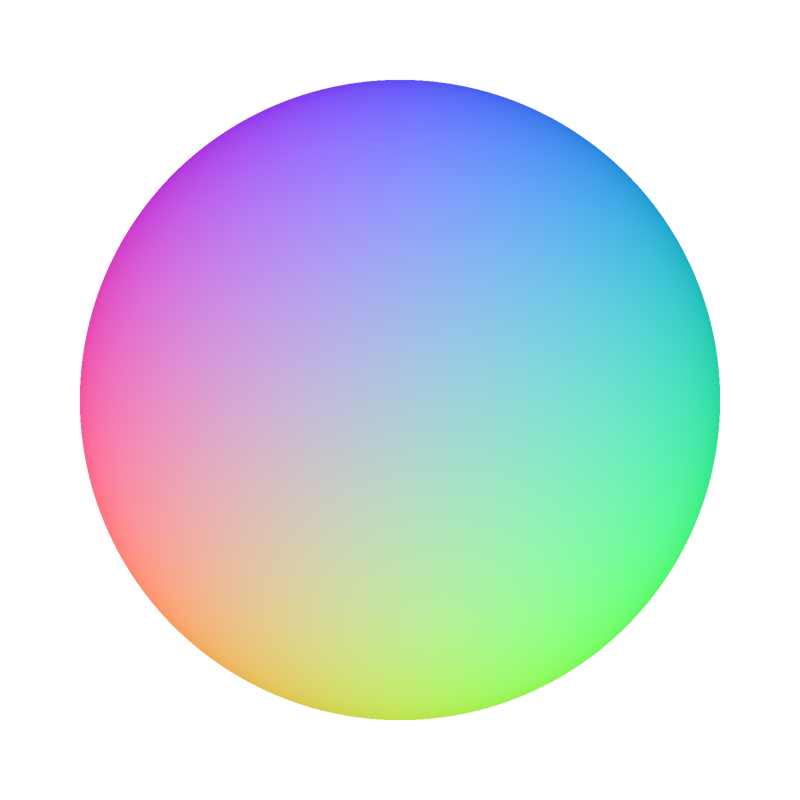}%
\end{subfigure}%
\begin{subfigure}[c]{0.166\linewidth}%
\begin{subfigure}[t!]{0.9\linewidth}%
\includegraphics[width=\linewidth]{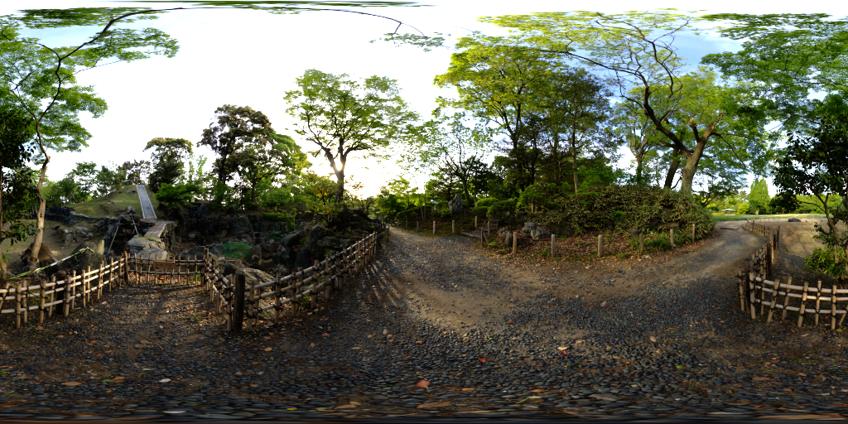}%
\end{subfigure}%
\\
\begin{subfigure}[b!]{0.9\linewidth}%
\includegraphics[width=\linewidth]{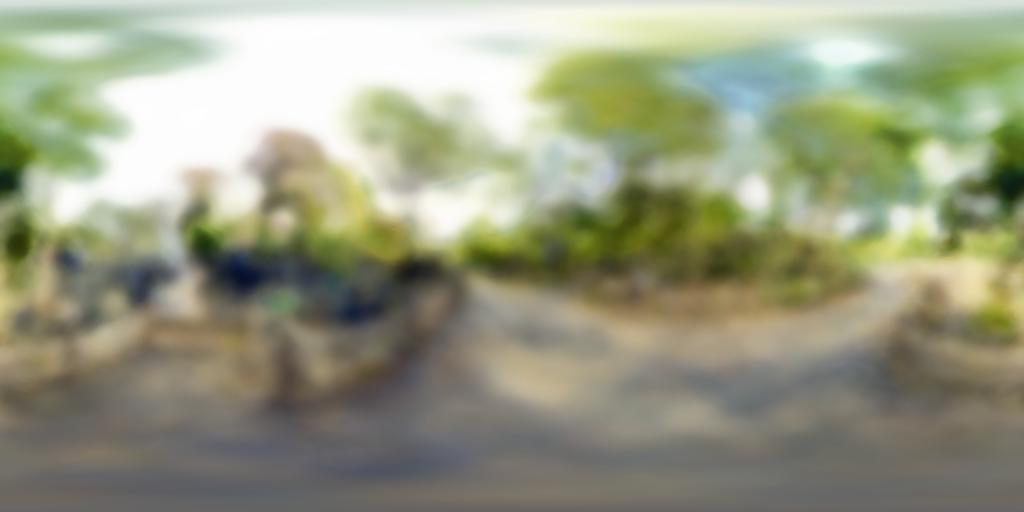}%
\end{subfigure}%
\end{subfigure}%
}
\\ \hdashline
%%%%%%%%%%%%%%%%%%%%%%
{\rotatebox{90}{car}} & 
\multicolumn{6}{c}
{
\begin{subfigure}[c]{0.166\linewidth}%
\includegraphics[width=0.94\linewidth]{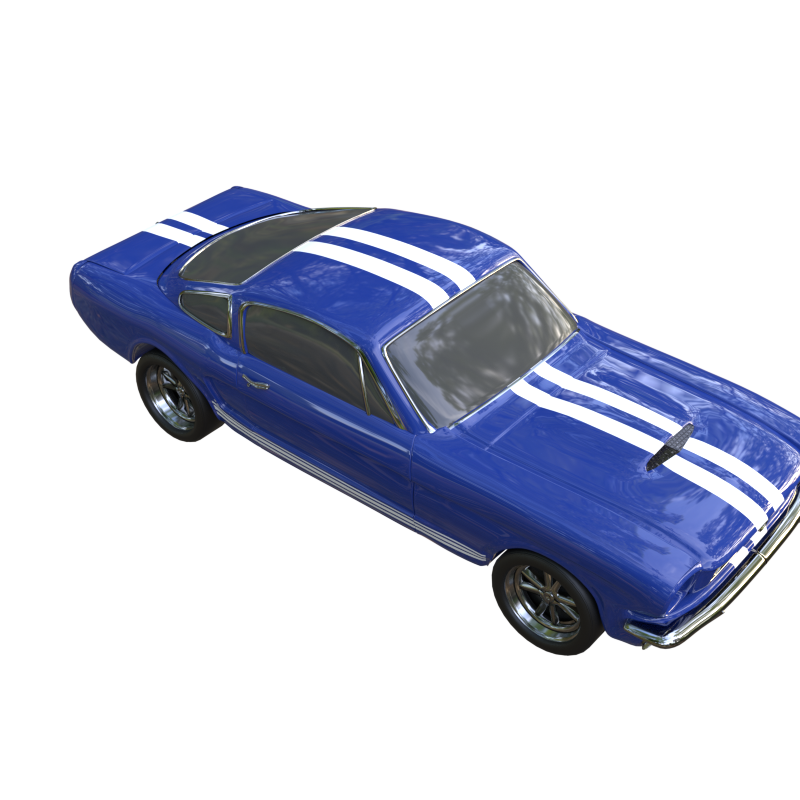}%
\end{subfigure}%
\begin{subfigure}[c]{0.166\linewidth}%
\includegraphics[width=0.94\linewidth]{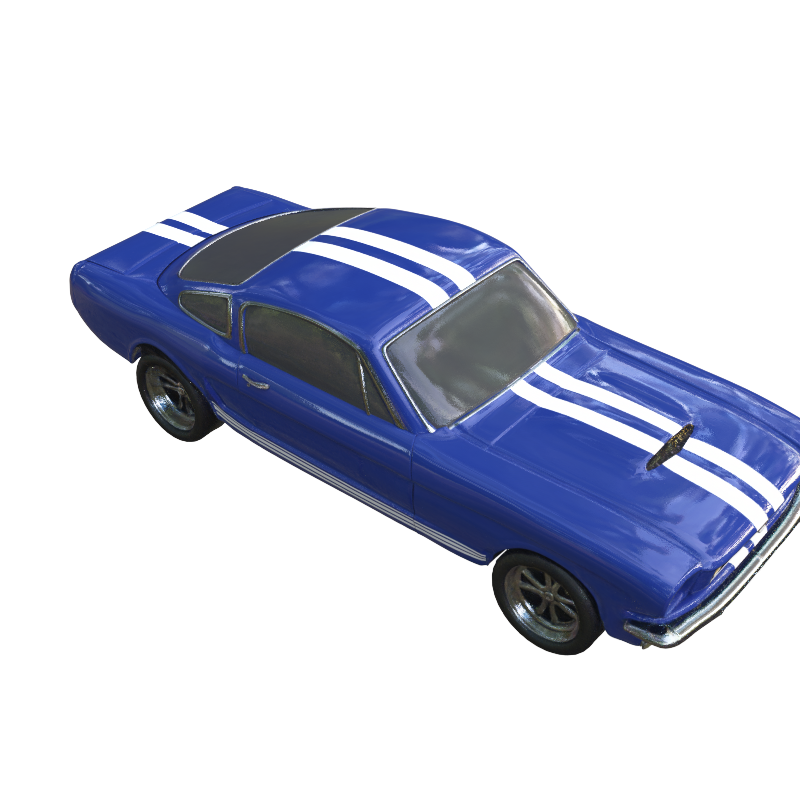}%
\end{subfigure}%
\begin{subfigure}[c]{0.166\linewidth}%
\includegraphics[width=0.94\linewidth]{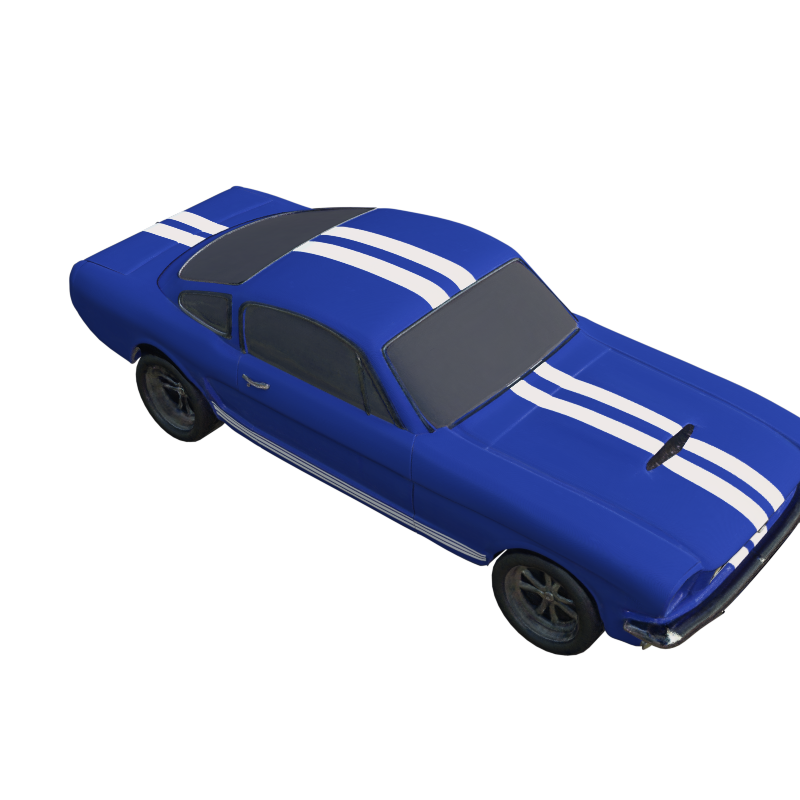}%
\end{subfigure}%
\begin{subfigure}[c]{0.166\linewidth}%
\includegraphics[width=0.94\linewidth]{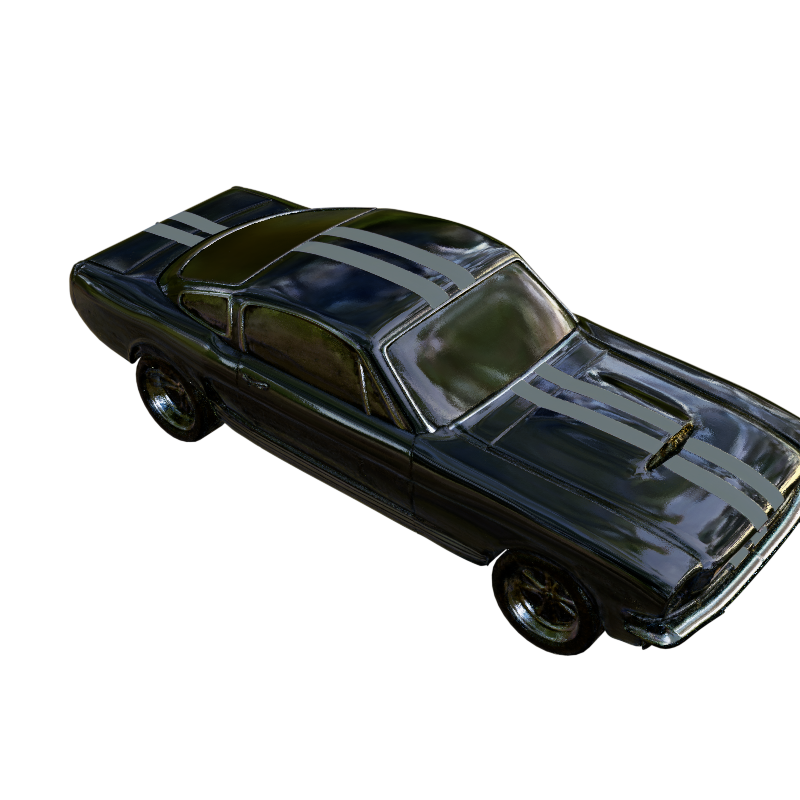}%
\end{subfigure}%
\begin{subfigure}[c]{0.166\linewidth}%
\includegraphics[width=0.94\linewidth]{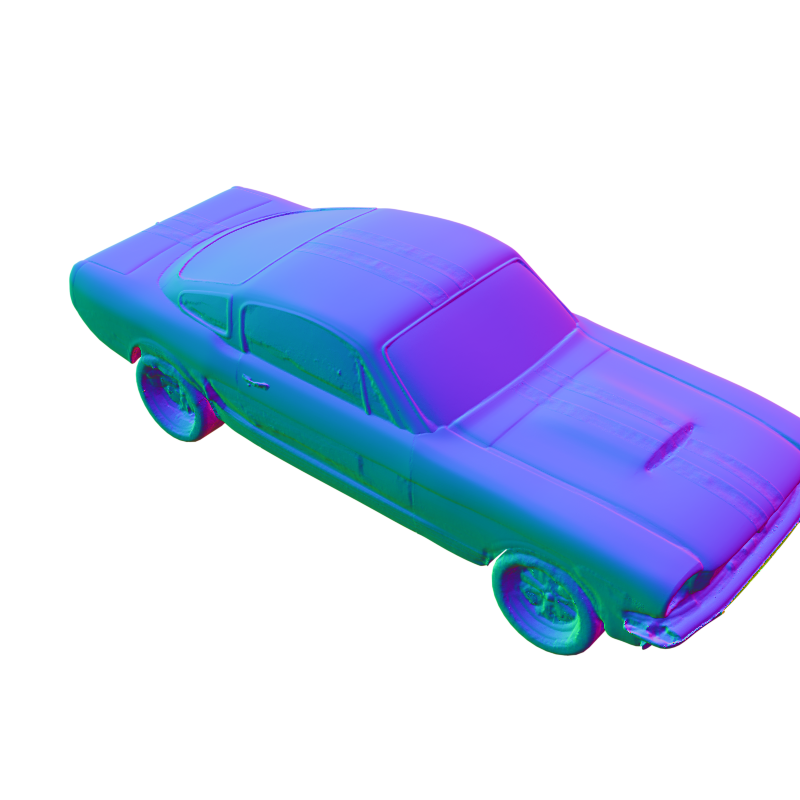}%
\end{subfigure}%
\begin{subfigure}[c]{0.166\linewidth}%
\begin{subfigure}[t!]{0.9\linewidth}%
\includegraphics[width=\linewidth]{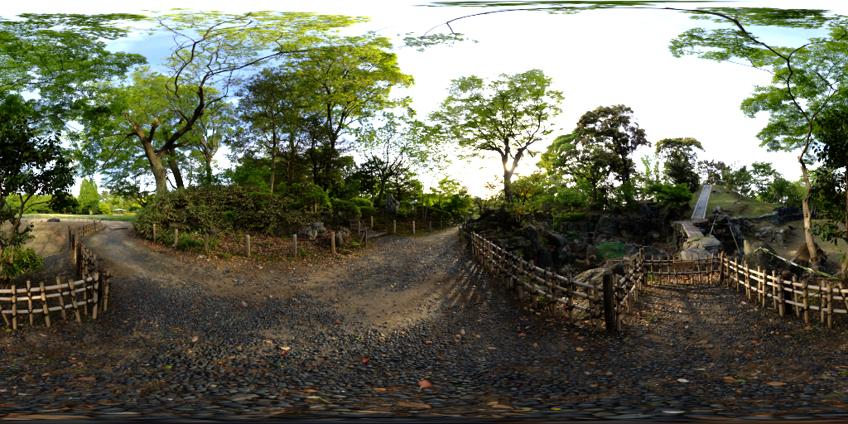}%
\end{subfigure}%
\\
\begin{subfigure}[b!]{0.9\linewidth}%
\includegraphics[width=\linewidth]{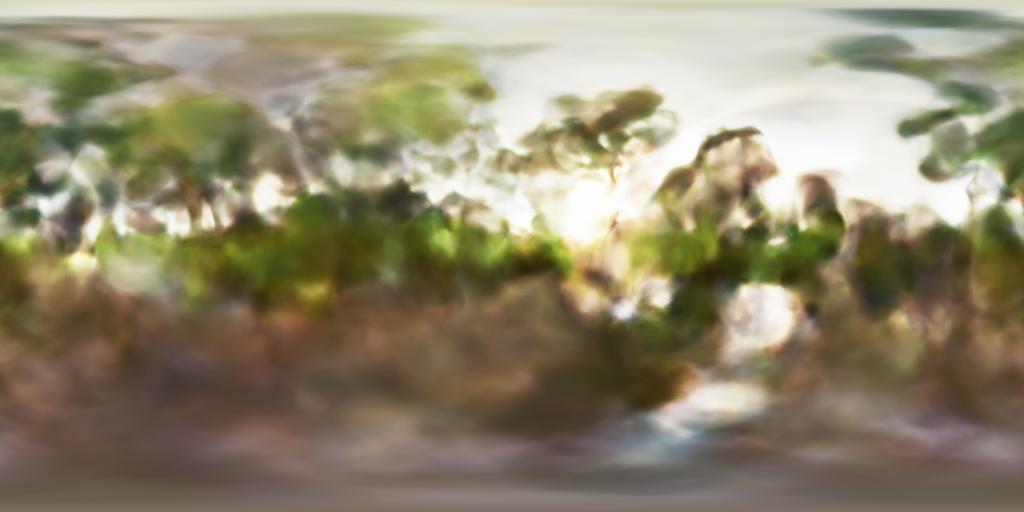}%
\end{subfigure}%
\end{subfigure}%
}
\\ \hdashline
%%%%%%%%%%%%%%%%%%%%%%
{\rotatebox{90}{toaster}} & 
\multicolumn{6}{c}
{
\begin{subfigure}[c]{0.166\linewidth}%
\includegraphics[width=0.94\linewidth]{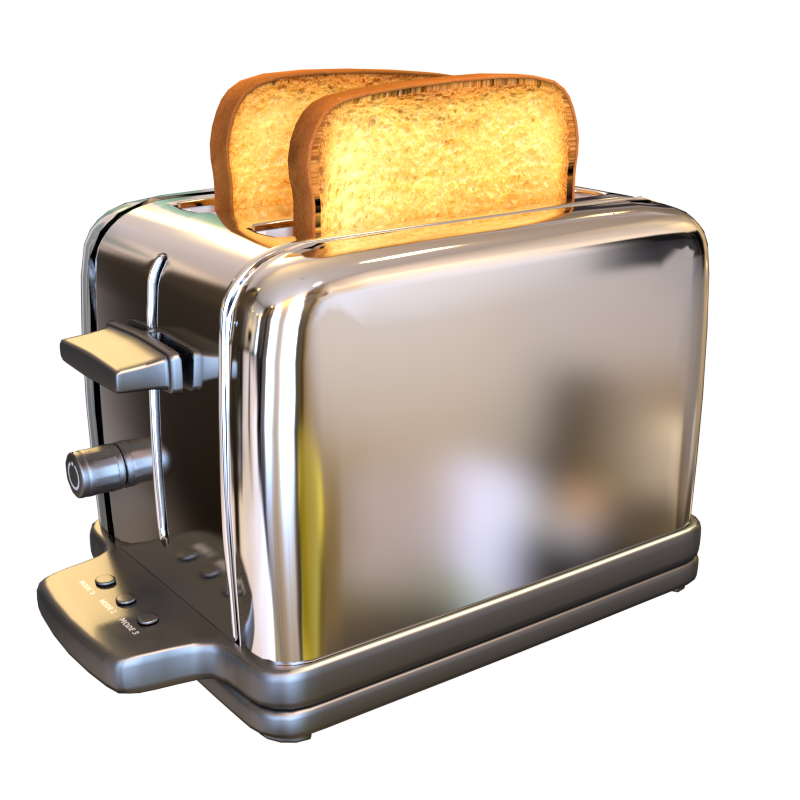}%
\end{subfigure}%
\begin{subfigure}[c]{0.166\linewidth}%
\includegraphics[width=0.94\linewidth]{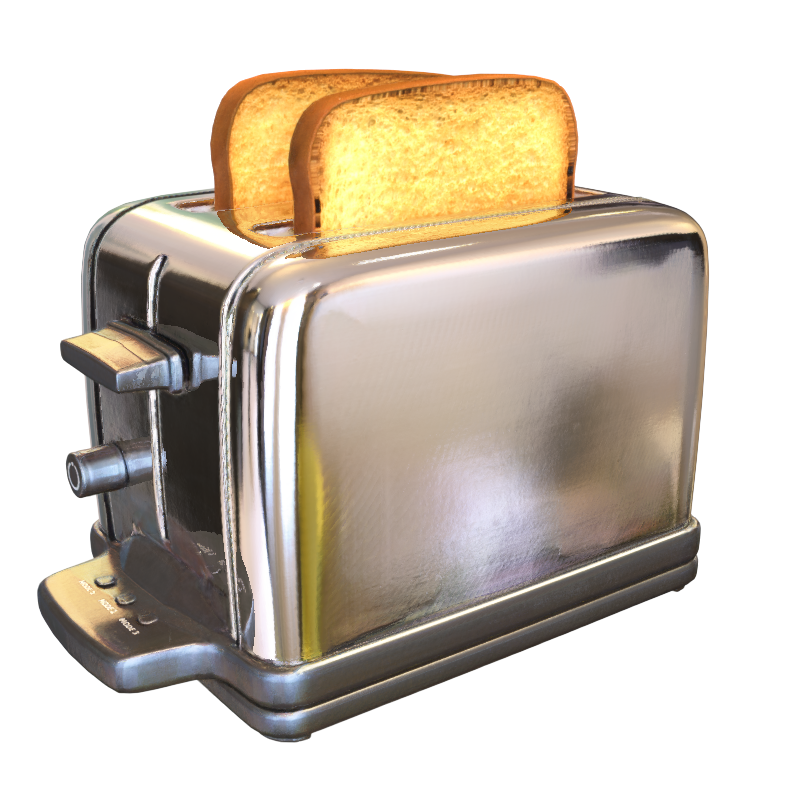}%
\end{subfigure}%
\begin{subfigure}[c]{0.166\linewidth}%
\includegraphics[width=0.94\linewidth]{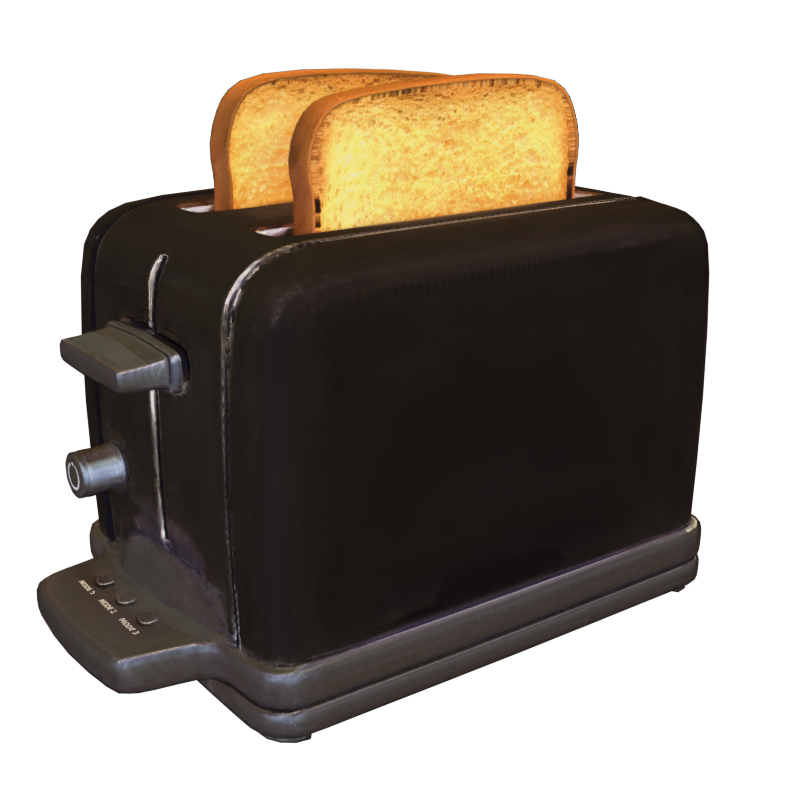}%
\end{subfigure}%
\begin{subfigure}[c]{0.166\linewidth}%
\includegraphics[width=0.94\linewidth]{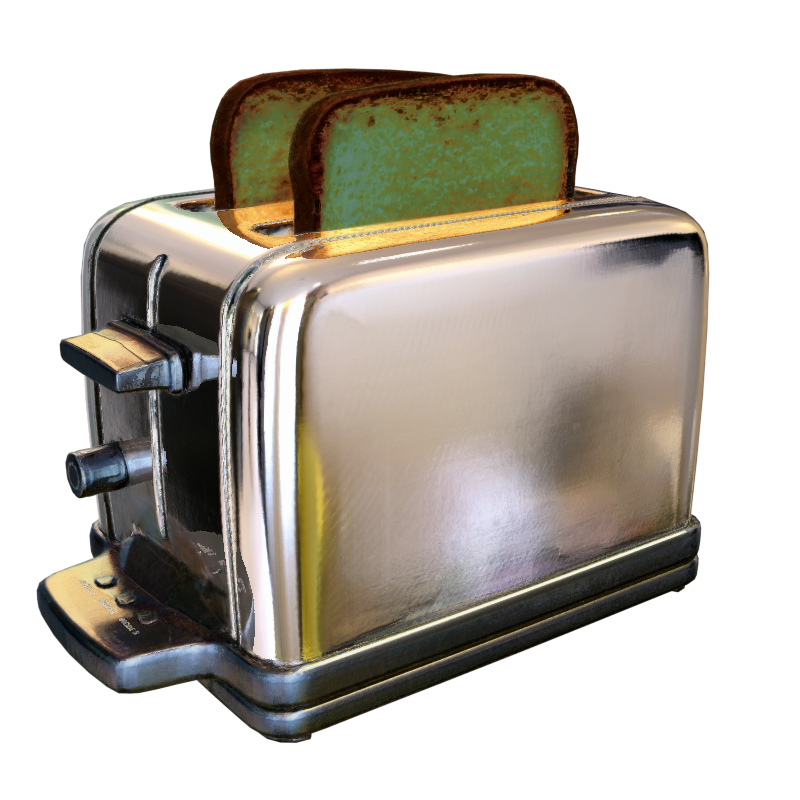}%
\end{subfigure}%
\begin{subfigure}[c]{0.166\linewidth}%
\includegraphics[width=0.94\linewidth]{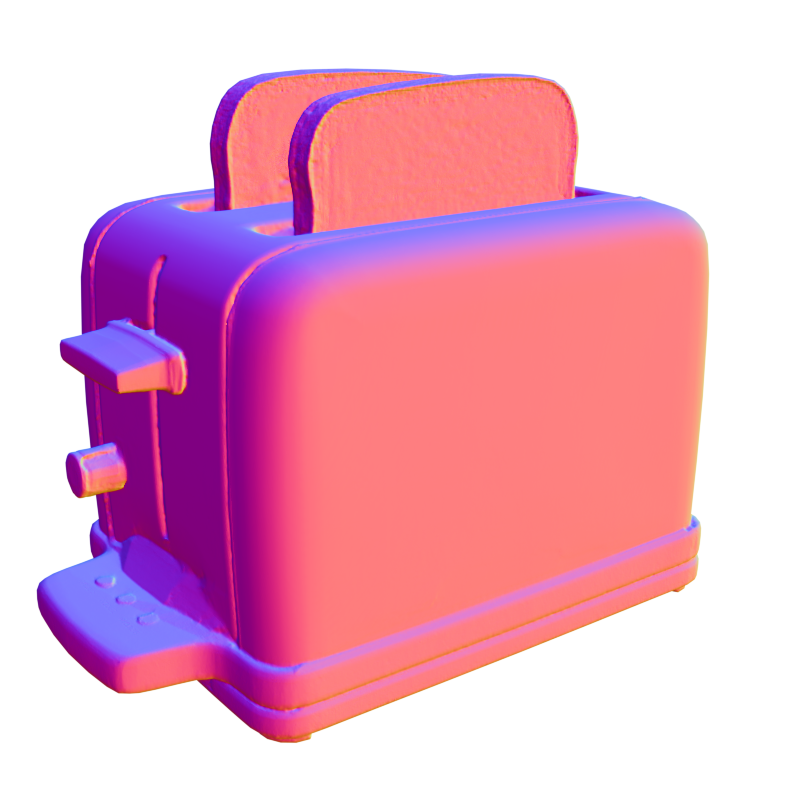}%
\end{subfigure}%
\begin{subfigure}[c]{0.166\linewidth}%
\begin{subfigure}[t!]{0.9\linewidth}%
\includegraphics[width=\linewidth]{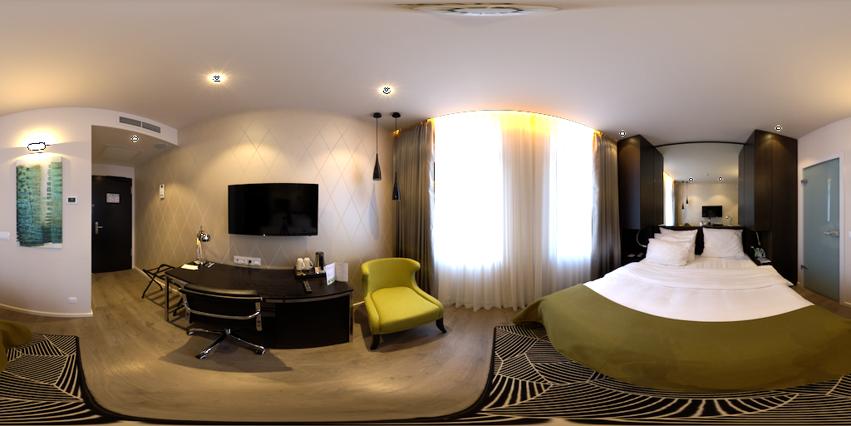}%
\end{subfigure}%
\\
\begin{subfigure}[b!]{0.9\linewidth}%
\includegraphics[width=\linewidth]{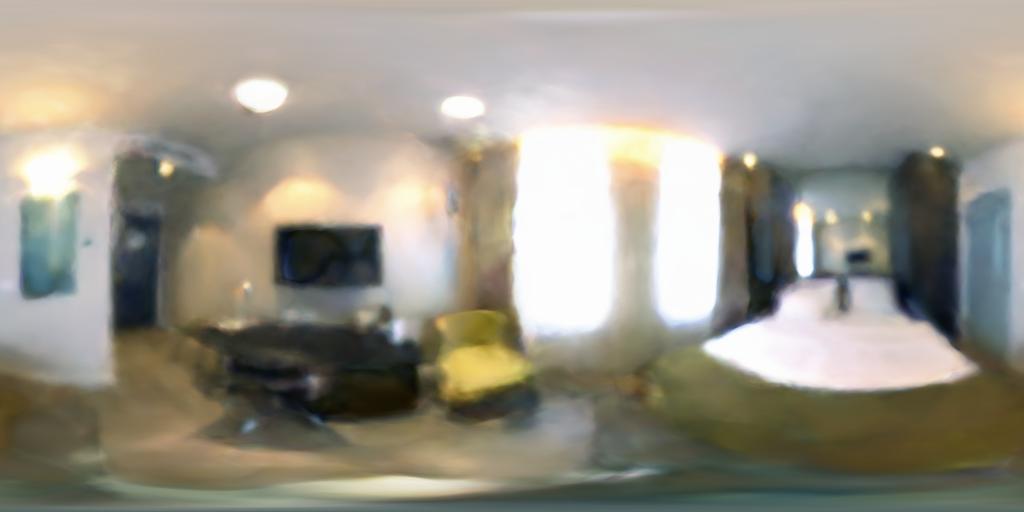}%
\end{subfigure}%
\end{subfigure}%
}
\\ \hdashline
%%%%%%%%%%%%%%%%%%%%%%
{\rotatebox{90}{helmet}} & 
\multicolumn{6}{c}
{
\begin{subfigure}[c]{0.166\linewidth}%
\includegraphics[width=0.94\linewidth]{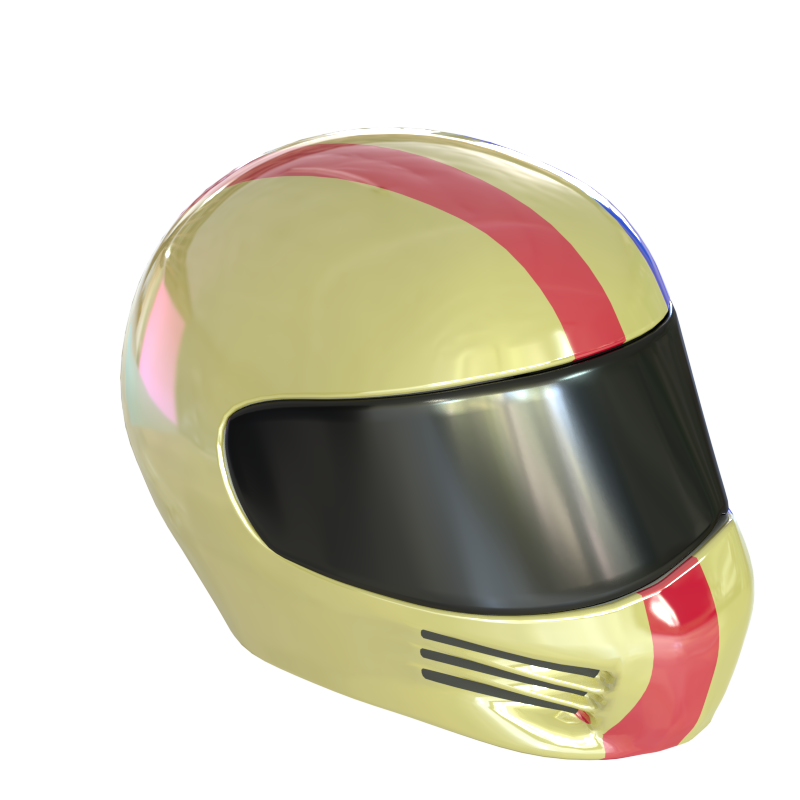}%
\end{subfigure}%
\begin{subfigure}[c]{0.166\linewidth}%
\includegraphics[width=0.94\linewidth]{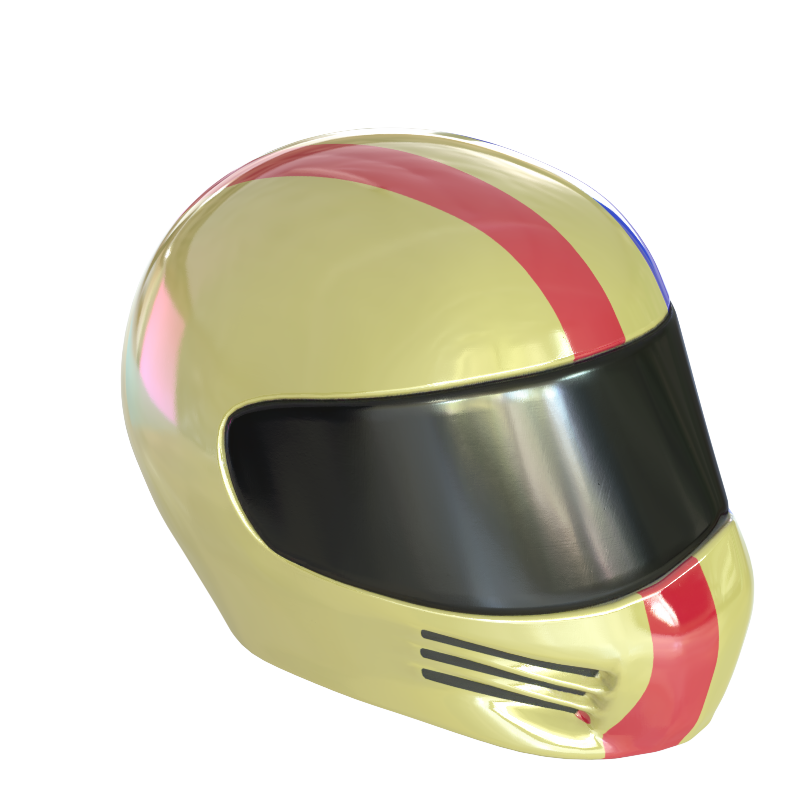}%
\end{subfigure}%
\begin{subfigure}[c]{0.166\linewidth}%
\includegraphics[width=0.94\linewidth]{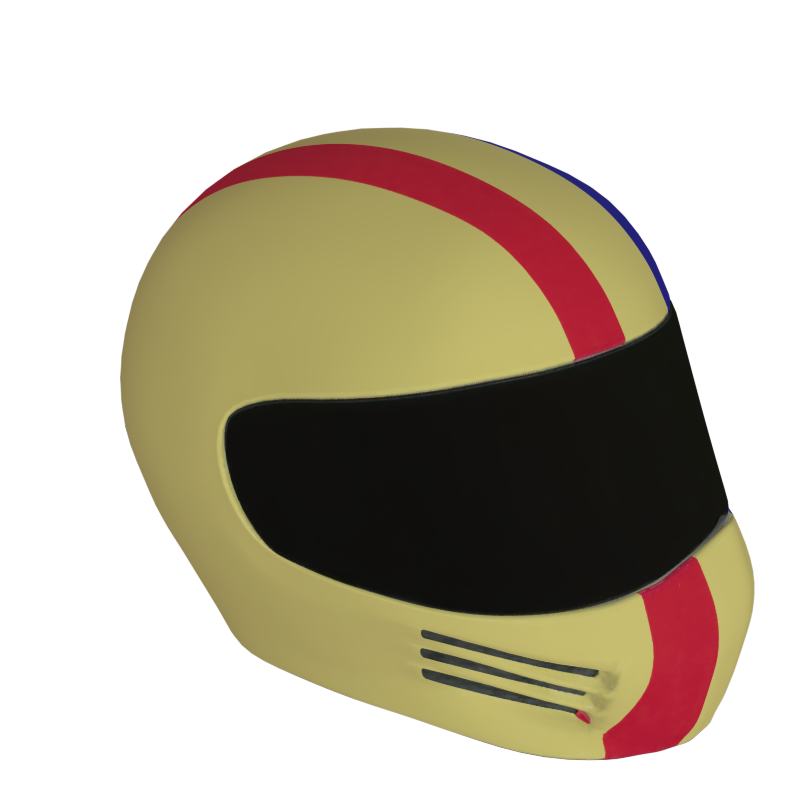}%
\end{subfigure}%
\begin{subfigure}[c]{0.166\linewidth}%
\includegraphics[width=0.94\linewidth]{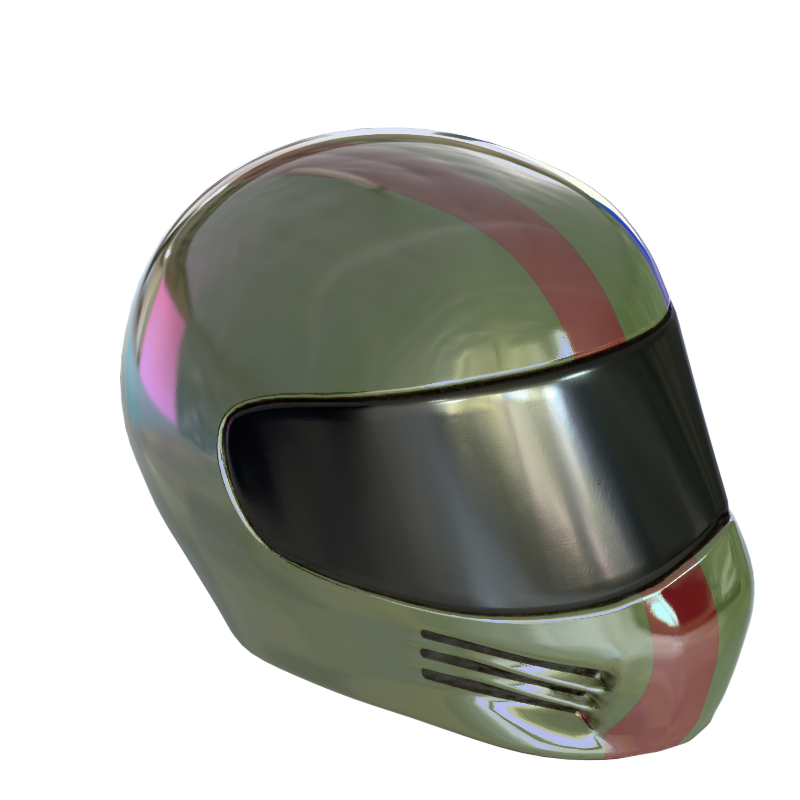}%
\end{subfigure}%
\begin{subfigure}[c]{0.166\linewidth}%
\includegraphics[width=0.94\linewidth]{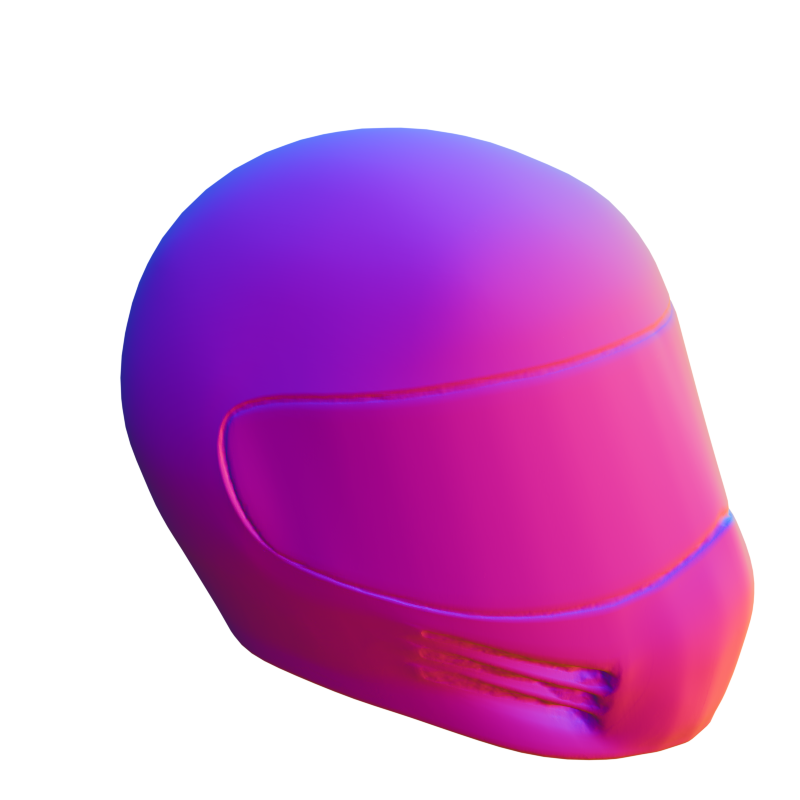}%
\end{subfigure}%
\begin{subfigure}[c]{0.166\linewidth}%
\begin{subfigure}[t!]{0.9\linewidth}%
\includegraphics[width=\linewidth]{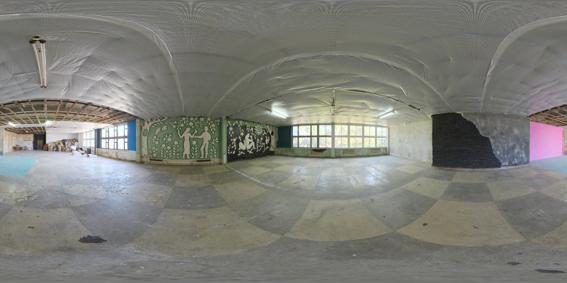}%
\end{subfigure}%
\\
\begin{subfigure}[b!]{0.9\linewidth}%
\includegraphics[width=\linewidth]{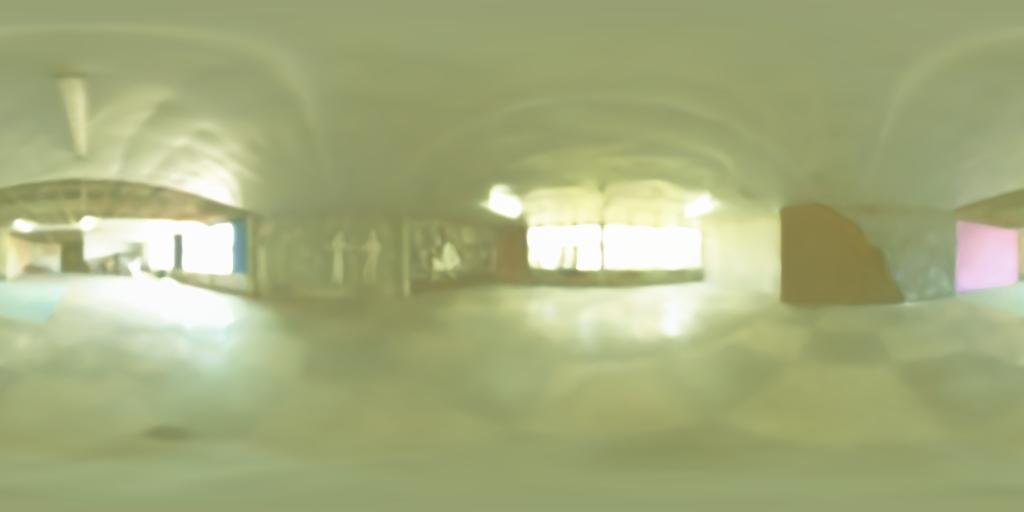}%
\end{subfigure}%
\end{subfigure}%
}
\\ \hdashline
%%%%%%%%%%%%%%%%%%%%%%
{\rotatebox{90}{teapot}} & 
\multicolumn{6}{c}
{
\begin{subfigure}[c]{0.166\linewidth}%
\includegraphics[width=0.94\linewidth]{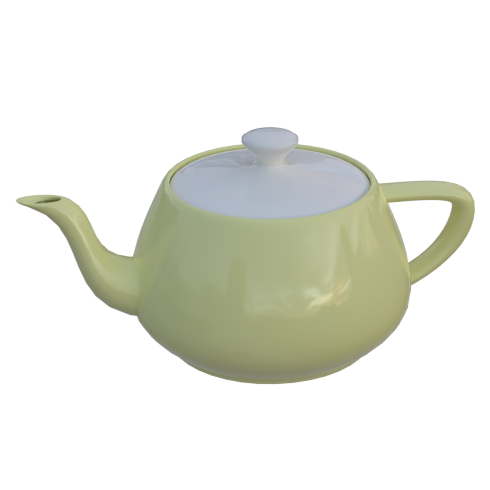}%
\end{subfigure}%
\begin{subfigure}[c]{0.166\linewidth}%
\includegraphics[width=0.94\linewidth]{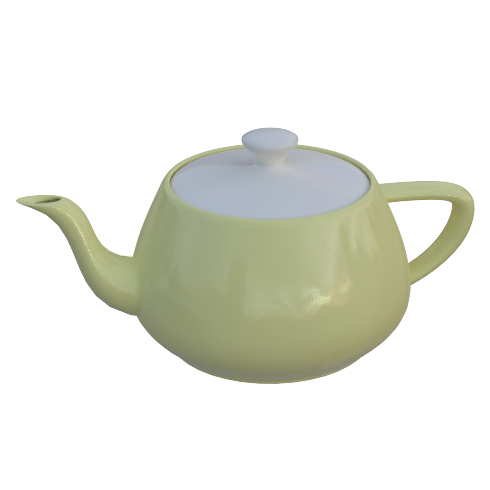}%
\end{subfigure}%
\begin{subfigure}[c]{0.166\linewidth}%
\includegraphics[width=0.94\linewidth]{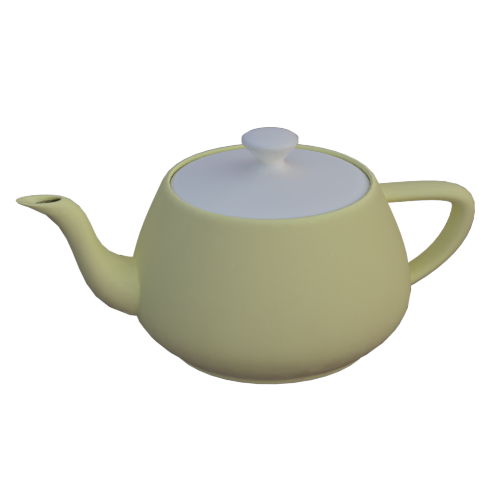}%
\end{subfigure}%
\begin{subfigure}[c]{0.166\linewidth}%
\includegraphics[width=0.94\linewidth]{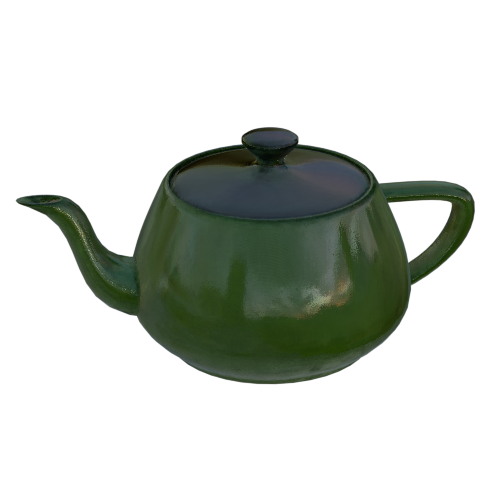}%
\end{subfigure}%
\begin{subfigure}[c]{0.166\linewidth}%
\includegraphics[width=0.94\linewidth]{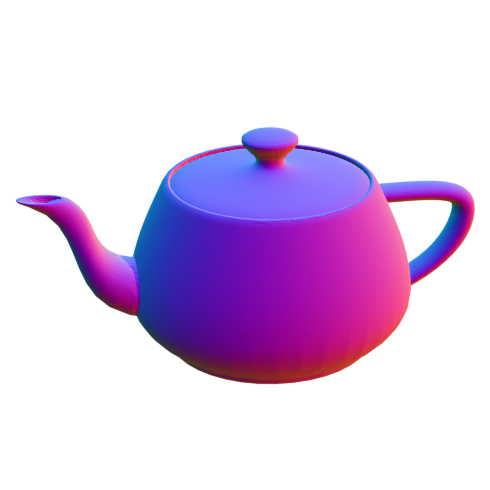}%
\end{subfigure}%
\begin{subfigure}[c]{0.166\linewidth}%
\begin{subfigure}[t!]{0.9\linewidth}%
\includegraphics[width=\linewidth]{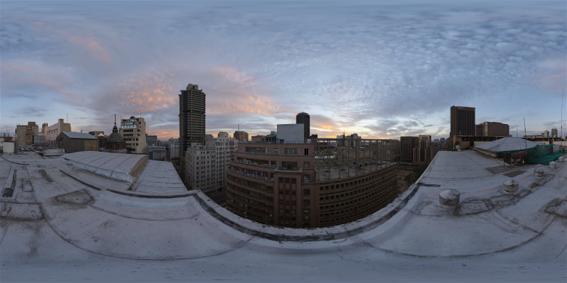}%
\end{subfigure}%
\\
\begin{subfigure}[b!]{0.9\linewidth}%
\includegraphics[width=\linewidth]{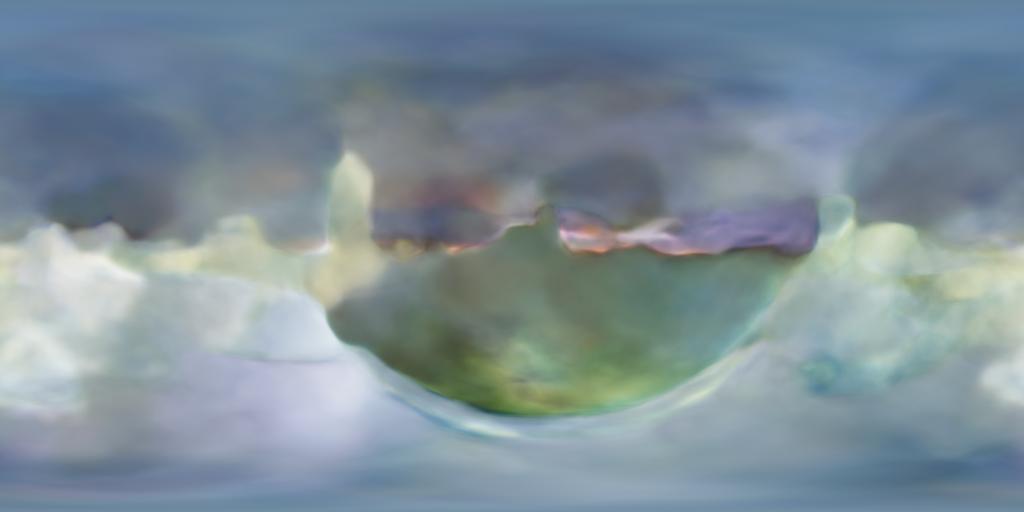}%
\end{subfigure}%
\end{subfigure}%
}
\\ \hdashline
%%%%%%%%%%%%%%%%%%%%%%
{\rotatebox{90}{coffee}} & 
\multicolumn{6}{c}
{
\begin{subfigure}[c]{0.166\linewidth}%
\includegraphics[width=0.94\linewidth]{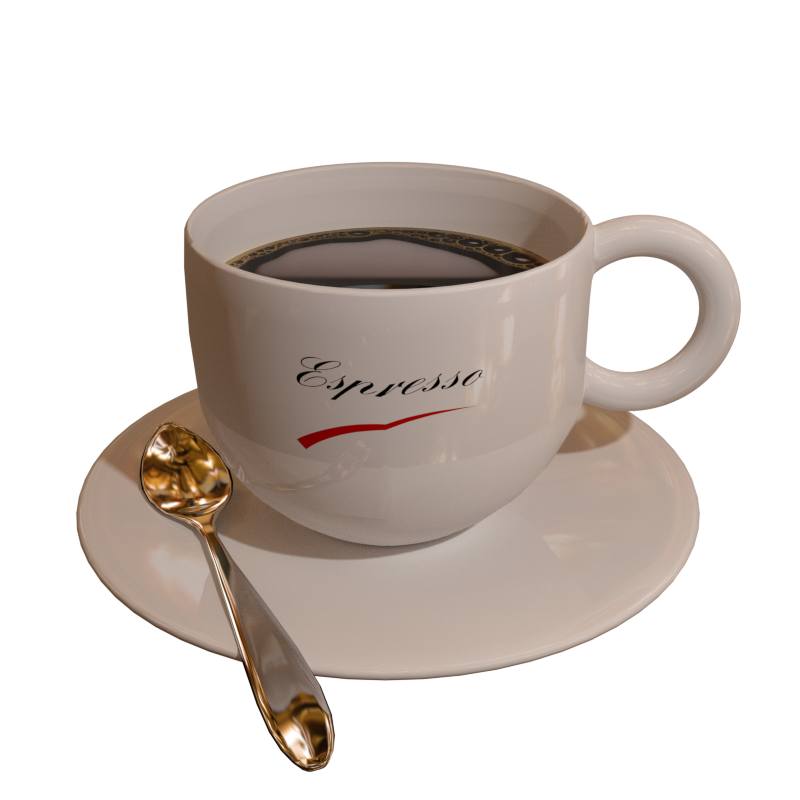}%
\end{subfigure}%
\begin{subfigure}[c]{0.166\linewidth}%
\includegraphics[width=0.94\linewidth]{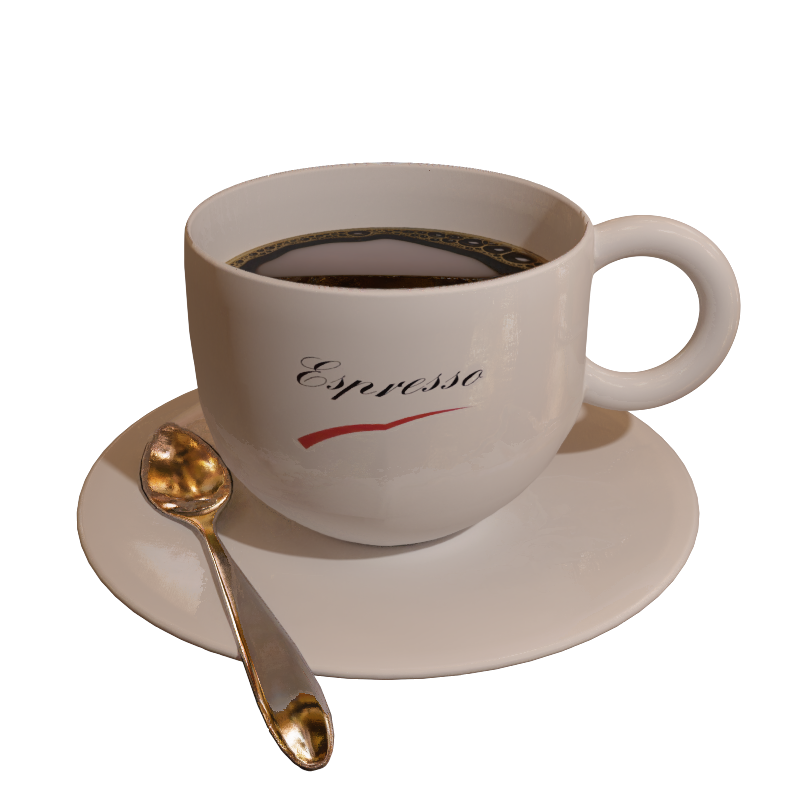}%
\end{subfigure}%
\begin{subfigure}[c]{0.166\linewidth}%
\includegraphics[width=0.94\linewidth]{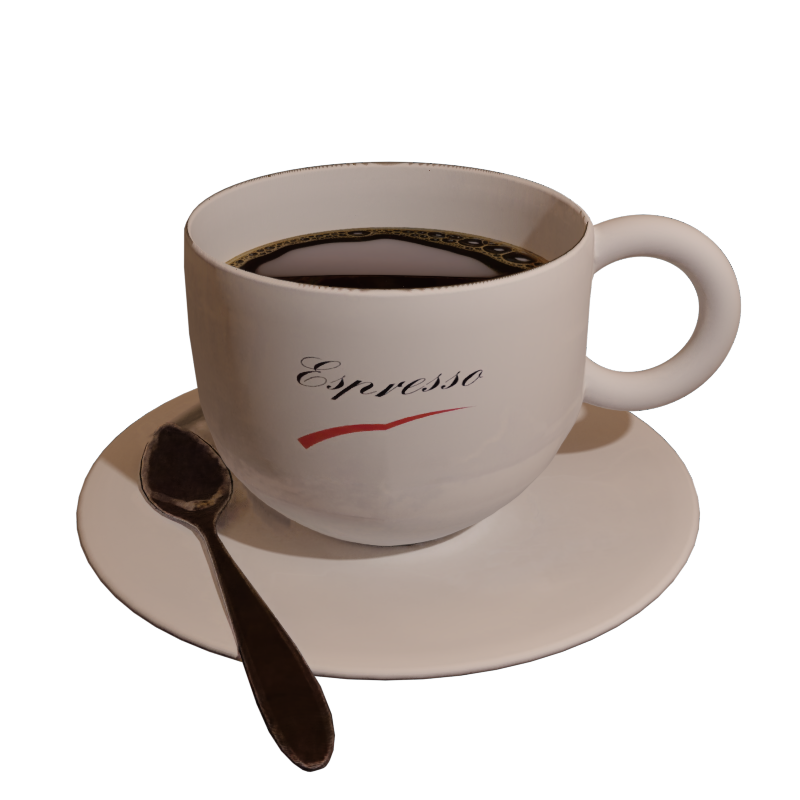}%
\end{subfigure}%
\begin{subfigure}[c]{0.166\linewidth}%
\includegraphics[width=0.94\linewidth]{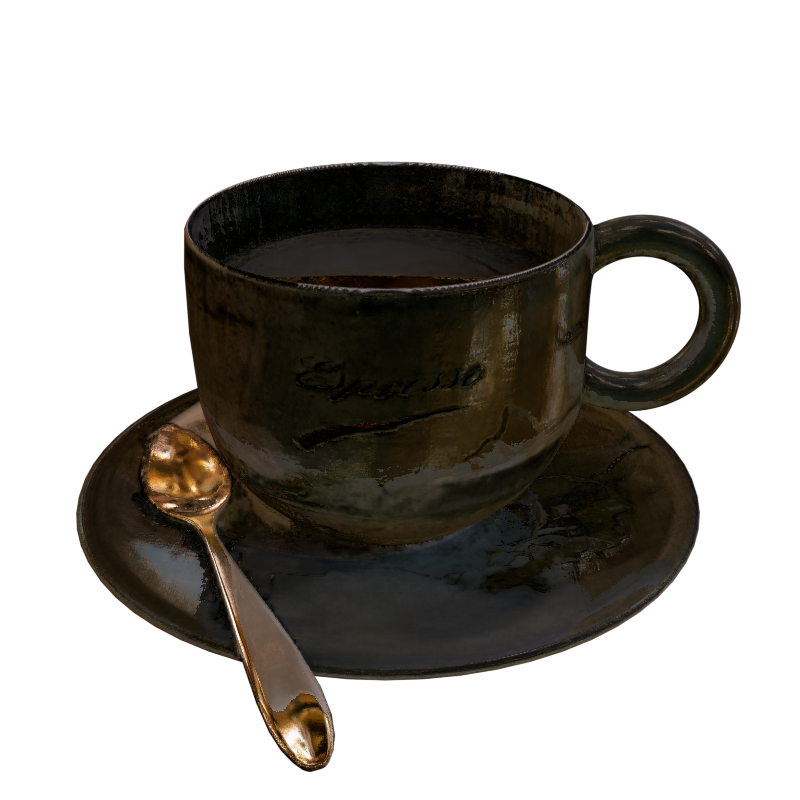}%
\end{subfigure}%
\begin{subfigure}[c]{0.166\linewidth}%
\includegraphics[width=0.94\linewidth]{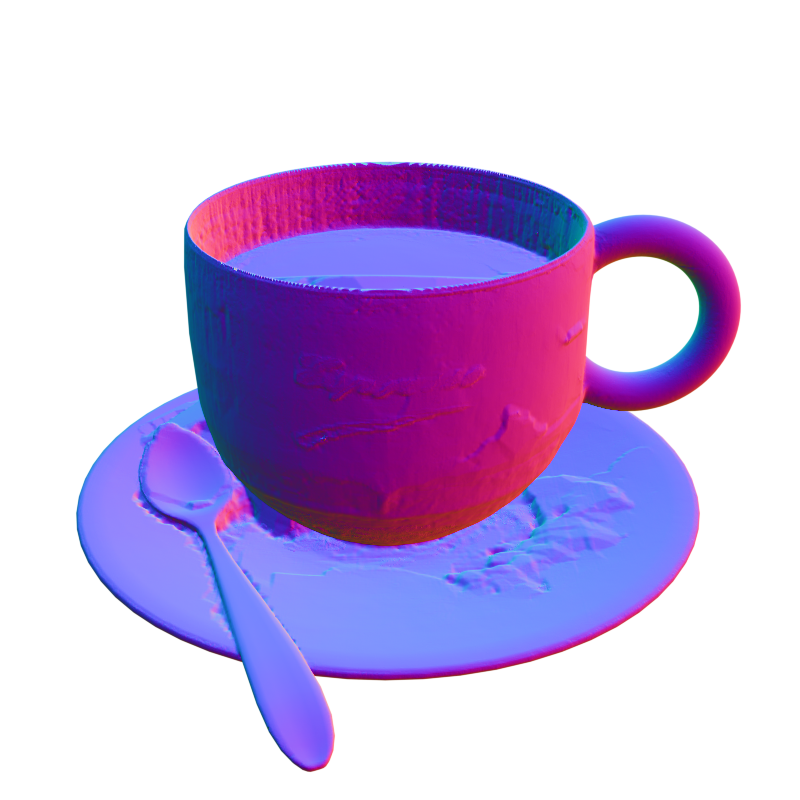}%
\end{subfigure}%
\begin{subfigure}[c]{0.166\linewidth}%
\begin{subfigure}[t!]{0.9\linewidth}%
\includegraphics[width=\linewidth]{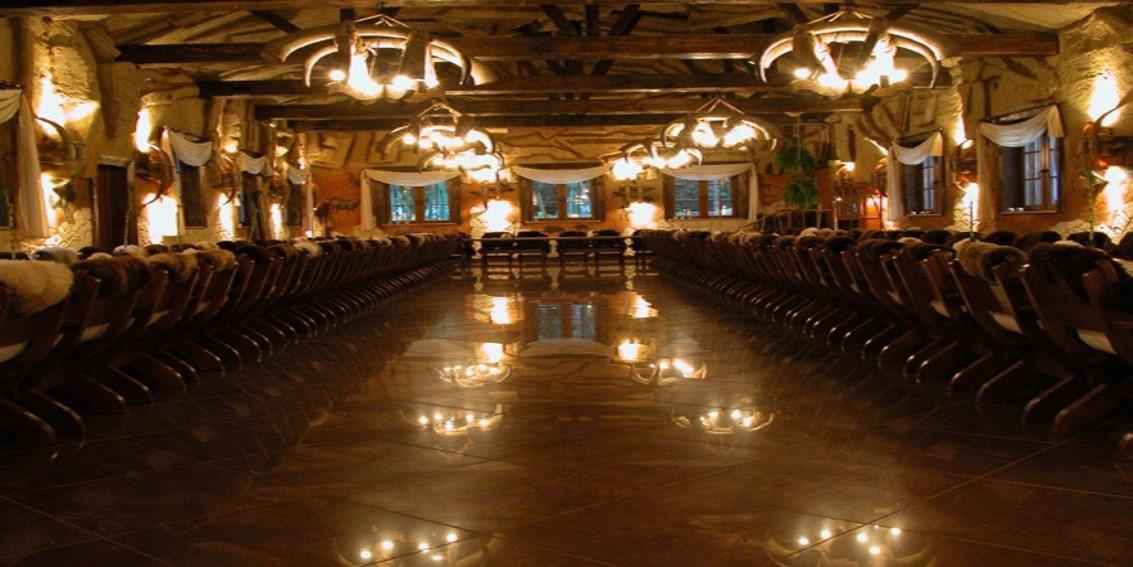}%
\end{subfigure}%
\\
\begin{subfigure}[b!]{0.9\linewidth}%
\includegraphics[width=\linewidth]{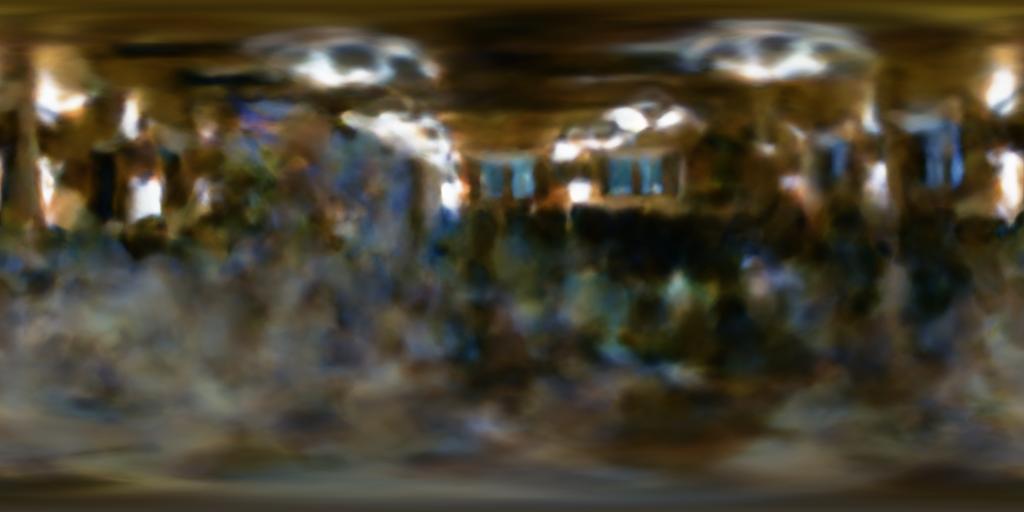}%
\end{subfigure}%
\end{subfigure}%
}
\\ \hdashline

     & Ground truth & Our Rendering & Diffuse & Specular & Normal & Probes
\end{tabularx}%
        
    \makeatletter\def\@captype{figure}\makeatother
    \caption{\small Our decompositions. In the ``Probes" column, the upper row shows the reference, and the lower row shows our estimation.
    \label{fig:perscene}}
\end{table*}

% \subsection{Failure Cases}

{\small
\bibliographystyle{ieee_fullname}
\bibliography{egbib}
}

\end{document}